\documentclass{elsarticle}

\newcommand{\myreferences}{../../../bibliography/article_arXiv}

\usepackage{amsmath}
\usepackage{url}
\usepackage{enumerate}
\usepackage{multirow}
\usepackage{paralist}
\usepackage{epsfig}
\usepackage{subfigure}
\usepackage{booktabs}
\usepackage{tabularx}
\usepackage{graphicx}
\usepackage[justification=justified]{caption}
\usepackage{color}
\usepackage{enumitem}
\usepackage{hyperref}
\usepackage{xcolor}
\usepackage{tablefootnote}
\usepackage{siunitx}

\usepackage{tikz}
\usepackage{tkz-euclide}
\usepackage{pgf}
\usepackage{pgfplots, pgfplotstable}
\usepackage{ifthen}
\usepackage{xifthen}
\usepackage{animate}
\usetikzlibrary{arrows,shapes,chains,calc,fit,matrix,positioning,scopes,decorations.pathmorphing,decorations.markings,decorations.pathreplacing,shadows,graphs,intersections,shapes.multipart,positioning,pgfplots.units,backgrounds,shapes.symbols}
\usepackage[noend]{algpseudocode}
\usepackage{algorithm,eqparbox,array}		

\usepackage{lineno}
\modulolinenumbers[5]

\begin{document}

\begin{frontmatter}

\title{Long-Range Trajectories from Global and Local Motion Representations}


\author[inesc_address,feup_address]{Eduardo M. Pereira\corref{correspondingauthor}}
\cortext[correspondingauthor]{Corresponding author. Tel.: +351 222094306}
\ead{ejmp@inesctec.pt}

\author[inesc_address,feup_address]{Jaime S. Cardoso}
\ead{jaime.cardoso@inesctec.pt}

\author[inesc_address,feup_address]{Ricardo Morla}
\ead{ricardo.morla@fe.up.pt}

\address[inesc_address]{INESC TEC, Campus da FEUP, Rua Dr. Roberto Frias, 4200 - 465 Porto, Portugal} 
\address[feup_address]{Faculdade de Engenharia, Universidade do Porto, Rua Dr. Roberto Frias, s/n, 4200 - 465 Porto, Portugal}

\begin{abstract}\label{sec:abstract}
Motion is a fundamental cue for scene analysis and human activity understanding in videos. It can be encoded in trajectories for tracking objects and for action recognition, or in form of flow to address behaviour analysis in crowded scenes. Each approach can only be applied on limited scenarios. We propose a motion-based system that represents the spatial and temporal features of the flow in terms of long-range trajectories. The novelty resides on the system formulation, its generic approach to handle scene variability and motion variations, motion integration from local and global representations, and the resulting long-range trajectories that overcome trajectory-based approach problems. We report the results and conclusions that state its pertinence on different scenarios, comparing and correlating the extracted trajectories of individual pedestrians, manually annotated. We also propose an evaluation framework and stress the diverse system characteristics that can be used for human activity tasks, namely on motion segmentation.
\end{abstract}


\end{frontmatter}


\section{Introduction} \label{sec:introduction}
Nowadays, almost any public space has a CCTV (Closed Circuit Television) system installed. This has fostered the implementation of (semi-)automatic systems to interpret real-world scenes by monitoring individuals and their activities, detecting common motion patterns, and identifying unusual behaviours. The key insight for this type of system is to exploit spatiotemporal relationships among subjects and motion patterns, while retaining structural information of the scene. Underlying motion representations such as trajectories are, by nature, intuitive and useful to build up solutions for such problems. To avoid ambiguity with related work and clarify our aim, throughout the paper the term \textbf{global motion trajectories} refers to trajectories that translate typical paths of pedestrians, normally used for video scene analysis \cite{DBLP:journals/ivc/JohnsonH96,conf/eccv/WangTG06,Wang:2011:TAS:2036798.2036814}.

Long-duration trajectories offer several advantages \cite{conf/icmcs/SunMYC10} over short-range tracklets \cite{Raptis:2010:TDA:1886063.1886107,Wang11,Matikainen09trajectons:action} for visual analysis tasks such as activity discovery and learning of semantic region models for event recognition. However, they imply to overcome occlusions, camera motions, and nonrigid deformations issues. In this paper, we provide a system for computing long-range global motion trajectories based on the combination of a rich global flow representation, that accurately captures spatiotemporal continuity and transitions of motion, with a local flow energy field in terms of entropy, which measures the degree of motion variability revealing salient regions for flow topology analysis. 

The type of scenarios addressed in this work are inherent to public spaces under CCTV systems. We aim to prove that the proposed system can approximate global motion trajectories for different pedestrian scenarios such as crowds, with structured and unstructured movement, and multi-tracking, with sparse and dense groups. Our contributions can be summarised as:
\begin{inparaenum}[i)]
\item integration of local motion information, based on information theory principles, with global motion information, based on temporal integration of flow, to capture longer spatial and temporal changes in the scene; where i.i) a new global technique for flow vector outliers removal is proposed; i.ii) and a fine-to-coarse quantisation of flow vectors and analysis of its impact on temporal flow integration is presented;
\item a complete and dynamic motion-based system that automatically extract long-range motion trajectories from different scenarios, that accounts with ii.i) a re-correlation algorithm to link broken streamlines and accurately form long-range streamlines which correspond to the global motion trajectories.
\end{inparaenum}

The paper's outline is as follows. In Section \ref{sec:related_work}, we survey the related work. Next, in Section \ref{sec:preliminary_concepts}, we briefly present the relevant theoretical concepts behind our system. The following Section \ref{sec:system_overview}, summarises an overview about the system's foundations. A description about the main system steps is presented next, in Section \ref{sec:system_details}. The experimental setup and results are reported in Section \ref{sec:experimental_results}. Finally, we formulate the conclusions and future work in Section \ref{sec:conclusion}.

\section{Related Work} \label{sec:related_work}
Considering various types of scenes, from dense crowd context to multi-tracking with sparse groups, the literature can be subdivided depending on the scene density and the object size. Indeed, human motion can be described at different levels from micro to macro scale. Each one implies specific motion analysis since their underlying relationships among individuals and space-context behaviour differ. Normally for high density scenes and low object resolution, motion is modelled at a global level and patterns are inferred \cite{Ali2007, conf/icpr/HuAS08a, MehranMooreShah_ECCV10}. On the other side for scenes with a small number of objects, multi-tracking approaches \cite{DBLP:CVPRW/2009/5206771,Zhao:2012:TUM:2403006.2403030} are preferred since they track objects individually and describe motion by their spatial position. The computer vision community has been addressing several research problems related to each scenario independently. 

Crowded scenes present two types of categories, structured and unstructured, depending if the movement of objects are defined by physical constraints or if they move freely in any direction, respectively. Related work focuses on modelling scene structures and on recognising the co-occurrences of crowd behaviours. For instance, the authors of \cite{Ali2007} proposed a framework that implements a Lagrangian Particle Dynamics to advect a grid of particles and use them for motion interpretation in the form of physically and dynamically distinguishable motion segments. This type of approaches overcomes the lack of optical flow in capturing long-range temporal dependencies, and do not suffer from problems faced by object-tracking-based approaches. However, they do not consider spatial changes, cause time delays, and imply high computational effort. 

For structured scenes, motion patterns are the most salient features that help understanding the scene \cite{Zhao:2012:TUM:2403006.2403030}. Some approaches \cite{DBLP:conf/cvpr/KimG09,DBLP:CVPRW/2009/5206771} consider the division of the video in spatiotemporal cuboids to identify prototypical motion pattern representations and variations within each one. It is not usual to extract motion trajectories from this type of conditions. However, to the best of our knowledge, there are some works \cite{DBLP:conf/eccv/AliS08,DBLP:conf/icpr/HuAS08,Ozturk:2010:DDM:1904935.1905400} that try to approximate the extraction of motion patterns in terms of \emph{super tracks}, but they did not measure their similarity with traditional object tracks and they did not test their approach on low density scenes.

For unstructured scenes, the concept of \emph{coherent motion} emerges. It describes the free collective movement of individuals in groups and try to infer collective behaviours. This type of approach models the crowd dynamics focusing on individuals and interactions among them \cite{DBLP:conf/eccv/ZhouTW12}. These approaches can follow two types of taxonomy: 
\begin{inparaenum}[i)]
\item macroscopic studies \cite{DBLP:conf/cvpr/ZhouWT12}, that consider groups as a collective and homogeneous block where the individual is transformed by the group;
\item microscopic approaches \cite{mehran09socialforce}, which analyse groups as the composition of individual agents that interact with each other and with the environment.
\end{inparaenum}
Macro models have statistical meaning and not physical, while micro present physical validation but are difficult to scale up to macro scale.

Scenarios with sparse and dense groups follow a single or multi-tracking approaches. Both present difficulties related to target's size, number of similar objects, and occlusions \cite{Zhao:2012:TUM:2403006.2403030}. For crowded scenes, tracking-based models disregard the correlation between pedestrians in a close vicinity. Motion trajectory mechanisms can also be performed at feature level \cite{Wang11,Matikainen09trajectons:action}, instead of object level, by tracking interest points. However, they face critical factors to solve: selection of good tracking features, correct mapping between selected features and actions of interest, trajectory discontinuity due to inconsistent point correspondence, among others. 

Our work borrows concepts from fluid dynamics to integrate macroscopic behaviour, in terms of global flow dynamics, with microscopic approach, in terms of local information theory concepts. From a technical point of view, the proposed system is novel since it integrates, in a pertinent way, different concepts into the pipeline to extract long-range trajectories that approximate the global motion trajectories. From a practical perspective, its pertinence surpass most common methods which can only be applied to limited scenarios. 

\section{Preliminary Concepts} \label{sec:preliminary_concepts}

\subsection{Flow Dynamics} \label{subsec:flow_dynamics}
Motion can be described by Lagrangian and Eulerian flow descriptions, which are formulated on different frames of reference and describe coherent structures of temporal dynamics in terms of trajectories. The Lagrangian coordinate system implies the advection and tracking of particles injected into the flow, and permits the observation of how the flow deforms and rotates the fluid. The Eulerian approach extracts a dense flow coverage since particles are computed at fixed positions, providing an overview over the entire flow at a specific time-step. 

In a time dependent vector field there are four types of characteristic curves: streamlines, pathlines, streaklines, and timelines. Streamlines and pathlines are described as curves tangent to the vector field. Streaklines can be computed from the spatial and temporal gradients of the flow map. For unsteady flows, directions of flow depends on time as well as on position therefore streamline, pathline, and streakline representations are different \cite{Jobard:2001:LAU:601671.601678}. In this work, we explore the streaklines and streamlines complementary representations.

\subsubsection{Vector Field Representation and Advection} \label{subsubsec:vector_field_advection}
A grid of particles is overlaid on the flow field. The scene's motion is quantified by particles' movement driven by dense optical flow. This advection process considers a video represented by a 3-dimensional array $W \times H \times T$, where $T$ is the number of frames, $W$ frame's width, and $H$ frame's height, and by the corresponding optical flow $(u_{w}(t), v_{h}(t))$, where $w \in [1, W]$, $h \in [1, H]$, and $t \in [1, T-1]$. The particle position $(x_{w}(t), y_{h}(t))$ at grid point $(w, h)$ at time $t$ is achieved by solving

\begin{gather}\label{eq:particle_integration} 
\begin{aligned}
x_{w}(t+1) &= x_{w}(t) + u(x_{w}(t), y_{h}(t), t) \\
y_{h}(t+1) &= y_{h}(t) + v(x_{w}(t), y_{h}(t), t)
\end{aligned}
\end{gather}

\noindent The repetition of this process at each frame yields a family of curves that represents the particle trajectory set. Since human motion  creates unsteady flow, each point can be represented by a set of pathlines, streaklines, and streamlines. 

\subsubsection{Streaklines: Computation and Derived Information} \label{subsubsec:streaklines_computation_information}
Streaklines are the locus of points that connect all the particles that had been originated from the same initial point in the past at a given time. Streaklines should not get too long due to shape inconsistency with the flow and instability on numerical integration solution. The authors of \cite{MehranMooreShah_ECCV10} revealed that streaklines are the most informative flow representation when compared with optical flow and particle flow. \emph{Streak Flow} can be obtained from time integration of the velocity field. Such representation fills the gaps of optical flow and captures faster immediate dynamic flow changes than traditional particle flow representation \cite{MehranMooreShah_ECCV10}.

\subsubsection{Streamlines: Computation and Derived Information} \label{subsec:streamlines_computation_information}
Streamlines can be obtained by bidirectional numerical integration of vector field using an autonomous ODE (Ordinary Differential Equation) system. They can be described as curves tangent to the vector field at every point in the flow \cite{10.1109/TVCG.2010.198}. The integration starts from a seed point and ends when it: reaches another streamline's neighbour or a critical point, hits the domain boundary or forms a closed path. Several streamline placement algorithms have been proposed on the literature including flow topology based methods \cite{Verma:2000:FSS:375213.383346}, evenly-spaced streamline placement method \cite{Jobard-1997-CES}, and a hybrid flow topology-evenly-spaced streamline algorithm \cite{Wu:2010:TES:1850479.1850497}. All of them share three common stages: 
\begin{inparaenum}[i)]
\item seed placement, 
\item diffusion process, 
\item stopping criteria.
\end{inparaenum}

\section{System Overview} \label{sec:system_overview}
\begin{figure}[h!]
\centering
\includegraphics[width=1.\textwidth]{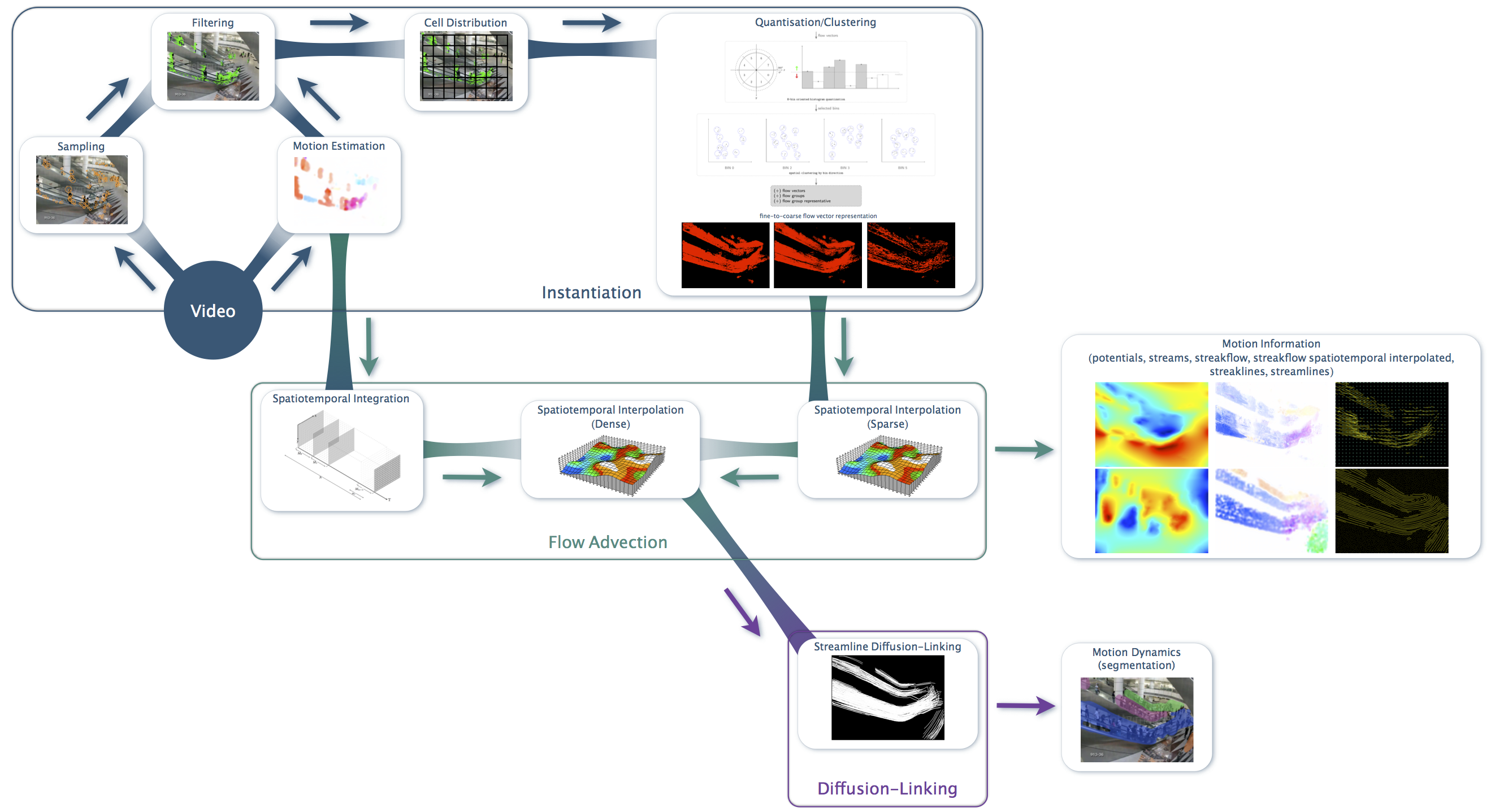}
\caption[System workflow.]{System workflow.}
\label{fig:system_flow_diagram}
\end{figure}

Traditional approaches for motion analysis that consist of detecting moving objects or features, tracking them, and analysing their tracks, miss the estimation of long-term motion representations that bring important cues for scene understanding \cite{lezama11}. Our approach is able to extract long-range motion trajectories that encode spatial and temporal changes in the scene, as well as local motion statistics around each trajectory point in the form of discrete distributions. It follows a Lagrangian perspective to integrate motion through temporal domain under an Eulerian view, similar to the \emph{extended particle} technique defined in \cite{MehranMooreShah_ECCV10}. Figure \ref{fig:system_flow_diagram} shows the overall system workflow.

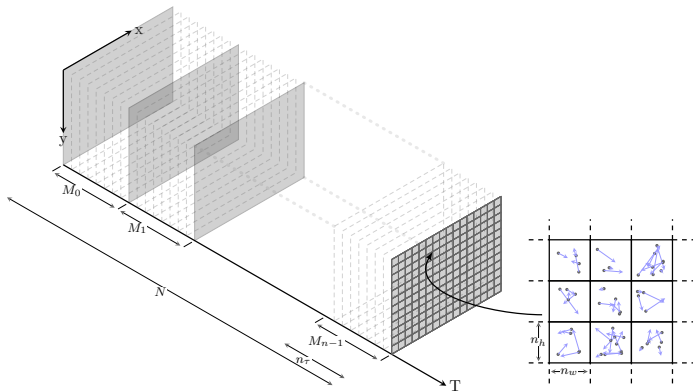
\begin{figure}[h!]
\centering
\scalebox{0.6}{

\def\scalefactor{0.7}
\def\vertexposx{2}
\def\vertexnegx{-2}
\def\gridinc{0.25}
\def\vertexposy{1.5}
\def\vertexnegy{-1.5}

\pgfmathtruncatemacro{\nframes}{8}
\pgfmathsetmacro{\incframe}{0.3}
\pgfmathtruncatemacro{\minibatches}{4}

\begin{tikzpicture}[
	scale=1.0,
	>=stealth, 
	inner sep=0pt, outer sep=2pt,
	axis/.style={thick,->},
	framedots/.style={thick, loosely dotted,line width=1.9pt, line cap=round,draw=black!60!white, opacity=0.15},
	batchframemini/.style={fill=black!60!white, opacity=0.3},
	batchframedots/.style={thin, <->,loosely dotted,draw=black!60!white},
	framein/.style={fill=white, draw=black!60!white, opacity=0.3, fill opacity=0., thin},
	tick/.style={fill=black!60!white,thin,-},
	gapminibatch/.style={fill=black!60!white,thin,<->},
	tickminibatchlabel/.style={fill=white, font=\sffamily\footnotesize},
	tickinlabel/.style={fill=white, font=\scriptsize}]

	\pgfmathsetmacro{\finalpos}{0}
	
	\begin{scope}[
		x={(0.866cm,-0.5cm)}, y={(0.866cm,0.5cm)}, z={(0cm,1cm)},
		scale=\scalefactor,
		grid/.style={thin,draw=black!60!white,opacity=0.3}] 
	
		\pgfmathsetmacro{\tickinitial}{0.1}
		\pgfmathsetmacro{\tickfinal}{0.2}
		\pgfmathsetmacro{\gapminibatch}{\nframes*\incframe}
		
	    	\coordinate (O) at (0, 0, 0);
		\coordinate (Otime) at (0, \vertexnegx, \vertexnegy);
		\coordinate (Oframe) at (0, \vertexnegx, \vertexposy);
	    	\draw[axis] (Otime) -- +(14, 0, 0) node [right] {T};
		\draw[axis] (Oframe) -- +(0, 2.5, 0) node [right] {x};
		\draw[axis] (Oframe) -- +(0, 0,  -2) node [below] {y};
		
		\coordinate (M0topleft) at (2*\nframes*\incframe, \vertexnegx, \vertexposy);
		\coordinate (M1topleft) at (3*\nframes*\incframe, \vertexnegx, \vertexposy+0.5);
		\draw[framedots] (M0topleft) -- +(M1topleft);
		\coordinate (M0topright) at (2*\nframes*\incframe, \vertexposx, \vertexposy);
		\coordinate (M1topright) at (1.3*\nframes*\incframe, \vertexposx, \vertexposy-3.5);
		\draw[framedots] (M0topright) -- +(M1topright);	
		\coordinate (M0bottomright) at (2*\nframes*\incframe, \vertexposx, \vertexnegy);
		\coordinate (M1bottomright) at (1.3*\nframes*\incframe, \vertexposx, \vertexnegy-0.5);
		\draw[framedots] (M0bottomright) -- +(M1bottomright);
				
		\draw[batchframemini] 
			(0, \vertexnegx, \vertexnegy) -- 
			(0, \vertexnegx, \vertexposy) -- 
			(0, \vertexposx, \vertexposy) -- 
			(0, \vertexposx, \vertexnegy) -- 
			cycle;
		\draw[tick] (0, \vertexnegx-\tickinitial, \vertexnegy) -- (0, \vertexnegx-2*\tickfinal, \vertexnegy);
		
		\foreach \minibatch in {0,1,\minibatches}
		{
			 \foreach \frame in {1,...,\nframes} 
			 {
			 	\pgfmathsetmacro{\framecurr}{\nframes*\minibatch+\frame}
				\pgfmathsetmacro{\posframe}{\framecurr*\incframe}
							 
				\ifthenelse{\equal{\frame}{\nframes}}
				{
					\draw[batchframemini] 
						(\posframe, \vertexnegx, \vertexnegy) -- 
						(\posframe, \vertexnegx, \vertexposy) -- 
						(\posframe, \vertexposx, \vertexposy) -- 
						(\posframe, \vertexposx, \vertexnegy) -- 
						cycle;
						
					\draw[tick] (\posframe, \vertexnegx-\tickinitial, \vertexnegy) -- (\posframe, \vertexnegx-2*\tickfinal, \vertexnegy);
					\ifthenelse{\equal{\minibatch}{\minibatches}}
					{
						\draw[tick] (\posframe-\gapminibatch, \vertexnegx-\tickinitial, \vertexnegy) -- (\posframe-\gapminibatch, \vertexnegx-2*\tickfinal, \vertexnegy);
						\draw[gapminibatch] (\posframe-\gapminibatch+\tickinitial, \vertexnegx-2.1*\tickfinal, \vertexnegy) -- (\posframe-\tickinitial, \vertexnegx-2.1*\tickfinal, \vertexnegy) node[tickminibatchlabel,left,midway] {$M_{n-1}$};
						\draw[gapminibatch] (\posframe-\gapminibatch+\tickinitial, \vertexnegx-8.1*\tickfinal, \vertexnegy) -- (\posframe-\tickinitial, \vertexnegx-8.1*\tickfinal, \vertexnegy) node[tickminibatchlabel,left,midway] {$n_{\tau}$};
						\draw[gapminibatch] (0, \vertexnegx-10.3*\tickfinal, \vertexnegy) -- (\posframe-\tickinitial, \vertexnegx-10.3*\tickfinal, \vertexnegy) node[tickminibatchlabel,left,midway] {$N$};
						
						\pgfmathsetmacro{\xincinit}{\vertexnegx+\gridinc}
						\pgfmathsetmacro{\xincfinal}{\vertexposx-\gridinc}
						\pgfmathsetmacro{\yincinit}{\vertexnegy+\gridinc}
						\pgfmathsetmacro{\yincfinal}{\vertexposy-\gridinc}
						
						\foreach \x in {\vertexnegx,\xincinit,...,\xincfinal,\vertexposx}
						{
							\foreach \y in {\vertexnegy,\yincinit,...,\yincfinal,\vertexposy}
							{
								\draw[grid] (\posframe,\x,\vertexnegy) -- (\posframe,\x,\vertexposy);
								\draw[grid] (\posframe,\vertexnegx,\y) -- (\posframe,\vertexposx,\y);
							}
						}
						
						\draw[-latex,thick] (\posframe+3,0.5) node[right]{} to [out=180,in=-120] (\posframe-1,0.5);
						\pgfmathparse{\posframe+2cm}
						\pgfmathsetmacro{\finalpos}{\pgfmathresult}
					}
					{
						\draw[gapminibatch] (\posframe-\gapminibatch+\tickinitial, \vertexnegx-2.1*\tickfinal, \vertexnegy) -- (\posframe-\tickinitial, \vertexnegx-2.1*\tickfinal, \vertexnegy) node[tickminibatchlabel,left,midway] {$M_{\minibatch}$};						
					}
				}
				{
					\draw[framein, densely dashed] 
						(\posframe, \vertexnegx, \vertexnegy) -- 
						(\posframe, \vertexnegx, \vertexposy) -- 
						(\posframe, \vertexposx, \vertexposy) -- 
						(\posframe, \vertexposx, \vertexnegy) -- 
						cycle;
						
				}
			 }
		}
		
	\end{scope}
	
	\begin{scope}[
		scale=\scalefactor*0.65,
		shift={(16cm,-14.5cm)}]
		
		\foreach \x in {5,7,9} 
		{
			\foreach \y in {1,3,5}
			{
				\node at (\x+0.5, \y+1.0){\scalebox{0.7}{

\def\minpos{0.0}
\def\maxpos{1.0}
\def\minvectors{3}
\def\maxvectors{10}

\def\vectorincolor{blue!40!white}

\pgfmathsetmacro{\vectorradius}{0.05}

\begin{tikzpicture}	

	\pgfmathrandominteger{\nvectors}{\minvectors}{\maxvectors} 
	
	\foreach \x in {1,...,\nvectors}
	{
		\pgfmathparse{rnd}
		\pgfmathsetmacro{\pax}{\pgfmathresult}
		
		\pgfmathparse{rnd}
		\pgfmathsetmacro{\pay}{\pgfmathresult}
		
		\fill[draw=black!60!white, fill=black!60!white] (\pax,\pay) circle[radius=1pt];
		
		\pgfmathparse{rnd}
		\pgfmathsetmacro{\pbx}{\pgfmathresult}
		
		\pgfmathparse{rnd}
		\pgfmathsetmacro{\pby}{\pgfmathresult}
		
		\draw[->,thick,draw=\vectorincolor, fill=\vectorincolor]  (\pax,\pay) -- (\pbx,\pby);
		
	} 

\end{tikzpicture}}};
        				\path[draw,thick] (\x,\y) rectangle (\x+2,\y+2);
   			}
		}
		
		\draw[gapminibatch] (4.5, 0, \vertexnegy) -- (6.3, 0, \vertexnegy) node[tickminibatchlabel,midway] {$n_{w}$};
		\draw[gapminibatch] (3.9, 0.5, \vertexnegy) -- (3.9, 2.4, \vertexnegy) node[tickminibatchlabel,midway] {$n_{h}$};
		
		\foreach \x in {5,7,9,11}
		{
			\path[draw,dashed,thick] (\x,0) -- (\x,1) (\x,7) -- (\x,8);
			\path[draw,dashed,thick] (4,\x-4) -- (5,\x-4) (11,\x-4) -- (12,\x-4);			
		}
	\end{scope}
		
\end{tikzpicture}}
\caption[Spatiotemporal video volume representation.]{Spatiotemporal video volume representation.}
\label{fig:video_frame_representation}
\end{figure}

The system subdivides temporally the video in a series of mini-batches without overlap, and subdivides spatially each frame in a static grid. In this way, spatiotemporal cells are created (see Figure \ref{fig:video_frame_representation}). On each frame, the flow vectors are formed from the combination of a sparse sampling strategy with the flow map obtained from an optical flow algorithm. The locations of the sampled key points are used as starting points and a median filtering is applied to the current flow map to obtain approximated and smoothed ending points for each flow vector. All flow vectors are collected by the corresponding enclosing cell for further quantisation and clustering, based on the orientation and spatial information respectively. Such step is performed on each cell at the end of each mini-batch. This dual-operation aims to reduce the number of flow vectors and leads to a fine-to-coarse representation, which defines three levels of granularity:
\begin{inparaenum}[i)]
\item \emph{flow vectors}, all vectors collected during the mini-batch duration;
\item \emph{flow groups}, dominant groups of vectors obtained after quantisation and clustering;
\item \emph{flow group representative}, the most dominant group, which is the one with the largest number of flow vectors within.
\end{inparaenum}
Flow information of local neighborhood around each cell is acquired and continuously updated. This information permits a richer distribution to accurately infer entropy and energy measures, which help to obtain vector field's characteristics such as topology.

From the advection scheme we can extract streaklines initiated at each particle position. Such process should not be too long to avoid the propagation of flow changes that are inconsistent with the actual flow, which can also be caused by the optical flow algorithms that introduce noisy flow vectors. Human-like motion does not produce so well-defined properties such as vorticity. Therefore, to approximate human-motion scenarios to fluid motion and reduce propagation artifacts, the advection is executed between mini-batches and not between individual frames. This is one of the novel features that distinguish our system from the approach presented in \cite{MehranMooreShah_ECCV10}.
 
Since pedestrian flow is unsteady, streaklines and streamlines curves are different in direction and shape. At the end of a set of consecutive mini-batches, so-called \emph{memory cell} size, an averaged streak flow map is combined with one of the fine-to-coarse flow representations to obtain a dense vector field, which is used as input for the streamline diffusion process. However, such discretisation could be insufficient to avoid broken streamlines. Instead of investigating about diffusion techniques of streamlines and improve them, we optimise globally the local short-range streamlines and obtain global long-range trajectories, which accurately correlate features related to flow decomposition and to the motion model, and can handle local ambiguities by spatiotemporal regularisation.

Our system, due to its novel formulation, is able to extract meaningful and correct long-range trajectories that represent either object tracks and motion patterns for different scenarios such as crowds, with structured and unstructured movement, and multi-tracking, with sparse and dense groups. Correct system initialisation and parametrisation of its different steps for the aforementioned scenarios are studied and evaluated in this work.

\section{System Details} \label{sec:system_details}
The input of the system is a monocular video sequence $I$ of $T$ frames, where its volume is $W \times H$ (pixels) $\times ~T$ (frames). The volume is subdivided into $N$ spatiotemporal cells of size $n_{w} \times n_{h} \times n_{\tau}$, without overlap, where $n_{w}$ is the width of the cell in pixels, $n_{h}$ is the height of the cell in pixels, and $n_{\tau}$ the number of frames in the cell, which is equal to the temporal duration of each mini-batch (Figure \ref{fig:video_frame_representation}). The video is composed by a set of mini-batches, $\mathcal M = \lbrace M_{n}\rbrace$. The output primitive is a trajectory $\Gamma = (\mathbf{S}(n), \mathbf{x}(n))$, where $\mathbf{S}(n)$ is the trajectory descriptor, and $\mathbf{x}(n) = (x(n), y(n))$ are the trajectory spatial coordinates at mini-batch $n$. The temporal coordinate, $n$, is integral (correspond to mini-batches) and the spatial coordinates, $\mathbf{x}(n)$, are in sub-pixel accuracy. The set of detected trajectories is denoted by $\mathcal T = \lbrace\Gamma_{i}\rbrace$. The Algorithm \ref{code:main_loop} summarizes the relevant steps of the system.

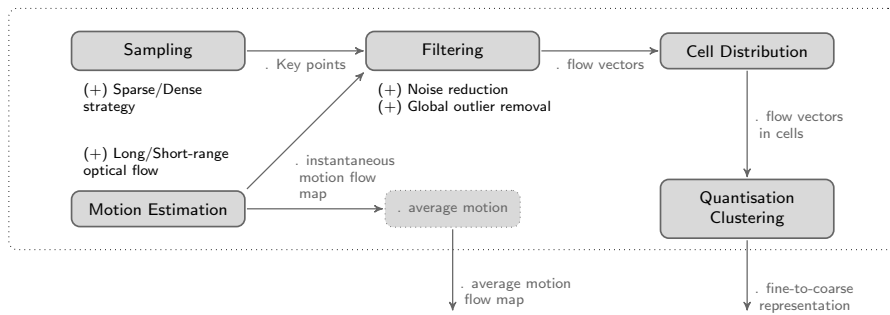
\begin{figure}[h!]
\centering
\resizebox{14.5cm}{!}{
\def\scalefactor{1.0}
\def\yshiftlabel{0.}

\begin{tikzpicture}[
	  scale=\scalefactor,
	  transform shape,
	  font=\sffamily,
	  every matrix/.style={ampersand replacement=\&,column sep=1cm,row sep=0.6cm, outer sep=1pt},
	  inputarrow/.style={anchor=east,single arrow, draw,line width=1pt, minimum height=1.5mm, minimum width=1.0cm, single arrow head extend=1.2mm, single arrow tip angle=120, draw=black!60!white,fill=white,font=\sffamily\footnotesize},
	  source/.style={draw,thick,rounded corners,fill=yellow!20,inner sep=.2cm},
	  process/.style={draw,thick,circle,fill=blue!20},
	  transforminput/.style={draw=black!60!white,dotted,rounded corners,inner sep=.2cm,fill=white,fill opacity=0.25,align=left,font=\sffamily\scriptsize,black!60!white,text opacity=1.0}, 
	  terminal/.style={draw=black!60!white,thick,rounded corners,inner sep=.2cm,fill=black!60!white,fill opacity=0.25,align=center,font=\sffamily\footnotesize,text opacity=1.0,text width=2.5cm*\scalefactor}, 
	  common_inner/.style={rectangle, draw, dotted,inner ysep=10pt,inner xsep=1pt,rounded corners}, 
	  comment/.style={rectangle, inner sep= 5pt, text width=2.5cm, node distance=0.25cm, align=left,font=\sffamily\scriptsize},
	  sink/.style={source,fill=green!20},
	  dots/.style={gray,scale=2},
	  edge/.style={semithick,font=\sffamily\scriptsize,black!60!white},
	  to/.style={->,>=stealth',shorten >=0pt,edge},
	  every node/.style={align=center},
	  skip loop/.style={to path={-- ++(0,#1) -| (\tikztotarget)}, rounded corners},
	  vh path/.style={to path={-| (\tikztotarget)}, rounded corners,font=\sffamily\footnotesize}]

  	\matrix (geral_matrix) [
		scale=\scalefactor,
		transform shape]
	{
		
		\& \node[terminal] (sampling) {Sampling}; \& \& \node[terminal] (filtering) {Filtering}; \& \& \node[terminal] (spatial_grid) {Cell Distribution}; \& \\ 
		
		\& \& \& \& \& \& \& \& \& \\
		\& \& \& \& \& \& \& \& \& \\
		
		\coordinate (A4_1) {}; \& \node[terminal] (optical_flow) {Motion Estimation}; \& \& \node[transforminput,,align=center] (average_optical_flow) {. average motion}; \& \& \node[terminal] (coarse_fine_quantization) {Quantisation\\Clustering}; \& \coordinate (A4_6) {}; \\	
		
		
		\& \& \& \& \& \& \& \& \& \\
		
		\& \& \& \coordinate (A4_3) {}; \& \& \coordinate (A4_5) {}; \& \\					
	};

	\node[common_inner, fit=(sampling) (optical_flow) (filtering) (spatial_grid) (A4_1) (A4_6)] (container_frame) {};
	

	\node [comment, below=\yshiftlabel of sampling] (comment-sampling) {(+) Sparse/Dense strategy};
	\node [comment, above=\yshiftlabel of optical_flow] (comment-optical_flow) {(+) Long/Short-range optical flow};
	\node [comment, below=\yshiftlabel of filtering, text width=3.5cm,  xshift=0.5cm] (comment-filtering) {(+) Noise reduction\\(+) Global outlier removal};		
			 
	\draw[to] (sampling) -- node[midway,below] {. Key points} (filtering);
	\draw[to] (optical_flow.north east) -- node[] {} (filtering.south west);
	\draw[to] (optical_flow) -- node[midway, above,align=left,xshift=0.5cm] {. instantaneous\\ \;motion flow\\ \;map} (average_optical_flow);
	\draw[to] (average_optical_flow) -- node[below right,align=left] {. average motion\\ \;flow map} (A4_3);
	\draw[to] (filtering) -- node[midway,below,align=left] {. flow vectors} (spatial_grid);
	\draw[to] (spatial_grid) -- node[right,align=left] {. flow vectors\\ \;in cells} (coarse_fine_quantization);
	\draw[to] (coarse_fine_quantization) -- node[below right,align=left] {. fine-to-coarse\\ \;representation} (A4_5);			
	
	\node [font=\scriptsize\ttfamily,text width=2.5cm,align=left] at (container_frame.north west) [above right] {};
      
\end{tikzpicture}}
\caption[System Instantiation step.]{System Instantiation step.}
\label{fig:system_instantiation_step}
\end{figure}

\subsection{Instantiation} \label{subsec:instantiation}
The beginning of this stage is composed by the sampling strategy and by the motion estimation represented by flow maps, which are indeed instantaneous velocities. Both information are combined to create the flow vectors. After, they undergo a filtering step to remove noise and outliers, and then are distributed by the enclosing cells, where are locally quantised and grouped. These steps form the \emph{instantiation} block of the system and are executed at each frame, excepting the quantisation and clustering operation. Its output is: 
\begin{inparaenum}[i)]
\item an averaged flow map from the instantaneous flow maps of the current mini-batch;
\item a global fine-to-coarse flow vector representation (see Figure \ref{fig:system_instantiation_step}).
\end{inparaenum}
\noindent All the operations taken in this stage are executed within each mini-batch.


%

\newcommand{\tikzmark}[1]{\tikz[overlay,remember picture] \node (#1) {};}

\renewcommand\algorithmiccomment[1]{\hfill \#\ \eqparbox{COMMENT}{\color{black!60!white}\scriptsize\textit{#1}}}
\algnewcommand{\LineComment}[1]{\State \#\ {\color{black!60!white}\scriptsize\textit{#1}}}

\newcommand*{\AddNote}[4]
{
	\begin{tikzpicture}[overlay, remember picture]
		\draw [decoration={brace,amplitude=0.5em,mirror},decorate,ultra thin,black!60!white]
			($(#3)!(#1.north)!($(#3.east)+(1,1)$)$) --  ($(#3)!(#2.south)!($(#3.east)-(0,1)$)$)
			node [align=left, text width=0.05cm, pos=0.1, anchor=east,xshift=-6pt] {#4};
	\end{tikzpicture}
}

\algnewcommand{\algorithmicgoto}{\textbf{go to}}
\algnewcommand{\Goto}[1]{\algorithmicgoto~\ref{#1}}

\begin{algorithm*}[t]
	\footnotesize
	\caption{Main Loop algorithm}\label{code:main_loop}
	\begin{algorithmic}[1]
		\Procedure{Main}{}\tikzmark{left}
			\For{$f\gets 1, T$}
				\LineComment{Computes sparse/dense sampling and returns key points.}
				\State $keyPoints\gets \Call{Sampling}{f}$ 
				\LineComment{Computes instantaneous flow maps from Optical Flow algorithm.}
				\State $instFlowMap\gets \Call{MotionEstimation}{f}$ 
				\LineComment{Filtering step with outlier removal technique.}
				\State $flowVectors\gets \Call{Filter}{keyPoints, instFlowMap}$ 
				\LineComment{Computes an averaged flow map for the current mini-batch.}
				\State $avgFlowMap\gets \Call{ComputeAvgFlowMap}{f}$ 
				\LineComment{Distributes flow vectors through the enclosed cell.}
				\State \Call{DistributeSpatially}{$flowVectors$} 
				\If{$newMinibatch$}
					\LineComment{Quantisation/Clustering for fine-to-coarse flow vector representation.}
					\State $flowRepresentation\gets \Call{QuantiseAndClusterCells}{cells}$
					\LineComment{Computes motion advection.}
					\State \Call{ComputeMotionAdvection}{$avgFlowMapPrev$, $avgFlowMapCurr$} 
					\LineComment{Computes an averaged streak map for the set of mini-batches.}
					\State $avgStreakMap\gets \Call{ComputeAvgStreakMap}{f}$
					\If{$nMinibatches=cellMemory$}
						\LineComment{Extract sparse interpolated flow map from flow vector representation.}
						\State $interpFlowMap\gets \Call{ComputeInterpFlowMap}{flowRepresentations}$ 
						\LineComment{Extract dense interpolated streak flow from combination between averaged}
						\LineComment{streak flow and fine-to-coarse flow vector representation.}
						\State $interpStreakFlowMap\gets \Call{SpatioTemporalInterpolation}{avgStreakMap, flowRepresentation}$
						\LineComment{Compute streamlines from diffusion technique.}
						\State $streamlines\gets \Call{ComputeStreamlines}{interpStreakFlowMap}$ 
						\LineComment{Link broken streamlines to form representative long motion trajectories.}
						\State $trajectories\gets \Call{LinkGloballyStreamlines}{streamlines}$ 
					\EndIf
					\State \textbf{end if}
				\EndIf
				\State \textbf{end if}

			\EndFor 
			\State \textbf{end for}
		\EndProcedure
		\State \textbf{end procedure}
	\end{algorithmic}
\end{algorithm*}

\subsubsection{Sampling and Motion Estimation} \label{subsubsec:flow_vectors_extraction}
The sampling extracts a set of key points from one of two possible distributions: dense or sparse. The dense sampling is highly computational demanding, and such effort is propagated through the subsequent steps. Also, it introduces noisy points that do not add discriminative value, therefore sparse sampling is preferred. The motion flow is estimated using optical flow algorithms, which differ from either frame-to-frame analysis or larger spatiotemporal displacements. We tested two algorithms: the short-classical Farneb\"{a}ck \cite{Farneback:2003:TME:1763974.1764031}, and the large-descriptor matching in variational model (LDOF) \cite{Brox:2011:LDO:1936329.1936562}. Both present good results but we adopted the Farneb\"{a}ck's method due to the trade-off between computational effort and robustness. See \cite{EPereira_flow_trajectories} for a more detailed description about these results.

\subsubsection{Filtering} \label{subsubsec:filtering}
Each flow vector is represented by $F_{i} = (x_{i}, y_{i}, u_{i}, v_{i})$, where $(x_{i}, y_{i})$ is the sampling point, and $(u_{i}, v_{i})$ are the motion field components in $x$ and $y$ directions, respectively. The filtering step builds each vector flow by assuming the key point location to be the initial vector's position, $P_{i}(t) = (x_{i}(t), y_{i}(t))$, and 
considering the flow vector's endpoint, $P_{f}(t) = (x_{f}(t), y_{f}(t))$, as the median (component-wise) of the flow field, $f = (u(t), v(t))$, in a neighborhood of size $K$. Therefore, the number of flow vectors is equal to the number of key points. The adopted method is a median filtering kernel that performs better than the bilinear interpolation \cite{Bro10e}.

A dual-threshold on flow magnitude is applied to remove flow vectors that have little motion information, as well as extremely high magnitudes. However, we empirically verified that such operation is not enough to remove the flow vectors resulting from the background noise. For this reason, a novel outlier removal technique is proposed. It consists on a rough approximation that fits the magnitude of the flow vectors into a unimodal gaussian distribution and estimates both lower and upper bounds inspired on the Chebyshev's theorem and on the skewness measure. 

\begin{figure}[h!]
\centering	
	\subfigure[][]{\includegraphics[width=.45\textwidth]{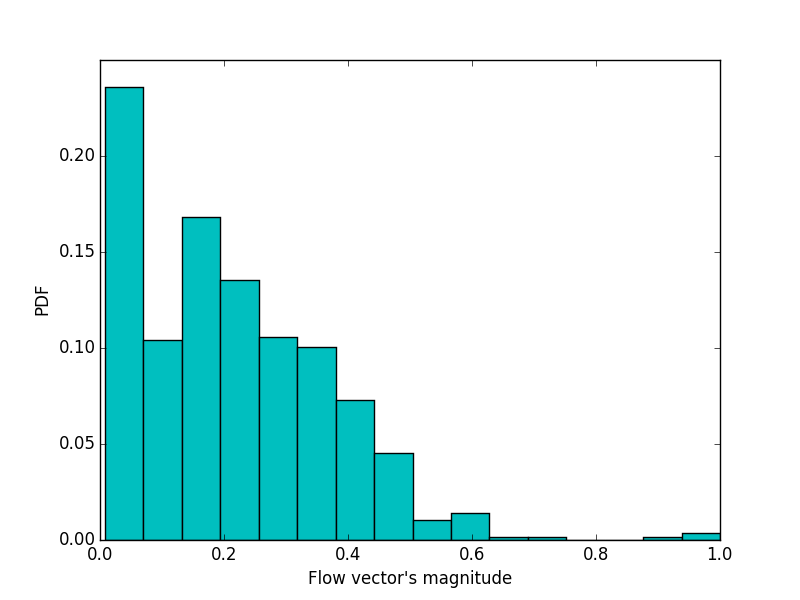}\label{fig:flow_vector_distribution_UMN_first}}
	\subfigure[][]{\includegraphics[width=.45\textwidth]{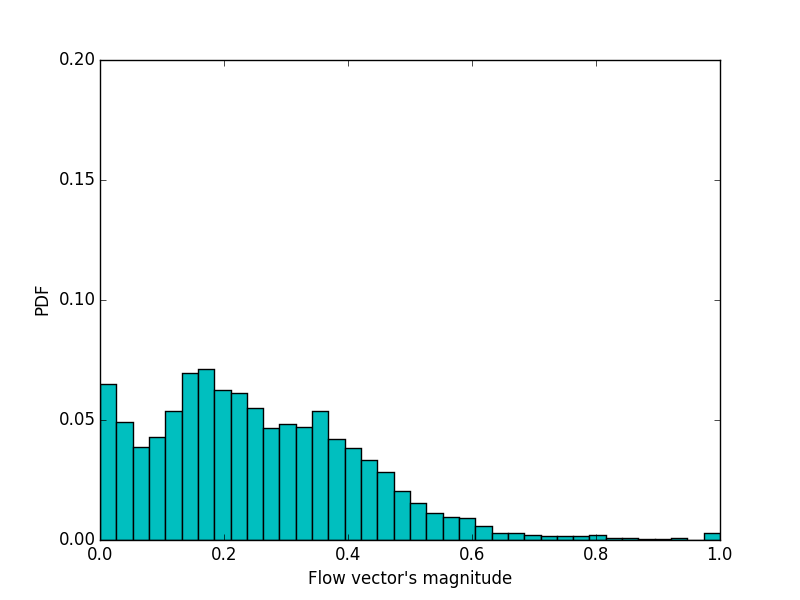}\label{fig:flow_vector_distribution_UMN_all}}	
	\caption[Flow vector's magnitude distribution (C.U. scenario) in:
			\subref{fig:flow_vector_distribution_UMN_first} just one frame;			
			\subref{fig:flow_vector_distribution_UMN_all} several frames.]{Flow vector's magnitude distribution (C.U. scenario) in:
			\subref{fig:flow_vector_distribution_UMN_first} just one frame;			
			\subref{fig:flow_vector_distribution_UMN_all} several frames.} 
	\label{fig:flow_vector_distribution}
\end{figure}

Since the magnitude of the flow vectors depend on the instantaneous velocities and on the kernel size, we expect that most of the flow vector's magnitude to be smaller than the mean magnitude. Therefore the distribution will not be symmetric and will be, normally, skewed to the right, as illustrated by Figure \ref{fig:flow_vector_distribution}.

The Chebyshev's theorem can be applied to any data set regardless of its distribution. In this case, due to the sharp skewed distribution, the Chebyshev's inequality is not good enough to estimate the non-symmetric, lower and upper, bounds, $\lbrace \ell_{-}, \ell_{+}  \rbrace$. Our technique estimates two parameters to obtain those bounds, a shift factor, $\chi$, and a scale factor, $\rho$. The intuition behind is that the nonparametric skew measure gives a shift factor to account with the difference between the median and the mean, accordingly with the distribution's tendency (positive or negative skew), while the ratio between the number of observations represented by the median and by the mode provides a factor to scale the shift amount (the greater the ratio, the smaller the scale factor, and, consequently, the larger will be the acceptance gap between the threshold limits). 

The nonparametric skew measure is given by $\gamma = (\mu - \nu)/\sigma$, which assumes values between $[-1, 1]$. Considering these boundaries, the shift factor is decomposed in the following limits

\begin{align} \label{eq:offset_limits}  
(\chi_{-}, \chi_{+}) & = \left\{
	\begin{array}{rl}
		(\textcolor{white}{\vert} \gamma \textcolor{white}{\rvert}, 0), \qquad \text{if } \gamma > 0 \nonumber \\
		(0, \lvert  \gamma \rvert), \qquad \text{if } \gamma < 0 
	\end{array} \right.  \\	
\end{align}

The scale factor, $\rho = 1 - \tilde{x}/\hat{x}$, results from the relation between the number of observations represented by the median, $\tilde{x}$, and the mode, $\hat{x}$, of the log-transformed flow vector's magnitude distribution. We employ the log-transformation to approximate the data to a symmetric distribution before measuring the amplitude relation. We adopt the Freedman-Diaconis rule to obtain the optimal bin width, $Bin_{w} = 2 \cdot IQR \cdot n^{-\frac{1}{3}}$, where $IQR$ is the interquartile range, and $n$ is the number of observations in distribution. The scale factors are given by $\lbrace s_{-}, s_{+} \rbrace = \lbrace \rho \cdot \chi_{-}, \rho \cdot \chi_{+} \rbrace, \in [0, 1]$, which produces the final non-symmetric bounds, $\lbrace \ell_{-}, \ell_{+} \rbrace = \lbrace \sigma \cdot s_{-}, \sigma \cdot s_{+} \rbrace$, which are used to estimate the final threshold limits for outlier removal, $\lbrace \lambda_{-},  \lambda_{+} \rbrace = \lbrace min + \ell_{-}, max - \ell_{+} \rbrace$, where $min$ and $max$ are the minimum and maximum values of the flow vector's magnitude distribution, respectively.

After several experiments on different datasets, this technique presented coherent and better results to remove global outliers on flow vector's data than state-of-the-art methods (see Section \ref{subsec:experimental_outlier_removal}). 

\subsubsection{Cell Distribution} \label{subsubsec:cell_distribution}
The video volume has a regular spatiotemporal distribution. Spatially each frame is divided by a grid, whose resolution is dependent on the frame size and is set at the beginning. Temporally the video duration is evenly divided. Each spatiotemporal region is denominated a cell, $C_{i}$, and contains the flow vectors whose initial positions lay inside it. Each flow vector is encoded by $F_{i} = (x_{i}, y_{i}, L_{i}, \theta_{i}, t_{i})$, where $(x_{i}, y_{i})$ is the sampling point, $L_{i}$ is the flow magnitude length, $\theta_{i}$ is the flow angle relative to positive \emph{x-axis}, and $t_{i}$ is the frame. This step as well as the previous ones are executed every frame.

\subsubsection{Quantisation and Clustering} \label{subsubsec:quantization_clustering}
\begin{figure}[ht]
\centering
\scalebox{0.75}{

\def\scaleimage{0.8}

\newcommand\nclusters{4}

\begin{tikzpicture} [
	 common_inner/.style={rectangle, draw, dotted,inner sep=1pt,rounded corners},
	 edge/.style={semithick,font=\sffamily\scriptsize,black!60!white},
	 to/.style={->,>=stealth',shorten >=0pt,edge},
	 terminal/.style={draw=black!60!white,thick,rounded corners,inner sep=.2cm,fill=black!60!white,fill opacity=0.25,align=center,font=\sffamily\footnotesize,text opacity=1.0,text width=2.5cm*\scalefactor}, 
	 framedots/.style={thick, loosely dotted,line width=1.9pt, line cap=round,draw=black!60!white, opacity=0.3},
	 every matrix/.style={ampersand replacement=\&,column sep=1cm,row sep=0.5cm, outer sep=1pt}]
	 
	 \matrix [
	 	column 1/.style={column sep=1mm},
		column 3/.style={column sep=1mm}]
	 {
	 	\node[shape=coordinate] (A11) {}; \\
		\node (orientation_quantization) {\scalebox{\scaleimage}{
\def\scaleimage{0.7}

\begin{tikzpicture} [
	 common_inner/.style={rectangle, draw, dotted,inner sep=1pt,rounded corners},
	 edge/.style={semithick,font=\sffamily\scriptsize,black!60!white},
	 to/.style={->,>=stealth',shorten >=0pt,edge}]

	\node (orientation_wheel) at (0, 0) {\scalebox{\scaleimage}{

\pgfmathtruncatemacro{\radius}{2}
\pgfmathsetmacro{\radiusinc}{0.25}
\pgfmathsetmacro{\axisinc}{1.0}

\begin{tikzpicture}
    
       	\draw[->] (-\radius-\axisinc,0) -- (\radius+\axisinc,0) node[right,fill=white] {$x$};
       	\draw[->]  (0,\radius+\axisinc) -- (0,-\radius-\axisinc) node[below,fill=white] {$y$};
        
	\draw[draw=black!60!white, thick] (0,0) circle (\radius);    
	
	\foreach \a in {0, 90,180,270}
		\draw[very thick,->,draw=black!60!white, opacity=0.3] (0, 0) -- (\a:\radius+\radiusinc);
		
	\foreach \a in {45, 135, 225, 315}
		\draw [very thick,->,draw=black!60!white,opacity=0.3] (0, 0) -- (\a:\radius+\radiusinc);
		
	\foreach \a in {0,...,7}
	{
		\pgfmathsetmacro{\posbin}{(\a+1)*45)-(45*0.5)}
		\pgfmathtruncatemacro{\bin}{7-\a}
		\draw (\posbin:1.5) node[fill=white] {$\bin$};
		
		\ifthenelse{\equal{\a}{0}}
		{
			\draw (\a:\radius+\axisinc-\radiusinc) node[right, below, font=\sffamily\footnotesize]{$0^\circ$};
		}
		{}
		\ifthenelse{\equal{\a}{7}}
		{
			\draw (\a:\radius+\axisinc-2.5*\radiusinc) node[right, font=\sffamily\footnotesize]{$360^\circ$};
		}
		{}
	}
		
	\draw[fill=black!60!white] (0,0) circle(0.7mm);
	\draw[,dashed,domain=6.24:0.02,samples=200,smooth, -<] plot (canvas polar cs:angle=\x r,radius={1.2 r});
	

\end{tikzpicture}}};
	\node [right of=orientation_wheel, node distance=7cm, yshift=0.45cm] (orientation_histogram) {\scalebox{\scaleimage}{


\pgfmathtruncatemacro{\nbins}{8}
\pgfmathsetmacro{\binheightmax}{3.7}
\pgfmathsetmacro{\histogrammedian}{1.9}
\pgfmathsetmacro{\binsize}{1.25}
\pgfmathsetmacro{\bininc}{0.0}
\pgfmathsetmacro{\binradius}{0.05}

\def\bincolorin{green}
\def\bincolorout{red}
\def\bincolor{black!60!white}

%

\begin{tikzpicture} [
	framedots/.style={thick, loosely dotted,line width=1.9pt, line cap=round,draw=black!60!white, opacity=0.3}]

	\pgfmathsetmacro{\margin}{0.2}
	\pgfmathsetmacro{\midbinsize}{\binsize*0.5}
	\pgfmathsetmacro{\axisX}{(\nbins+1)*(\binsize+\bininc)}
	
       	\draw[->] (0,0) -- (\axisX,0) node[right,fill=white] {};
       	\draw[->]  (0,0) -- (0,\binheightmax+\binsize+\bininc) node[above,fill=white] {};
	
	\draw[-,framedots] (-\binsize-\bininc,\histogrammedian) -- (\axisX,\histogrammedian) node[fill=white, font=\sffamily\footnotesize,below] {median};
	
	\draw[->, color=red, thick, line width=2pt] (-\midbinsize-\bininc,\histogrammedian-\margin) -- (-\midbinsize-\bininc,\histogrammedian-\margin-0.5);
	\draw[->, color=green, thick, line width=2pt] (-\midbinsize-\bininc,\histogrammedian+\margin) -- (-\midbinsize-\bininc,\histogrammedian+\margin+0.5);
	
	\pgfplotsarraynewempty\probe
	\pgfplotsarraypushback eins\to\probe
	\pgfplotsarraypushback [probe]\to\probe
	\pgfplotsarraypushback [probe2]\to\probe
	\pgfplotsarraypushback [probe3]\to\probe
	
	\pgfplotsarrayforeach\probe\as\curarrayelem{\curarrayelem \par}
	
	\pgfplotsarraynew\fooarray{Eins\\Zwei\\Drei\\}
	\pgfplotsarrayforeach\fooarray\as\foo{Element \foo\par}	
	
	\def\binvalues{2.0,0.8,2.5,3.3,0.3,2.8,1.2,1.6}\def\dim{8}
	
	\foreach \n [count=\bin] in \binvalues
	{
		
		\pgfmathsetmacro{\binheight}{\n}
		\pgfmathsetmacro{\binpos}{(\bin-1)*(\binsize+\bininc)}
		
		\pgfmathparse{ifthenelse(\binheight > \histogrammedian,"1", "0")}
		\let\pass\pgfmathresult;
		
		\ifthenelse{\equal{\pass}{1}}
		{			
			\draw[color=\bincolor, opacity=0.9, fill=\bincolor, fill opacity=0.3] (\binpos,0) rectangle (\binpos+\binsize, \binheight);
		}
		{
			\draw[color=\bincolor, opacity=0.9, fill=\bincolor, fill opacity=0.] (\binpos,0) rectangle (\binpos+\binsize, \binheight);
		}				
		
		\pgfmathsetmacro{\binpoint}{\binpos+\midbinsize}
		\draw[draw=\bincolor, fill=\bincolor] (\binpoint,\binheight) circle (\binradius);   
		
		\pgfmathparse{ifthenelse(\bin > 0,"1", "0")}
		\let\pass\pgfmathresult;
		\ifthenelse{\equal{\pass}{1}}
		{
			
		}
		{
		}
	}

\end{tikzpicture}}};
	
	\node[common_inner, fit=(orientation_wheel) (orientation_histogram)] (orientation_quantization) {};
	
	
	\node [font=\scriptsize\ttfamily,text width=5.5cm,align=center] at (orientation_quantization.south) [above] {8-bin oriented histogram quantisation};

\end{tikzpicture}}}; \\
		\node (spatial_clustering) {\scalebox{\scaleimage}{
\def\scaleimage{0.5}

\begin{tikzpicture} [
	 common_inner/.style={rectangle, draw, dotted,inner sep=1pt,rounded corners},
	 edge/.style={semithick,font=\sffamily\scriptsize,black!60!white},
	 to/.style={->,>=stealth',shorten >=0pt,edge},
	 framedots/.style={thick, loosely dotted,line width=1.9pt, line cap=round,draw=black!60!white, opacity=0.3}]

	\def\binindexes{0,2,3,5}
	\pgfmathsetmacro{\clusterdistance}{4.7}

	\foreach \n [count=\cluster] in \binindexes
	{
		\pgfmathsetmacro{\clusterpos}{\cluster*\clusterdistance}
		\node (\cluster) at (\clusterpos, 0) {\scalebox{\scaleimage}{\input{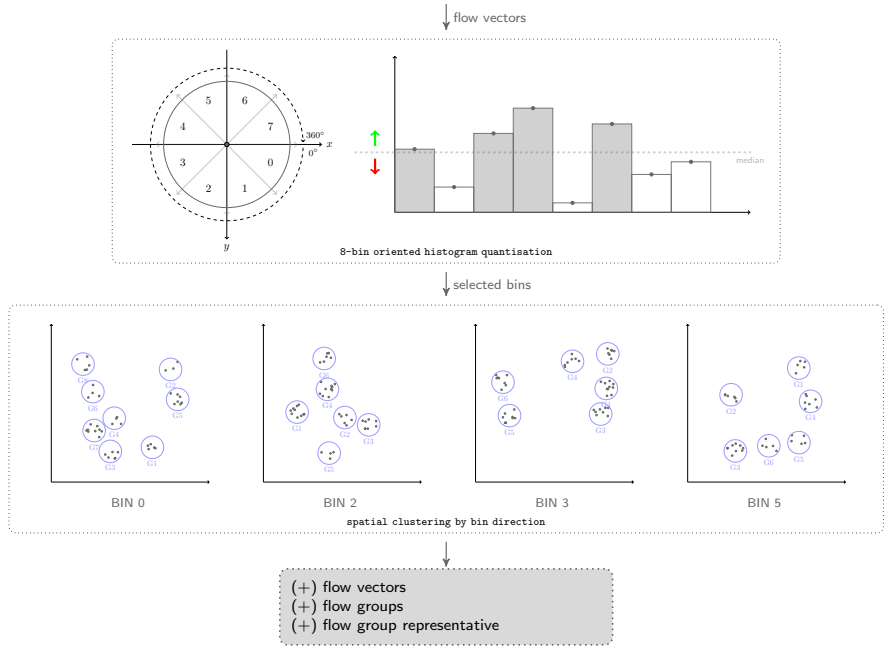}}};
		\node [font=\sffamily\footnotesize,text opacity=1.0,text width=2.5cm,align=center,black!60!white] at (\cluster.south) [below] {BIN \n};
	}
	
	\node[common_inner, fit=(1) (2) (3) (4), inner ysep=15pt, yshift=-0.39cm, inner xsep=14pt] (spatial_clustering) {};
	
	
	
	
	
	
	\node [font=\scriptsize\ttfamily,text width=5.5cm,align=center] at (spatial_clustering.south) [above] {spatial clustering by bin direction};

\end{tikzpicture}}}; \\
		\node[terminal,dotted, text width=5.5cm, align=left] (output) {(+) flow vectors\\(+) flow groups\\(+) flow group representative}; \\
	 };
	 
	 \draw[to] (A11) -- node[midway,right] {flow vectors} (orientation_quantization);
	 \draw[to] (spatial_clustering) -- node[midway,below] {} (output);
	 \draw[to] (orientation_quantization) -- node[midway,right] {selected bins} (spatial_clustering);
	 
\end{tikzpicture}}
\caption[Quantisation and Clustering step by cell.]{Quantisation and Clustering step by cell.}
\label{fig:quantization_clustering_cell}
\end{figure}

In order to obtain a fine-to-coarse flow vector representation that could permit to model different levels of patterns, a two-step quantisation and clustering approach is applied on each cell at the end of each mini-batch. This operation considers all the flow vectors collected along the duration of the mini-batch, therefore the number of key points is much greater than the number of cells. The aim is to reduce the number of flow vectors, while maintaining the geometric structure of the flow field, and to obtain different representations of local dominant motion flows in a fine-to-coarse scale. Figure \ref{fig:quantization_clustering_cell} illustrates this process.

The first quantisation step uses the flow vector angle and considers a full-degree histogram (360\si{\degree}) with 8 bins to represent orientation groups. Only the major groups, with weight above the histogram's median value, are taken into account for next clustering step. This eliminates noisy flow vectors whose orientation fall apart the expected local distribution. The second step uses the flow vector position and applies a spatial clustering on each valid orientation group. A k-means approach with center initialisation \cite{Arthur:2007:KAC:1283383.1283494} is adopted. To select a robust $k$ value, we compute k-means with increasing $k$ until the compactness measure, $C_{k}$, satisfies the condition

\begin{equation}\label{eq:cluster_convergence}
\frac{C_{k+1} - C_{k}}{C_{k+1}} < t_{c} 
\end{equation}

\noindent with $C_{k} = \sum_{i=1}^n \|s_{i}-c_{l_{i}} \|^2 $, where $s_{i}$ is the input sample $i$, $c_{l_{i}}$ is the $l$ index clustering center to which sample $i$ belongs, and $t_{c}$ is the compactness ratio threshold, which is normally selected very low ($\approx 0.01$), and we set $k_{opt} = k+1$.

Several clusters per orientation group are obtained, the so-called dominant groups, which are weighted by the number of flow vectors that belong to them and are ordered in a descendent-way. These groups represent the local dominant flows, which are described by $L_{i} = (x_{i}, y_{i}, n_{i}, \theta_{i})$, where $(x_{i}, y_{i})$ is the average position, $n_{i}$ is the total number of flow vectors, and $\theta_{i}$ is the average orientation angle. A fine-to-coarse representation, with three levels of flow vector granularity, is obtained:
\begin{inparaenum}[i)]
\item \emph{flow vectors}, the set of all flow vectors collected and filtered, $\mathcal F = \lbrace F_{i} \rbrace$;
\item \emph{flow groups}, the prototypes of the set of local dominant groups, $\mathcal L = \lbrace L_{i} \rbrace$;
\item \emph{flow group representative}, the principal local dominant group, $L_{rep}$.
\end{inparaenum}
They are useful for computational consumption requirements and to investigate their length and time scale impact on the dynamics of the motion advection system. This three-level global flow vector representation is obtained per mini-batch. However, it can be accumulated during the last $b$ mini-batches, so-called \emph{memory cell}, in order to create dense flow field representations for different discriminative levels. In this paper, we explore these representations for streamline diffusion.

At this stage, each cell is able to estimate local region parameters using information from its neighborhood. One of them is the entropy, whose computation might not be reliable if a small number of samples is presented. To overcome this problem, for each cell we collected as samples its flow vectors, the \emph{flow groups} of the cells that belongs to its neighborhood of size $K \times K$, and replicate its flow vectors at boundary cells. The entropy calculation relies on the probability of each angular bin $x_{i}$, $p(x_{i}) = C(x_{i})/\sum\nolimits_{i=1}^n C(x_{i})$, where $C(x_{i})$ corresponds to the number of vectors in bin $x_{i}$. Entropy is then measured and used on this work as an indicative of: 
\begin{inparaenum}[i)]
\item degree of vector variation in its local neighborhood, which measures the saliency value and highlights the possibility of being a critical point or belonging to a separation line in the global vector field;
\item degree of vector variation between subsequent mini-batches, which measures the difference in the information content between both vector fields, considering the distribution shape difference and highlighting a possible new local motion pattern.
\end{inparaenum}

\subsection{Flow Model Advection} \label{subsec:flow_model_advection}
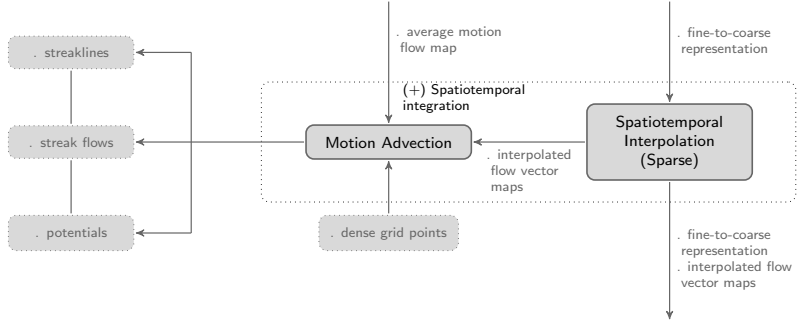
\begin{figure}[h!]
\centering
\resizebox{13.5cm}{!}{
\def\scalefactor{1.0}
\def\yshiftlabel{0.}

\begin{tikzpicture}[
	  scale=\scalefactor,
	  transform shape,
	  font=\sffamily,
	  every matrix/.style={ampersand replacement=\&,column sep=1cm,row sep=0.6cm, outer sep=1pt},
	  inputarrow/.style={anchor=east,single arrow, draw,line width=1pt, minimum height=1.5mm, minimum width=1.0cm, single arrow head extend=1.2mm, single arrow tip angle=120, draw=black!60!white,fill=white,font=\sffamily\footnotesize},
	  source/.style={draw,thick,rounded corners,fill=yellow!20,inner sep=.2cm},
	  process/.style={draw,thick,circle,fill=blue!20},
	  transforminput/.style={draw=black!60!white,dotted,rounded corners,inner sep=.2cm,fill=white,fill opacity=0.25,align=left,font=\sffamily\scriptsize,black!60!white,text opacity=1.0}, 
	  terminal/.style={draw=black!60!white,thick,rounded corners,inner sep=.2cm,fill=black!60!white,fill opacity=0.25,align=center,font=\sffamily\footnotesize,text opacity=1.0,text width=2.5cm*\scalefactor}, 
	  common_inner/.style={rectangle, draw, dotted,inner ysep=10pt,inner xsep=1pt,rounded corners}, 
	  comment/.style={rectangle, inner sep= 5pt, text width=2.5cm, node distance=0.25cm, align=left,font=\sffamily\scriptsize},
	  sink/.style={source,fill=green!20},
	  dots/.style={gray,scale=2},
	  edge/.style={semithick,font=\sffamily\scriptsize,black!60!white},
	  to/.style={->,>=stealth',shorten >=0pt,edge},
	  every node/.style={align=center},
	  skip loop/.style={to path={-- ++(0,#1) -| (\tikztotarget)}, rounded corners},
	  vh path/.style={to path={-| (\tikztotarget)}, rounded corners,font=\sffamily\footnotesize}]

  	\matrix (geral_matrix) [
		scale=\scalefactor,
		transform shape]
	{
		\& \& \& \& \coordinate (A1_6) {}; \& \& \coordinate (A1_7) {}; \& \\
		
		\& \node[transforminput,text width=1.8cm,align=center] (streaklines) {. streaklines};  \& \coordinate (A2_4) {}; \& \& \& \& \& \\
		
		\& \node[transforminput,text width=1.8cm,align=center] (streakflows) {. streak flows};  \& \coordinate (A3_4) {}; \& \& \node[terminal] (motion_advection) {Motion Advection}; \& \& \node[terminal] (spatio_temporal_interp) {Spatiotemporal\\Interpolation\\(Sparse)}; \\
		
		\& \node[transforminput,text width=1.8cm,align=center] (potentials) {. potentials}; \& \coordinate (A4_4) {}; \& \&\node[transforminput,,align=center] (grid_points) {. dense grid points}; \& \& \\
		
		\& \& \& \& \& \& \& \& \& \\
		
		\& \& \& \& \& \& \coordinate (A6_5) {}; \& \& \\

		\& \& \& \coordinate (A4_3) {}; \& \& \& \coordinate (A4_5) {}; \& \\					
	};

	\node[common_inner, left=0.01cm*\scalefactor, inner xsep=21pt, fit=(motion_advection) (spatio_temporal_interp)] (container_minibatch) {};

	\node [comment, above=\yshiftlabel of motion_advection, xshift=1.5cm] (comment-motion_advection) {(+) Spatiotemporal integration};
			 
	\draw[to] (A1_7) -- node[above right,align=left, yshift=-0.2cm] {. fine-to-coarse\\ \;representation} (spatio_temporal_interp);
	\draw[to] (A1_6) -- node[above right,align=left] {. average motion\\ \;flow map} (motion_advection);
	\draw[to] (grid_points) -- node[right,align=left] {} (motion_advection);
	\draw[to,<-] (streakflows) -- node[] {} (motion_advection);
	\draw[edge] (A3_4) -- node[] {} (A2_4);
	\draw[edge] (A3_4) -- node[] {} (A4_4);
	\draw[to,<-] (streaklines) -- node[] {} (A2_4);
	\draw[to,<-] (potentials) -- node[] {} (A4_4);
	\draw[to,<-] (motion_advection) -- node[below,align=left] {. interpolated\\ \;flow vector\\ \;maps} (spatio_temporal_interp);
	\draw[edge] (streaklines) -- node[] {} (streakflows);
	\draw[edge] (streakflows) -- node[] {} (potentials);
	\draw[to] (spatio_temporal_interp) -- node[below right,align=left, yshift=0.5cm] {. fine-to-coarse\\ \;representation\\. interpolated flow\\ \;vector maps} (A6_5);		
	
	\node [font=\scriptsize\ttfamily,text width=2.5cm,align=left] at (container_frame.north west) [above right] {};
      
\end{tikzpicture}}
\caption[System Flow Model Advection step.]{System Flow Model Advection step.}
\label{fig:system_flow_advection}
\end{figure}

This stage captures long-range temporal dependencies to represent spatial and temporal features of the flow. It is responsible to advect the motion considering an average flow map along the entire mini-batch and a dense grid of particles. It uses the global fine-to-coarse flow vector representation in a two-fold way: 
\begin{inparaenum}[i)]
\item interpolated to produce sparse vector maps;
\item in combination with a sampling strategy to enrich the representation.  
\end{inparaenum}
This stage is executed at each mini-batch and its output is a set of motion representations such as streaklines, potentials, streak flow, and interpolated flow vector maps, among others (see Figure \ref{fig:system_flow_advection}).

\subsubsection{Motion Advection and Spatiotemporal Interpolation} \label{subsubsec:motion_advection}
A dense grid of particles is considered. Each particle has fluid properties and their initial position correspond to each pixel on image. This characteristic follows the assumption that the computation of the streakline vector field needs a dense path line integration \cite{Weinkauf:2010:SLT:1907651.1908009}. All particles are integrated over time accordingly to an average of the optical flow maps along the current mini-batch, which is restarted at the beginning of the next mini-batch. On each time step a particle on position $p$ is created and all other particles previously initialised on same position follows the flow field. This process is expressed by equation \eqref{eq:particle_integration} and is repeated along the \emph{memory cell} size to obtain the streaklines. The streak flow is extracted from temporal integration of the velocity field. We use the Runge-Kutta-Fehlberg (a.k.a. RKF45) for this purpose.

Considering that each streakline is a collection of particles, 
we get a set of 3D data points for $x$ and $y$ flow directions. To compute the streak flow, $\Omega_{s} = (u_{s}(x, y), v_{s}(x,y))$, with sub-pixel level accuracy, we adopt a multi-resolution method based on B-spline refinement to approximate scattered data by error minimization on both dimensions. The 3D data points are given as input, the tensor product B-spline surfaces are produced, and the result is a least square approximation to the scattered data with B-splines for each flow direction that represents the streak flow on each direction $(x, y)$ \footnote[1]{We used the Least Squares Approximation of Scattered Data with B-splines Library (\url{http://www.sintef.no/Projectweb/Geometry-Toolkits/LSMG/}).}. The same procedure is used to obtain the spatiotemporal interpolated flow maps of each fine-to-coarse representation, considering instead the accumulated flow vectors of each one.

\subsection{Streamline Diffusion-Linking} \label{subsec:streamlines_linking}
\begin{figure}[h!]
\centering
\scalebox{0.75}{
\def\scalefactor{1.0}
\def\yshiftlabel{0.}

\begin{tikzpicture}[
	  scale=\scalefactor,
	  transform shape,
	  font=\sffamily,
	  every matrix/.style={ampersand replacement=\&,column sep=1cm,row sep=0.6cm, outer sep=1pt},
	  inputarrow/.style={anchor=east,single arrow, draw,line width=1pt, minimum height=1.5mm, minimum width=1.0cm, single arrow head extend=1.2mm, single arrow tip angle=120, draw=black!60!white,fill=white,font=\sffamily\footnotesize},
	  source/.style={draw,thick,rounded corners,fill=yellow!20,inner sep=.2cm},
	  process/.style={draw,thick,circle,fill=blue!20},
	  transforminput/.style={draw=black!60!white,dotted,rounded corners,inner sep=.2cm,fill=white,fill opacity=0.25,align=left,font=\sffamily\scriptsize,black!60!white,text opacity=1.0}, 
	  terminal/.style={draw=black!60!white,thick,rounded corners,inner sep=.2cm,fill=black!60!white,fill opacity=0.25,align=center,font=\sffamily\footnotesize,text opacity=1.0,text width=2.5cm*\scalefactor}, 
	  common_inner/.style={rectangle, draw, dotted,inner ysep=10pt,inner xsep=1pt,rounded corners}, 
	  comment/.style={rectangle, inner sep= 5pt, text width=2.5cm, node distance=0.25cm, align=left,font=\sffamily\scriptsize},
	  sink/.style={source,fill=green!20},
	  dots/.style={gray,scale=2},
	  edge/.style={semithick,font=\sffamily\scriptsize,black!60!white},
	  to/.style={->,>=stealth',shorten >=0pt,edge},
	  every node/.style={align=center},
	  skip loop/.style={to path={-- ++(0,#1) -| (\tikztotarget)}, rounded corners},
	  vh path/.style={to path={-| (\tikztotarget)}, rounded corners,font=\sffamily\footnotesize}]

  	\matrix (geral_matrix) [
		scale=\scalefactor,
		transform shape]
	{
		\& \& \& \coordinate (A1_6) {}; \& \& \coordinate (A1_7) {}; \& \& \\
		
		\& \& \& \node[transforminput,,align=center] (grid_points) {. dense grid points}; \& \&  \& \\
		
		\& \coordinate (A3_2) {}; \& \& \node[terminal] (interpolated_flow_vectors) {Spatiotemporal\\Interpolation\\(Dense)};  \& \& \coordinate (A3_8) {}; \\
		
		\& \& \& \& \& \& \& \\
		
		\& \& \& \node[terminal] (streamline_diffusion) {Streamline Diffusion-Linking}; \& \& \& \\
		
		\&  \coordinate (A5_2) {}; \&  \& \coordinate (A5_5) {}; \& \& \coordinate (A5_8) {}; \\

%
%
%
%
%
%
		\& \& \& \coordinate (A4_3) {}; \& \& \coordinate (A4_5) {}; \& \\					
	};

	\node[common_inner, inner xsep=25pt, inner ysep=5pt, fit=(interpolated_flow_vectors) (streamline_diffusion)] (container_cell_history) {};
			 
	 \node at (container_cell_history.south) [terminal,dotted, text width=5.5cm, align=left, yshift=-1.cm*\scalefactor] (output) {(+) Representation of Motion Dynamics\\(+) Physical Motion Patterns};
	 			 
	\draw[to] (grid_points) -- node[right,align=left] {} (interpolated_flow_vectors);
	\draw[to] (A3_2) -- node[below left,align=left] {. average\\ \;streak map} (interpolated_flow_vectors);
	\draw[to] (A3_8) -- node[below right, align=left] {. fine-to-coarse\\ \;representation} (interpolated_flow_vectors);
	\draw[to] (interpolated_flow_vectors) -- node[right,align=left] {. interpolated\\ \;streak vector map} (streamline_diffusion);
	\draw[to] (streamline_diffusion) -- node[] {} (A5_5);	
	
      
\end{tikzpicture}}
\caption[System Streamline Diffusion-Linking step.]{System Streamline Diffusion-Linking step.}
\label{fig:system_streamline_linking}
\end{figure}
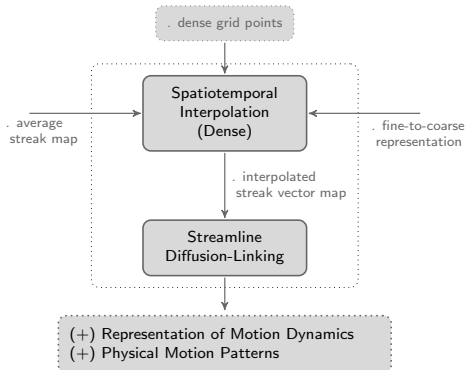

Due to the ending conditions of the streamline diffusion process, described on Section \ref{subsec:streamlines_computation_information}, and since we are interest in extracting long-range streamlines to represent global motion trajectories, we include a post-processing step that links short streamlines. This stage uses the flow information collected over a sequence of mini-batches, of \emph{memory cell} size (see Figure \ref{fig:system_streamline_linking}). 

Streamline diffusion requires as input a vector field obtained from a dense flow field and the largest it represents the temporal and local changes over time, the longer the streamlines, and the better they emphasise the global field temporal coherency and the better they describe the topology of the flow. To this end, we use a combination of: 
\begin{inparaenum}[i)]
\item the averaged streak flow from the current set of mini-batches; 
\item the set of flow vectors from a specific fine-to-coarse representation collected along the current set of mini-batches;
\item a dense grid of particles.
\end{inparaenum}
The resulting flow field is formed by B-spline interpolation, as explained on Section \ref{subsubsec:motion_advection}, considering as input the set of flow vectors from the fine-to-coarse representation superimposed to the averaged streak flow. This flow field is converted into a vector field using a grid-based discretisation and the filtering step explained on Section \ref{subsubsec:filtering}. We adopted the state-of-the-art farthest point seeding method \cite{10.1109/VIS.2005.39} as the streamline diffusion technique \footnote[2]{Algorithm is implemented on the Computational Geometry Algorithms library (CGAL) \url{http://www.cgal.org/}.}. 

The streamline linking process permits to obtain long-range streamlines. It is formulated as a combinatorial matching problem that considers compatibility in terms of flow appearance, motion, and spatiotemporal regularisation among all short-streamlines. We adopted a discrete Markov Random Field (MRF) process to encode association constraints between \emph{query} and \emph{candidate} streamlines, re-correlate the set of short-streamlines and extract an optimal linkage between them. This undirected model was inspired on \cite{Rubinstein12Towards}. Our formulation in terms of probability of linkage, $\mathcal L$, between streamlines is defined by

\begin{equation}\label{eq:linkage} 
    P(\mathcal L) = \prod_{i} \phi_{i}(l_{i}) \prod_{i, j \in \mathcal N(i)} \psi_{i, j} (l_{i}, l_{j}),
\end{equation}

\noindent  where $\phi_{i}(l_{i})$ are the unary potentials that model the compatibility between a query streamline, $q_{i}$, and a candidate streamline, $c_{j}$; $\psi_{i, j} (l_{i}, l_{j})$ are the pairwise potentials for link regularisation, in case of tracking ambiguities, between a pair of query streamlines, $q_{i}$ and $q_{j}$, considering the candidate streamlines that lay in their spatiotemporal neighborhood, $\mathcal N(i)$. The global optimisation problem, given by equation \ref{eq:linkage}, is inferred using a tree-reweighted belief propagation. Under this context, the streamlines taken on the MRF process are called tracks.

The compatibility term is divided into three components:
\begin{inparaenum}[a)]
\item the appearance similarity, $\phi_{a}$, which models the flow properties;
\item the motion similarity, $\phi_{m}$, which takes into account the velocity information;
\item the prior on motion model, $\phi_{p}$, which approximates a motion model to predict next streamline's position in case of large discontinuities.
\end{inparaenum}
Appearance and motion similarity terms consider a symmetrically weighted average comparison of features along the last $n$ elements of the query track, $q_{i}$, and the first $n$ elements of the candidate track, $c_{j}$. The weight is an exponentially decaying factor, $w_{t}(k) = \alpha^k, 0 < \alpha < 1$, that works as a confidence parameter.

Instead of considering an individual average information (motion or appearance) for each track and then take their difference in the similarity term, as used in \cite{Rubinstein12Towards}, we adopted point-to-point operations. We formulate the appearance term between tracks $\Gamma_{i}$ and $\Gamma_{j}$ based on the cosine similarity of the streak flow's angle at each track's position given by

\begin{gather}\label{eq:appearance_term} 
\begin{aligned}
    \mathbf{s}_{ij} =  \frac{1}{Z} \sum\limits_{k=0}^{n_{a}-1} & (S_{i}(t_{i}^{end} - k)w_{o}(t_{i}^{end} - k) - 
    &S_{j}(t_{j}^{start} + k) w_{o}(t_{j}^{start} + k)) w_{t}(k) 
\end{aligned}
\end{gather}

\noindent where $S_{i}(t)$ is the track's, $\Gamma_{i}$, cosine of the streak flow angle at time $t$, $w_{o}(t)$ is an outlier weight that measures how well $S_{i}(t)$ fits the appearance characteristics of the entire track, $\Gamma_{i}$, which is modelled by a Gaussian distribution of the track's streak flow angles. The same is defined for track $\Gamma_{j}$. The normalisation factor is expressed by 

\begin{equation}\label{eq:appearance_normalization_factor} 
    Z = \sum\limits_{k=0}^{n_{a}-1} (w_{o}(t_{i}^{end} - k) - w_{o}(t_{j}^{start} + k)) w_{t}(k) 
\end{equation}

\noindent and the appearance similarity is defined by

\begin{equation}\label{eq:appearance_similarity} 
    \phi_{a} = \exp(-\frac{1}{\sigma_{a}^2}\lVert \mathbf{s}_{ij} \rVert)
\end{equation}

The motion term considers the velocity variation. Similarly, the velocity difference is taken by a point-to-point track relation stated by

\begin{equation}\label{eq:motion_term} 
    \mathbf{v}_{ij} = \sum\limits_{k=0}^{n_{v}-1} (v_{i}(t_{i}^{end} - k) - v_{j}(t_{j}^{start} + k)) w_{t}(k),
\end{equation}

\noindent where $v_{i}(t)$ is the track's, $\Gamma_{i}$, velocity at time $t$. The same is defined for track $\Gamma_{j}$. The motion similarity is expressed by

\begin{equation}\label{eq:motion_similarity} 
    \phi_{m} = \exp(-\frac{1}{\sigma_{m}^2}\lVert \mathbf{v}_{ij} \rVert)
\end{equation}

The prior on motion model that predicts track's movement on discontinuities considers linear kinematic equations to estimate the closest point of the query track, $\Gamma_{i}$, to the initial point of the candidate track, $\Gamma_{j}$. The motion integration is done until the distance travelled equals the length between the last point of $\Gamma_{i}$ and the first point of $\Gamma_{j}$, and is governed by

\begin{gather}\label{eq:motion_linear_model} 
\begin{aligned}
    \mathbf{x}_{i}(t+1) & = \mathbf{x}_{i}(t) + \mathbf{v}_{i}(t) + \frac{1}{2}\mathbf{a}_{i}(t) + \mathbf{v}_{i}^{flow}(t) \\
    \mathbf{a}_{i}(t+1) & = \mathbf{v}_{i}(t+1) -  \mathbf{v}_{i}(t) 
\end{aligned}    
\end{gather}

\noindent where $\mathbf{v}_{i}^{flow}(t)$ is the flow vector velocity at position $\mathbf{x}_{i}(t)$. The next velocity, $\mathbf{v}_{i}(t+1)$, is randomly chosen from a Gaussian distribution of the velocities of the track $\Gamma_{i}$. After this, a weighted distance, that includes spatial and angular values between the last predicted point of the query track and the initial point of the candidate track, is used to obtain the motion discontinuity similarity term

 \begin{equation}\label{eq:motion_discontinuity_similarity} 
    \phi_{p} = \exp(-\frac{1}{\sigma_{p}^2} \bigl( \alpha \lVert \mathbf{x}_{j}^{start}(t) - \mathbf{x}_{i}^{end}(t) \rVert + (1 - \alpha) \angle (\mathbf{l}_{i},\mathbf{l}_{j}) \bigr) )
\end{equation}

\noindent where $\mathbf{l}_{i}$ is the last segment of the query track, $\mathbf{l}_{j}$ is the first segment of the candidate track, $\angle (\mathbf{l}_{i},\mathbf{l}_{j})$ is the angle between both segments, and $\alpha$ is a weighted factor (in this case $0.5$). For each set of mini-batches, the system extracts a set of streamlines, which are connected with the streamlines obtained from the subsequent set of mini-batches. 

Every node in the graph, i.e. every streamline, has an additional state with a predefined cost to represent the terminal state and to avoid a forced linking. The graph just defines unary potentials among streamlines that present a compatibility term, $\phi_{i}$, below a predetermined threshold. Candidate track's formation is evaluated under geometrical constraints:
\begin{inparaenum}[i)]
\item length ($d_{thr}$), which defines the spatial distance between the last point of the query track and the first point of the candidate track;
\item continuity direction ($\theta_{dir}$), which returns the angle between the last segment of the query track and the segment that links the last point of the query track with the first point of the candidate track;
\item direction difference ($\delta_{dif}$), which states the angular difference between the last segment of the query track and the first segment of the candidate track.
\end{inparaenum}
Only the ones that satisfy pre-defined thresholds are included in the graph. In the same way, the compatibility term between query tracks, $\psi_{i, j}$, follows a geometrical pruning with the same constraints, excluding the $\delta_{dif}$ constraint. In terms of temporal neighboring, only the tracks which belong to the same set of mini-batches are considered in the link regularisation step. This pruning process effectively reduces the computational effort without affecting the final results. 

We took further advantage of the system's characteristics and used the cell's entropy to detect the areas with high entropy values. The query tracks whose ending points and candidate track whose starting points fall on these regions are not considered in the linking process. This highly improves efficiency.

\section{Experimental Results} \label{sec:experimental_results}
We present results on several datasets according to the following criteria:
\begin{inparaenum}[i)]
\item crowd scenario, both structured and unstructured movement;
\item multi-tracking scenario, both sparse and dense groups (see Table \ref{table:datasets}).
\end{inparaenum}

\begingroup
\renewcommand*{\thefootnote}{\alph{footnote}}
\begin{table}[ht] 
\centering 
\scalebox{0.72}
{
\begin{tabular}{cccccc>{\centering\bfseries}m{1in} >{\centering}m{1in} >{\centering}m{1in} >{\centering\arraybackslash}m{1in}}
 \toprule
  Name & Category & Frame Size & Fps & N frames \\
  \midrule
  UCF 913-36l \footnotemark[1]
  & Crowd Structured (C.S.) & 480x360 & 25 & 467 \\ 
  UMN seq3 \footnotemark[2] 
  & Crowd Unstructured (C.U.) & 320x240 & 30 & 658 \\ 
  
  PETS2013 S2 L1 Time12-34 View001 & Dense Multi-Tracking (D.MT.)  & 768x576 & -- & 794 \\
  
  PETS2013 S2 L3 Time14-41 View001 & Sparse Multi-Tracking (S.MT.) & 768x576 & -- & 240 \\  
  \bottomrule
\end{tabular}
}
\caption{Datasets characteristics.}\label{table:datasets} 
\end{table}
\footnotetext[1]{UCF Crowd Segmentation dataset \url{http://crcv.ucf.edu/data/crowd.php}}
\footnotetext[2]{Unusual Crowd Activity dataset \url{http://mha.cs.umn.edu/Movies/Crowd-Activity-All.avi}}
\endgroup

\subsection{System Parameters}\label{subsec:experimental_system_parameters}
In our previous work \cite{EPereira_flow_trajectories}, we tested the instantiation stage with satisfactory results. A baseline was reached, namely we adopted the FAST sampling, the median filtering kernel size of $K = (13, 13)$, the LDOF's optical flow algorithm \cite{Brox:2011:LDO:1936329.1936562}, and a spatial cell size of $C = (15, 15)$. In this work, considering empirical research, we fixed the neighborhood size to $\mathcal{N} = (3, 3)$.

\begin{table}[!ht] 
\centering
\parbox{0.40\linewidth}
{
\centering
\resizebox{\linewidth}{!}
{
\begin{tabular}{cccccc>{\centering\bfseries}m{1in} >{\centering}m{1in} >{\centering}m{1in} >{\centering\arraybackslash}m{1in}}
 \toprule
  \multicolumn{1}{c}{Memory Cell Size} & \multicolumn{5}{c}{Minibatch Size} \\
  \midrule
  \multicolumn{1}{c}{10} & 2 & 4 & 6 & 8 & 10  \\
\end{tabular}
}
}

\vspace{0.5em}

\parbox{0.46\linewidth}
{
\centering
\resizebox{\linewidth}{!}
{  
\begin{tabular}{ccccccc>{\centering\bfseries}m{1in} >{\centering}m{1in} >{\centering}m{1in} >{\centering\arraybackslash}m{1in}}
  \toprule
  \multicolumn{1}{c}{Minibatch Size} & \multicolumn{6}{c}{Memory Cell Size} \\
  \midrule
  \multicolumn{1}{c}{5} & 3 & 6 & 9 & 12 & 15 & 20 \\
  \bottomrule
\end{tabular}
}
}
\caption{System parameters.}\label{table:system_parameters} 
\end{table}

We identified four system parameters:
\begin{inparaenum}[i)]
\item cell size, which influences the granularity from which motion is aggregated, filtered, and advected;
\item kernel size, which conveys a rule for common kernel operations such as flow vector computation and filtering;
\item \emph{minibatch} size, which sets the length of the streaklines, as well as the flow vectors quantisation;
\item \emph{memory cell} size, which sets the number of mini-batches for streamline formation. 
\end{inparaenum}
After some experiences, we verified that the most relevant are the last two. For further analysis, we vary \emph{minibatch} size fixing \emph{memory cell} size, and vice versa according to Table \ref{table:system_parameters}.

\subsection{Outlier Removal}\label{subsec:experimental_outlier_removal}
The outlier removal technique assumes a predominant role on system accuracy. Figure \ref{fig:outliers} provides a qualitative confirmation. A quantitative comparison of the proposed technique with several outlier removal methods is provided in Table \ref{table:outlier_removal}, reported under three metrics, namely true positive (TP) rate, true negative (TN) rate and the mean of both, here called the true balance (TB) rate. Results were computed considering the (C.S.) and (C.U.) scenarios. For the former, nearly 20 frames were masked, since it is a large video sequence with many persons per frame, while for the latter, all the frames were masked. Pedestrians were manually annotated with bounding boxes and the masks are the inner ellipses within each one. The flow vectors that lay outside the masks are considered the background, i.e. the outliers, which are considered as negative samples.

\begin{table}[!ht] 
\centering 
\scalebox{0.57}
{
\begin{tabular}{ccccccccc>{\centering\bfseries}m{1in} >{\centering}m{1in} >{\centering}m{1in} >{\centering\arraybackslash}m{1in}}
 \toprule
   & Std & Thopmson-Tau \cite{DBLP:journals/Thompson85} & Mzscore & Zscore & Adj-Boxplot \cite{Hubert:2008:ABS:1393650.1393903} & ExpSM \cite{DBLP:journals/Kimber82} & Grubbs \cite{Frank_Grubbs:Glenn_Beck:1972} & Ours \\
 \midrule
 \multicolumn{9}{c}{UMN seq3} \\ 

   \midrule
  TP & 97.9 ($\pm$1.3) & 18.9 ($\pm$5.5) & 27.4 ($\pm$22.6) & \textbf{98.7} ($\pm$0.9) & 58.9 ($\pm$10.7) &  68.0 ($\pm$4.6) & 92.2 ($\pm$15.9) & 79.7 ($\pm$6.5) \\ 

  TN & 1.7 ($\pm$1.2) & 19.2 ($\pm$8.1) & 16.5 ($\pm$6.7) & 1.2 ($\pm$0.8) & 12.7 ($\pm$5.5) & 21.0 ($\pm$4.6) & 3.6 ($\pm$4.4) &  \textbf{82.7} ($\pm$6.0)  \\   
  
  TB & 49.8 & 19.1 & 21.9 & 49.5 & 35.8 & 44.5 & 47.9 & \textbf{81.2}  \\   
  
  \midrule
  \multicolumn{9}{c}{UCF 913-36l} \\
   \midrule
  TP & 96.6 ($\pm$5.7) & 13.8 ($\pm$13.6) & 29.3 ($\pm$33.7) & \textbf{97.9} ($\pm$3.6) & 58.6 ($\pm$21.7) & 77.3 ($\pm$12.1) & 76.8 ($\pm$31.6) & 79.9 ($\pm$14.2) \\ 

  TN & 2.3 ($\pm$1.8) & 24.8 ($\pm$21.9) & 17.8 ($\pm$10.7) & 1.7 ($\pm$1.3) & 21.3 ($\pm$21.1) & 16.5 ($\pm$7.2) & 7.8 ($\pm$6.2) &  \textbf{89.7} ($\pm$6.4)  \\   
  
   TB & 48.4 & 19.3 & 23.5 & 49.8 & 39.9 & 46.9 & 42.3 & \textbf{84.8}  \\
  \bottomrule
\end{tabular}
}
\caption{Sensitivity and Specificity of various outlier removal techniques ($\%$).}\label{table:outlier_removal} 
\end{table}

\begin{table}[!ht] 
\centering
\begin{tabular} {llll}

	\includegraphics[width=.35\textwidth]{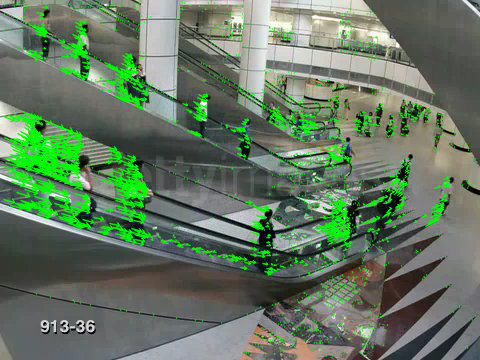}\label{fig:outlier_flow_vectors_UCF} &
	\includegraphics[width=.35\textwidth]{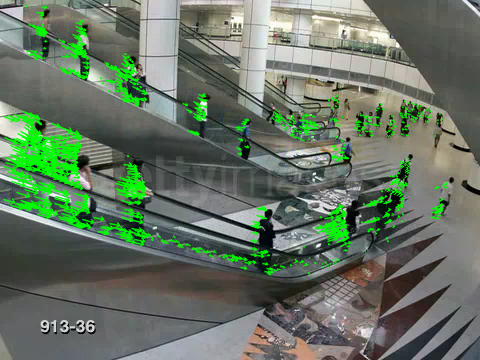}\label{fig:flow_vectors_UCF}  \\
	
	\includegraphics[width=.35\textwidth]{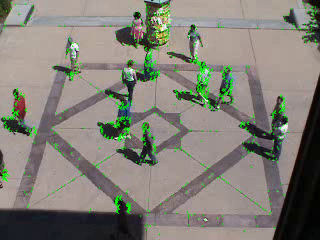}\label{fig:outlier_flow_vectors_MHA} &
	\includegraphics[width=.35\textwidth]{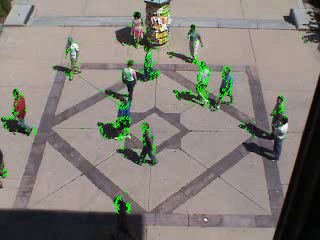}\label{fig:flow_vectors_MHA} \\
	
	\includegraphics[width=.35\textwidth]{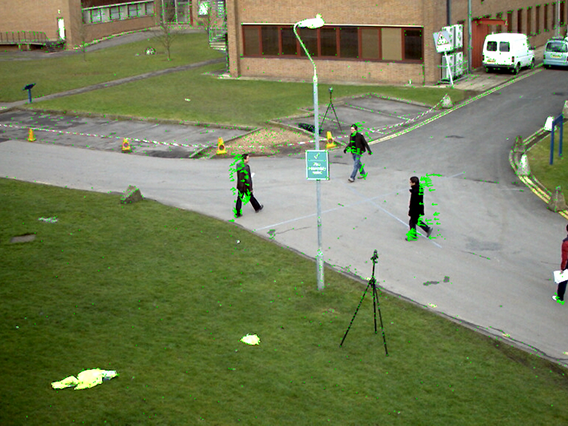}\label{fig:outlier_flow_vectors_PETS_S2_L1} &
	\includegraphics[width=.35\textwidth]{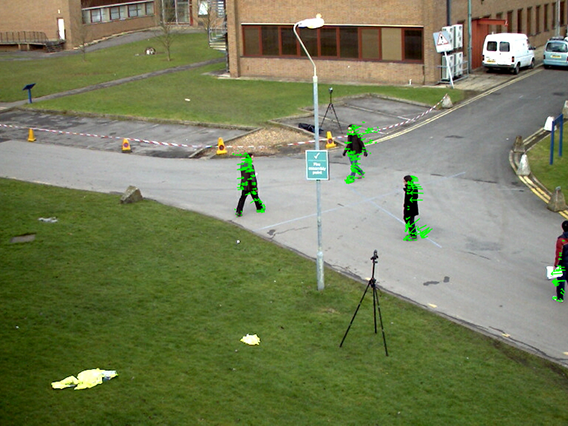}\label{fig:flow_vectors_PETS_S2_L1} \\
	
	\includegraphics[width=.35\textwidth]{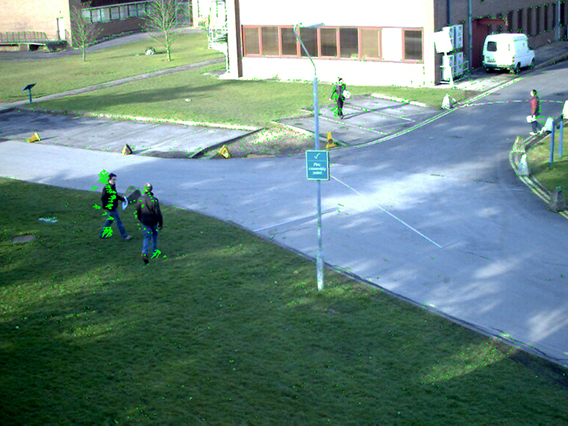}\label{fig:outlier_flow_vectors_PETS_S2_L3} &
	\includegraphics[width=.35\textwidth]{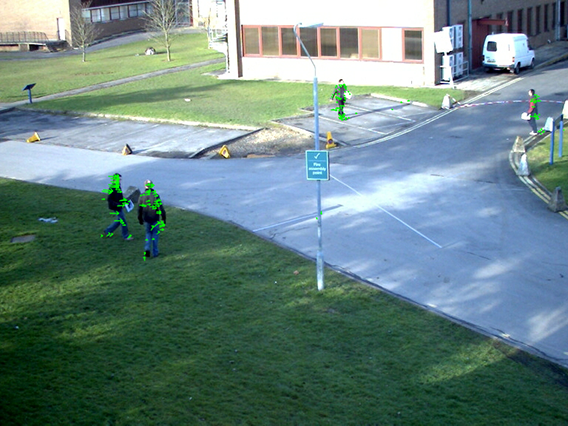}\label{fig:flow_vectors_PETS_S2_L3}\\
	
\end{tabular}
	\captionof{figure}[Comparison results with and without the proposed outlier removal technique. By row: from left to right, with outliers and after outlier removal. By column: from top to bottom, C.S., C.U., D.MT., and S.MT.]{Comparison results with and without the proposed outlier removal technique. By row: from left to right, with outliers and after outlier removal. By column: from top to bottom, C.S., C.U., D.MT., and S.MT.} 
	\label{fig:outliers}			
\end{table}

We have tested with simple and more complex techniques (for most of them, we used the available source code). For instance, the most \emph{na{\"i}ve} approach, the \emph{Std} method, considers as outliers the samples that distance from the mean more than $3\sigma$, while more robust approaches follow statistical models that deal with skewed data. All techniques are applied on the log-normalised magnitude distribution of flow vectors. Results show that our technique keeps a high TP rate and, more importantly, it demonstrates a higher TN rate, proving its effectiveness on removing background's flow vectors. None of the others techniques are able to achieve satisfactory TN rate, which support our evidence to propose a new outlier removal technique for this problem. The combined TB rate clearly shows the overall supremacy of our technique.  We conclude that the combination of the Chebyshev's theorem with the skewness metric brings a stabilisation factor to the initial rough approximation of the unimodal normal distribution for the formulation of our outlier removal technique.

\begin{figure}[h!]
\centering
	\subfigure[][]{\includegraphics[width=.31\textwidth]{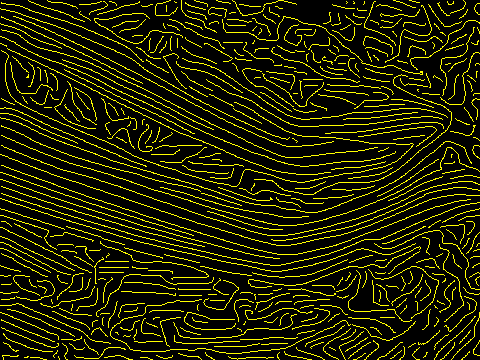}\label{fig:streamline_outlier_UCF}}
	\subfigure[][]{\includegraphics[width=.31\textwidth]{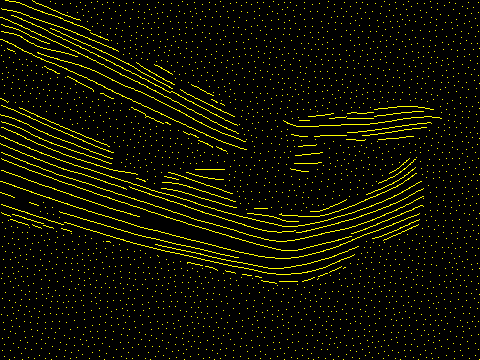}\label{fig:streamline_UCF}}
	\caption[Effects caused by the outlier removal technique on streamline formation (C.S. scenario):
			\subref{fig:streamline_outlier_UCF} without outlier removal;			
			\subref{fig:streamline_UCF} with outlier removal (single points are seeds that were, correctly, not diffused by lack of meaningful flow field).]{Effects caused by the outlier removal technique on streamline formation (C.S. scenario):
			\subref{fig:streamline_outlier_UCF} without outlier removal;			
			\subref{fig:streamline_UCF} with outlier removal (single points are seeds that were, correctly, not diffused by lack of meaningful flow field).} 
	\label{fig:outlier_interference}
\end{figure}

This step is also crucial for further motion advection step. Figure \ref{fig:outlier_interference} confirms the high perturbation that undesired flow vectors introduce on the streamline formation. The great advantage of the proposed outlier removal technique is that it is data-driven, therefore it automatically adjusts to several scene contexts, as well as to large motion variations during the same scenario. We verified that normally the percentage of flow vectors removed is between $[65\%,75\%]$, but it could varied around $[20\%,85\%]$, depending on the scene category and the video characteristics.

\subsection{Global Motion Trajectories}\label{subsec:experimental_trajectories}
Our aim in this paper is to show that our system is able to extract long-range global motion trajectories that approximate the ones obtained from multi-tracking approaches, along with the advantage to be useful and effective in different scenarios. 

The final trajectories are obtained by the linking process. The parameters of the selected farthest point seeding method were fixed to the default values, $d_{sep} = 4.3$ and $d_{rat} = 1.3$, both in sub-pixel accuracy. The former controls the spacing distance of the density field, and the latter expresses the saturation ratio to trigger the seeding of a new streamline. It is necessary to execute a pruning step to remove redundant and short streamlines. Such process removes trajectories with a number of points below the mean value of the whole set of trajectories extracted on all set of mini-batches of the entire sequence. We used in this step the number of points since after the diffusion process the points are very close to each other (nearly one pixel), therefore its value is approximately equal to the trajectory length. Table \ref{table:npoints_trajectories} states the significant reduction of trajectories achieved with the pruning and linking processes.

\begin{table}[ht]
\centering 
\parbox{0.85\linewidth}
{
\centering
\resizebox{\linewidth}{!}
{
\begin{tabular}{ccccccccccccccccc>{\centering\bfseries}m{1in} >{\centering}m{1in} >{\centering}m{1in} >{\centering\arraybackslash}m{1in}}
    \toprule
    \multirow{3}{*}{Dataset} & \multicolumn{15}{c}{Minibatch Size (for memory cell size = 10)} \\ 
    
    \cline{2-16} & \multicolumn{3}{c|}{2} & \multicolumn{3}{c|}{4} & \multicolumn{3}{c|}{6} & 
    \multicolumn{3}{c|}{8} & \multicolumn{3}{c}{10} \\
    
    \cline{2-16} & \multicolumn{1}{c}{BP} & \multicolumn{1}{c}{AP} & \multicolumn{1}{c|}{AL} &
    \multicolumn{1}{c}{BP} & \multicolumn{1}{c}{AP}  & \multicolumn{1}{c|}{AL} & 
    \multicolumn{1}{c}{BP} & \multicolumn{1}{c}{AP} & \multicolumn{1}{c|}{AL} & 
    \multicolumn{1}{c}{BP} & \multicolumn{1}{c}{AP} & \multicolumn{1}{c|}{AL} & 
    BP & AP & AL\\
    
    \midrule
    C.S. & \multicolumn{1}{c}{2058} & \multicolumn{1}{c}{574} & \multicolumn{1}{c|}{434} &  
    \multicolumn{1}{c}{918} & \multicolumn{1}{c}{263} & \multicolumn{1}{c|}{222} &
    \multicolumn{1}{c}{573} & \multicolumn{1}{c}{165} & \multicolumn{1}{c|}{143} &
    \multicolumn{1}{c}{426} & \multicolumn{1}{c}{121} & \multicolumn{1}{c|}{106} &
    \multicolumn{1}{c}{405} & \multicolumn{1}{c}{106} & \multicolumn{1}{c}{89}\\
    
    C.U. & \multicolumn{1}{c}{2508} & \multicolumn{1}{c}{995} & \multicolumn{1}{c|}{169} &  
    \multicolumn{1}{c}{1381} & \multicolumn{1}{c}{540} & \multicolumn{1}{c|}{153} &
    \multicolumn{1}{c}{891} & \multicolumn{1}{c}{346} & \multicolumn{1}{c|}{136} &
    \multicolumn{1}{c}{770} & \multicolumn{1}{c}{283} & \multicolumn{1}{c|}{128} &
    \multicolumn{1}{c}{580} & \multicolumn{1}{c}{196} & \multicolumn{1}{c}{88}\\
    
    D.MT. & \multicolumn{1}{c}{6252} & \multicolumn{1}{c}{2088} & \multicolumn{1}{c|}{1171} &  
    \multicolumn{1}{c}{4156} & \multicolumn{1}{c}{1184} & \multicolumn{1}{c|}{691} &
    \multicolumn{1}{c}{3597} & \multicolumn{1}{c}{1020} & \multicolumn{1}{c|}{566} &
    \multicolumn{1}{c}{2695} & \multicolumn{1}{c}{707} & \multicolumn{1}{c|}{389} &
    \multicolumn{1}{c}{2263} & \multicolumn{1}{c}{616} & \multicolumn{1}{c}{319}\\
    
    S.MT. & \multicolumn{1}{c}{2111} & \multicolumn{1}{c}{516} & \multicolumn{1}{c|}{247} &  
    \multicolumn{1}{c}{797} & \multicolumn{1}{c}{235} & \multicolumn{1}{c|}{127} &
    \multicolumn{1}{c}{714} & \multicolumn{1}{c}{194} & \multicolumn{1}{c|}{128} &
    \multicolumn{1}{c}{424} & \multicolumn{1}{c}{125} & \multicolumn{1}{c|}{87} &
    \multicolumn{1}{c}{356} & \multicolumn{1}{c}{100} & \multicolumn{1}{c}{60}\\
    
    \bottomrule
\end{tabular}
}
}

\vspace{1.0em}

\parbox{1.0\linewidth}
{
\centering
\resizebox{\linewidth}{!}
{  
\begin{tabular}{cccccccccccccccccccc>{\centering\bfseries}m{1in} >{\centering}m{1in} >{\centering}m{1in} >{\centering\arraybackslash}m{1in}}  
    \toprule
    \multirow{3}{*}{Dataset} & \multicolumn{18}{c}{Memory Cell Size (for minibatch size = 5)} \\
    
    \cline{2-19} & \multicolumn{3}{c|}{3} & \multicolumn{3}{c|}{6} & \multicolumn{3}{c|}{9} & 
    \multicolumn{3}{c|}{12} & \multicolumn{3}{c|}{15} & \multicolumn{3}{c}{20} \\
    
    \cline{2-19} & \multicolumn{1}{c}{BP} & \multicolumn{1}{c}{AP} & \multicolumn{1}{c|}{AL} &
    \multicolumn{1}{c}{BP} & \multicolumn{1}{c}{AP}  & \multicolumn{1}{c|}{AL} &
    \multicolumn{1}{c}{BP} & \multicolumn{1}{c}{AP} & \multicolumn{1}{c|}{AL} & 
    \multicolumn{1}{c}{BP} & \multicolumn{1}{c}{AP} & \multicolumn{1}{c|}{AL} & 
    \multicolumn{1}{c}{BP} & \multicolumn{1}{c}{AP} & \multicolumn{1}{c|}{AL} &
    BP & AP & AL\\
    
    \midrule
    C.S. & \multicolumn{1}{c}{3043} & \multicolumn{1}{c}{818} & \multicolumn{1}{c|}{598} & 
    \multicolumn{1}{c}{1372} & \multicolumn{1}{c}{368} & \multicolumn{1}{c|}{297}  & 
    \multicolumn{1}{c}{879} & \multicolumn{1}{c}{244} & \multicolumn{1}{c|}{199} & 
    \multicolumn{1}{c}{618} & \multicolumn{1}{c}{165} & \multicolumn{1}{c|}{143} &
    \multicolumn{1}{c}{497} & \multicolumn{1}{c}{145} & \multicolumn{1}{c|}{122} &
    \multicolumn{1}{c}{347} & \multicolumn{1}{c}{100} & \multicolumn{1}{c}{83} \\
    
    C.U. & \multicolumn{1}{c}{3514} & \multicolumn{1}{c}{1398} & \multicolumn{1}{c|}{236} & 
    \multicolumn{1}{c}{1746} & \multicolumn{1}{c}{662} & \multicolumn{1}{c|}{175}  & 
    \multicolumn{1}{c}{1180} & \multicolumn{1}{c}{441} & \multicolumn{1}{c|}{157} & 
    \multicolumn{1}{c}{877} & \multicolumn{1}{c}{339} & \multicolumn{1}{c|}{132} &
    \multicolumn{1}{c}{716} & \multicolumn{1}{c}{267} & \multicolumn{1}{c|}{117} &
    \multicolumn{1}{c}{535} & \multicolumn{1}{c}{176} & \multicolumn{1}{c}{89} \\
       
    D.MT. & \multicolumn{1}{c}{3948} & \multicolumn{1}{c}{1294} & \multicolumn{1}{c|}{775} & 
    \multicolumn{1}{c}{5705} & \multicolumn{1}{c}{1656} & \multicolumn{1}{c|}{1001}  & 
    \multicolumn{1}{c}{4167} & \multicolumn{1}{c}{1178} & \multicolumn{1}{c|}{706} & 
    \multicolumn{1}{c}{3521} & \multicolumn{1}{c}{973} & \multicolumn{1}{c|}{552} &
    \multicolumn{1}{c}{2949} & \multicolumn{1}{c}{814} & \multicolumn{1}{c|}{456} &
    \multicolumn{1}{c}{2132} & \multicolumn{1}{c}{570} & \multicolumn{1}{c}{317} \\
    
    S.MT. & \multicolumn{1}{c}{2978} & \multicolumn{1}{c}{711} & \multicolumn{1}{c|}{351} & 
    \multicolumn{1}{c}{1260} & \multicolumn{1}{c}{332} & \multicolumn{1}{c|}{190}  & 
    \multicolumn{1}{c}{741} & \multicolumn{1}{c}{222} & \multicolumn{1}{c|}{124} & 
    \multicolumn{1}{c}{704} & \multicolumn{1}{c}{186} & \multicolumn{1}{c|}{115} &
    \multicolumn{1}{c}{347} & \multicolumn{1}{c}{113} & \multicolumn{1}{c|}{81} &
    \multicolumn{1}{c}{367} & \multicolumn{1}{c}{104} & \multicolumn{1}{c}{66} \\
    
    \bottomrule
\end{tabular}
}
}
\caption{Number of trajectories per system parameters on each dataset at different steps (BP: before pruning; AP: after pruning; AL: after linking).}
\label{table:npoints_trajectories}
\end{table}

The geometrical constraints for the MRF graph were fixed to $d_{thr}=45$, $\theta_{dir}=42^{\circ}$, and $\delta_{dif}=40^{\circ}$. For the remaining parameters used on the similarities terms, we follow \cite{Rubinstein12Towards} with the exception of $n_{p}$ and $n_{v}$, which are the number of points taken for appearance and velocity similarity terms, respectively. They were fixed heuristically considering a percentage of the trajectory with the minimum number of points ($n_{min}$), $n_{p}=0.7\times n_{min}$ and $n_{v}=0.35\times n_{min}$. The velocity term is lower due to a higher expected variation, therefore less point-to-point measures are considered. 

To the best of our knowledge there is not any evaluation framework that deals with comparison between manual trajectories and automatic long-range motion trajectories of pedestrians. We extend our previous work \cite{EPereira_flow_trajectories} to propose an evaluation methodology that supports clustering of trajectories by similarity to obtain the most representatives, and measures the correspondence between extracted and annotated trajectories. We follow a similar approach to \cite{Ochs:2014:SMO:2693343.2693376} and use a one-to-one distance function between each annotated trajectory and each auto-generated one to obtain a distance matrix, and solve the assignment problem with the Hungarian algorithm \cite{Kuhn1955}. We report the quality of the matching process with the miss detection (FN) and false positive (FP) rates. 

The distance matrix passes through a regularisation process before applying the Hungarian algorithm. This step favours configurations with small residuals and down-weights large errors, that could otherwise dominate the matching process. We evaluate four alternatives to regularise the distance matrix: 
\begin{inparaenum}[i)]
\item \emph{clustering threshold,} where a K-means is applied to the distances and the $max$ of the cluster with the lowest values is taken as the desired threshold to truncate the matrix entries up this value;
\item \emph{quartile threshold}, where the third quartile of the distances distribution is taken as threshold, and a similar truncation process is applied;
\item \emph{median RLS}, where the median of the distances is used as the $\sigma$ parameter for the robust least square (RLS) approach expressed by $\rho(u, \sigma) = u^2/(\sigma^2 + u^2)$, where $u$ is each distance value;
\item \emph{local scaling RLS}, where the same robust least square approach is taken, but instead of using the median value, the mean of a clustering based on local scaling is used. Such value is computed as follows: 
\begin{inparaenum}[1)]
\item a K-means is applied to the distances, 
\item the $max$ of each cluster is used to obtain a distance based on local scaling as stated on \cite{Zelnik-manor04self-tuningspectral}, 
\item the mean of those distances is considered to be the $\sigma$ value.
\end{inparaenum}
\end{inparaenum}

An annotated trajectory is considered correctly matched if its distance to an auto-generated one is below a certain threshold. The threshold is also used to evaluate the relationship between the accumulated error, sum of the distances between the trajectories correctly matched, and the false positive rate. Such threshold is considered to be the $max$ of the cluster with lowest values after applying the same K-means process detailed on Section \ref{subsubsec:quantization_clustering}. Before computing the distance matrix, all trajectories are resampled using a cubic spline interpolation scheme, where the resampling value is global and is equal to the length of the trajectory with the minimum number of points. 

\begin{figure}[h!]
\centering
	\subfigure[][]{\includegraphics[width=.44\textwidth]{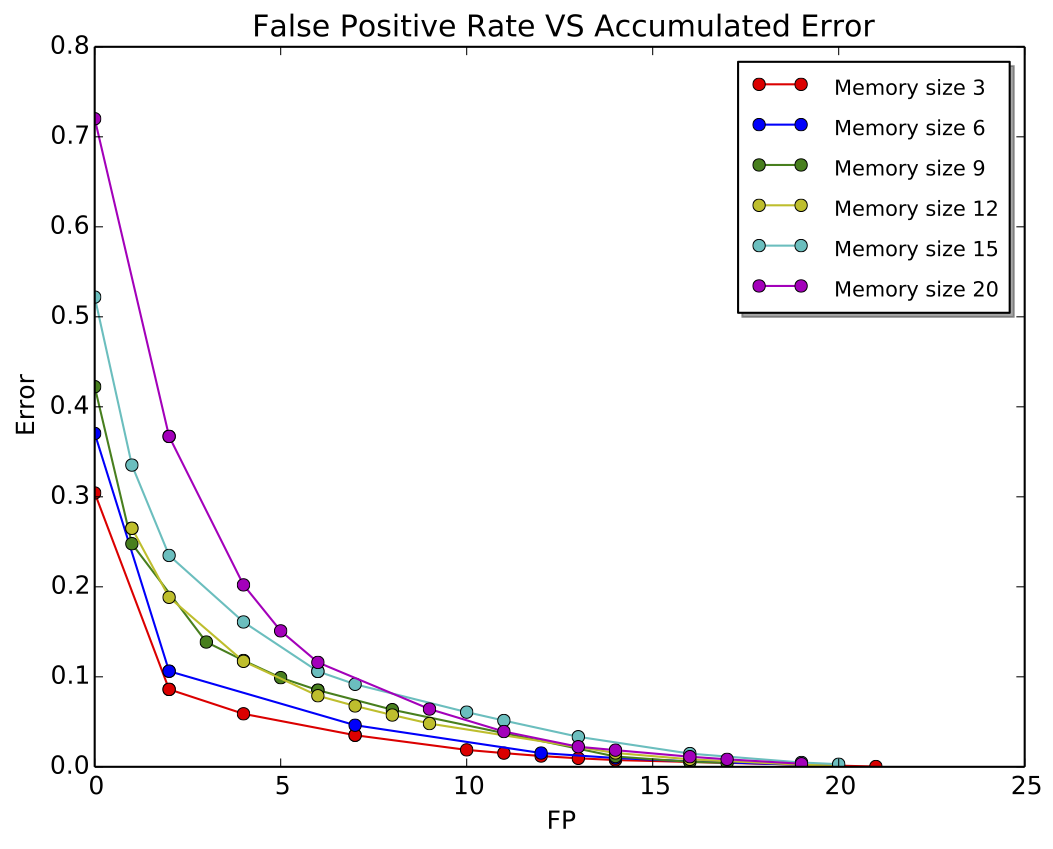}\label{fig:FPvsError_memory_DTW_RLS_local_scaling_UCF}}
	\subfigure[][]{\includegraphics[width=.44\textwidth]{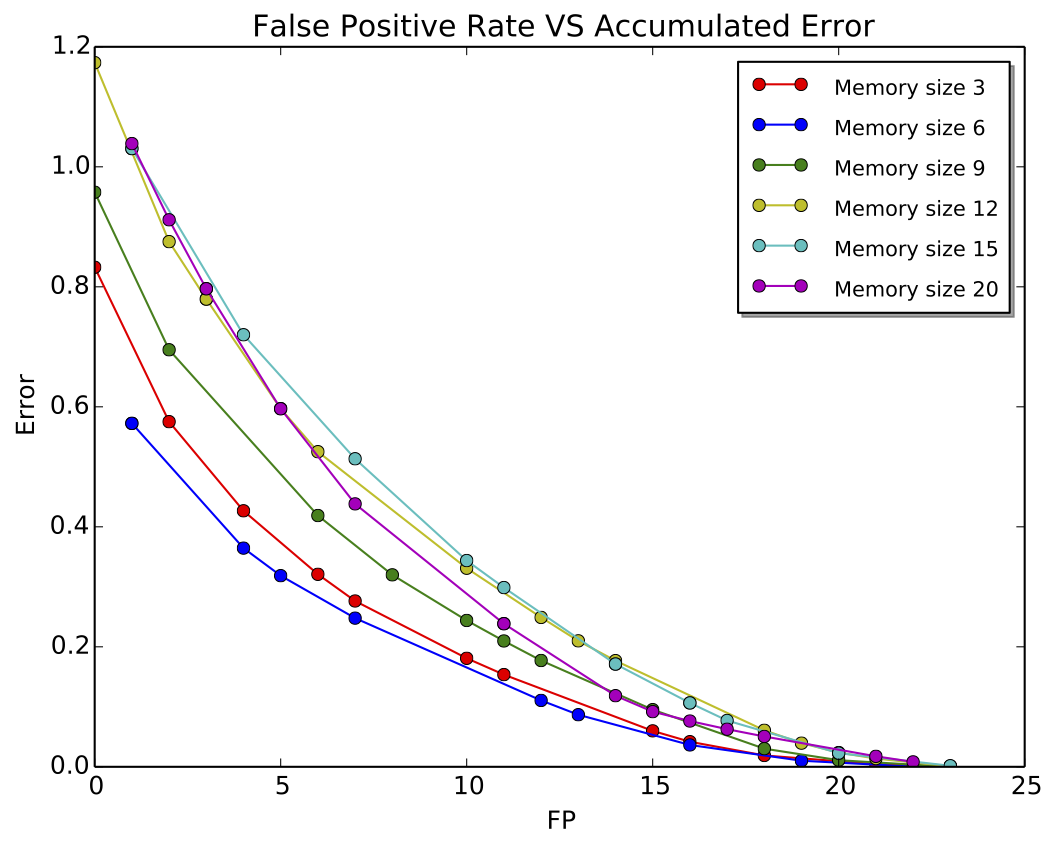}\label{fig:FPvsError_memory_euclidean_RLS_local_scaling_UCF}} \\
	\subfigure[][]{\includegraphics[width=.44\textwidth]{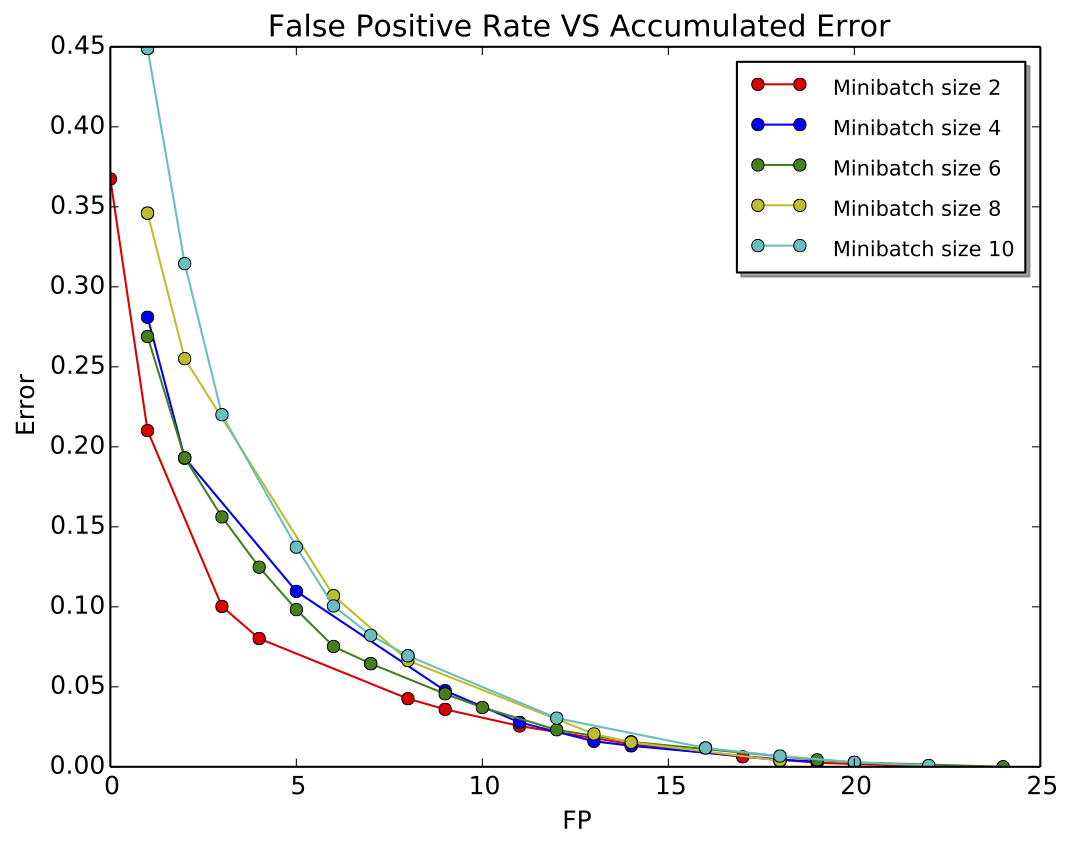}\label{fig:FPvsError_minibatch_DTW_RLS_local_scaling_UCF} }	
	\subfigure[][]{\includegraphics[width=.44\textwidth]{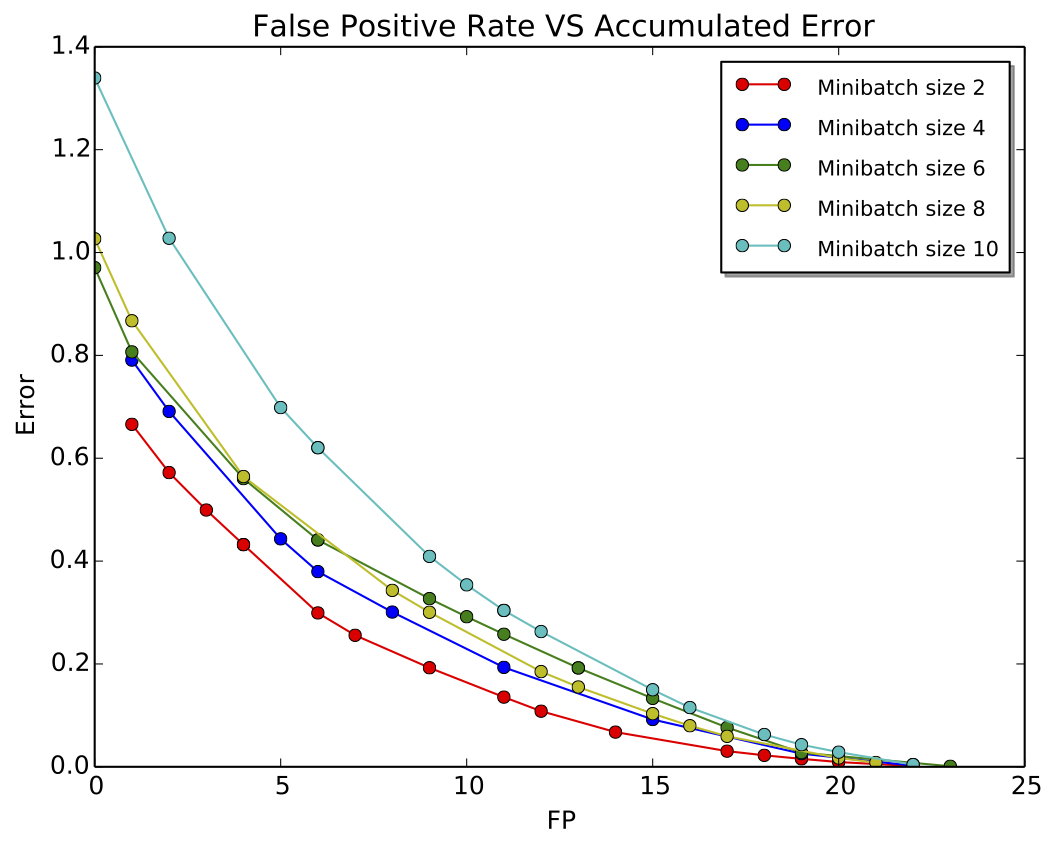}\label{fig:FPvsError_minibatch_euclidean_RLS_local_scaling_UCF}}
	\caption[False positive rate and accumulated error relation on C.S. scenario with RLS local scaling regularization for:
			\subref{fig:FPvsError_memory_DTW_RLS_local_scaling_UCF} \emph{memory} variation and DTW metric;			
			\subref{fig:FPvsError_memory_euclidean_RLS_local_scaling_UCF} \emph{memory} variation and euclidean metric;
			\subref{fig:FPvsError_minibatch_DTW_RLS_local_scaling_UCF} \emph{minibatch} variation and DTW metric;
			\subref{fig:FPvsError_minibatch_euclidean_RLS_local_scaling_UCF} \emph{minibatch} variation and euclidean metric.]{False positive rate and accumulated error relation on C.S. scenario with RLS local scaling regularization for:
			\subref{fig:FPvsError_memory_DTW_RLS_local_scaling_UCF} \emph{memory} variation and DTW metric;			
			\subref{fig:FPvsError_memory_euclidean_RLS_local_scaling_UCF} \emph{memory} variation and euclidean metric;
			\subref{fig:FPvsError_minibatch_DTW_RLS_local_scaling_UCF} \emph{minibatch} variation and DTW metric;
			\subref{fig:FPvsError_minibatch_euclidean_RLS_local_scaling_UCF} \emph{minibatch} variation and euclidean metric.}
	\label{fig:CS_FPvsError}	
\end{figure}

Next, we present some results for each dataset that will help us to answer the following questions:
\begin{inparaenum}[i)]
\item what is the influence of \emph{memory cell} size;
\item what is the influence of \emph{minibatch} size;
\item what is the influence of the regularisation matrix distance method on the assignment process;
\item what is the most suitable distance function.
\end{inparaenum}
We took conclusions based on the relation between the false positive rate and the accumulated error, varying the threshold for the incorrectly classified matches. The following results are presented accordingly with the regularisation step and the distance function. For instance, the chosen distance metrics between trajectories are Euclidean, Hausdorff, Dynamic Time Warping (DTW) and Longest Common Subsequence (LCS), while the regularisation that leads most times to the lowest error is selected. The distance matrix is normalised by the minimum and maximum, therefore the accumulated error for each metric can be compared. The distance functions are calculated considering four feature's trajectory, $(x,y,dx,dy)$, which are the point coordinates and the normalised vector direction between subsequent trajectory segments. Further individual analyses will be based on the results reported on Figures \ref{fig:CS_FPvsError}, \ref{fig:CU_FPvsError}, \ref{fig:DMT_FPvsError}, \ref{fig:SMT_FPvsError}.
 
\subsubsection{Crowd Structured (C.S.) Scenario} \label{sec:results_crowd_structured}
For this scenario a manual annotation of pedestrians was performed. Since the motion is structured, i.e. is represented by common motion patterns, for each one we selected and annotated several persons that undergo a motion pattern. At the end, we obtained representatives trajectories for each pattern, which led to 25 trajectories.

Both \emph{memory cell} size and \emph{minibatch} size have similar behaviour, namely lower values have less error and FP rate associated, therefore for this type of scenario \emph{memory} and \emph{minibatch} size should be kept low. We also notice that \emph{memory} size is less sensitive than \emph{minibatch} size, therefore \emph{memory} size could be higher than the \emph{minibatch} size. The chosen regularisation was the local scaling RLS, since, in general, it presents lower error. The DTW measure presents a steeper monotonically decreasing behaviour which shows a better compromise between the error and the FP rate. It also produces less false positives. Euclidean distance also performs well, while LCS measure presents the worst results, since it has almost a linear behaviour (see Figure \ref{fig:CS_FPvsError}).

\begin{figure}[h!]
\centering
	\subfigure[][]{\includegraphics[width=.15\textwidth]{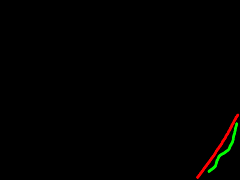}\label{fig:image_relation_01_UCF}}
	\subfigure[][]{\includegraphics[width=.15\textwidth]{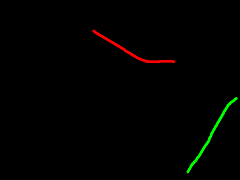}\label{fig:image_relation_02_UCF}}
	\subfigure[][]{\includegraphics[width=.15\textwidth]{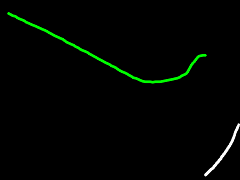}\label{fig:image_relation_03_UCF}}	
	\subfigure[][]{\includegraphics[width=.15\textwidth]{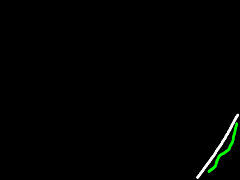}\label{fig:image_relation_04_UCF}}
	\subfigure[][]{\includegraphics[width=.15\textwidth]{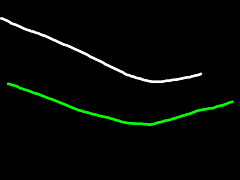}\label{fig:image_relation_05_UCF}}
	\subfigure[][]{\includegraphics[width=.15\textwidth]{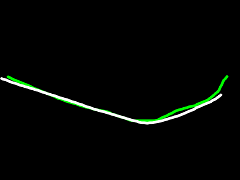}\label{fig:image_relation_06_UCF}}
	\caption[Assignment using DTW metric (\emph{minibatch}=2, \emph{memory}=10, C.S. scenario):
			\subref{fig:image_relation_01_UCF} false negative;			
			\subref{fig:image_relation_02_UCF} miss-match detected successfully;
			\subref{fig:image_relation_03_UCF} false positive not detected;
			\subref{fig:image_relation_04_UCF}, \subref{fig:image_relation_05_UCF}, \subref{fig:image_relation_06_UCF} matches detected.]{Assignment using DTW metric (\emph{minibatch}=2, \emph{memory}=10):
			\subref{fig:image_relation_01_UCF} false negative;			
			\subref{fig:image_relation_02_UCF} miss-match detected successfully;
			\subref{fig:image_relation_03_UCF} false positive not detected;
			\subref{fig:image_relation_04_UCF}, \subref{fig:image_relation_05_UCF}, \subref{fig:image_relation_06_UCF} matches detected.}
	\label{fig:assignments_UCF}
\end{figure}

Figure \ref{fig:assignments_UCF} presents some matching results, where manual trajectories are represented by green, and auto-generated trajectories are illustrated by red if a false positive is considered, or by white if a correct match is assigned. This nomenclature is kept for subsequent figures. Figures \ref{fig:image_relation_01_UCF} and \ref{fig:image_relation_04_UCF} show the threshold influence when deciding if it is a false positive or not, the former is a false negative assignment, and the latter is a correct match. Figure \ref{fig:image_relation_03_UCF} presents a wrong match, probably due to the global minimisation problem associated with the Hungarian algorithm. However, we state by Figure \ref{fig:image_relation_05_UCF} that our system is capable to generate a possible assignment for the previous miss match. We verify valid assignments and false positive detections. It is important to highlight that just some trajectories were annotated for this scenario, which has hundreds of trajectories that could be more similar to the automatic trajectories.

\subsubsection{Crowd Unstructured (C.U.) Scenario} \label{sec:results_crowd_unstructured}
This is the most challenging scenario for our system. It is a low resolution video that represents a crowded scene where pedestrians move randomly in various directions, causing constant occlusions. We manually annotated all the pedestrians, which lead to a total of 15 trajectories.

\begin{figure}[h!]
\centering
	\subfigure[][]{\includegraphics[width=.44\textwidth]{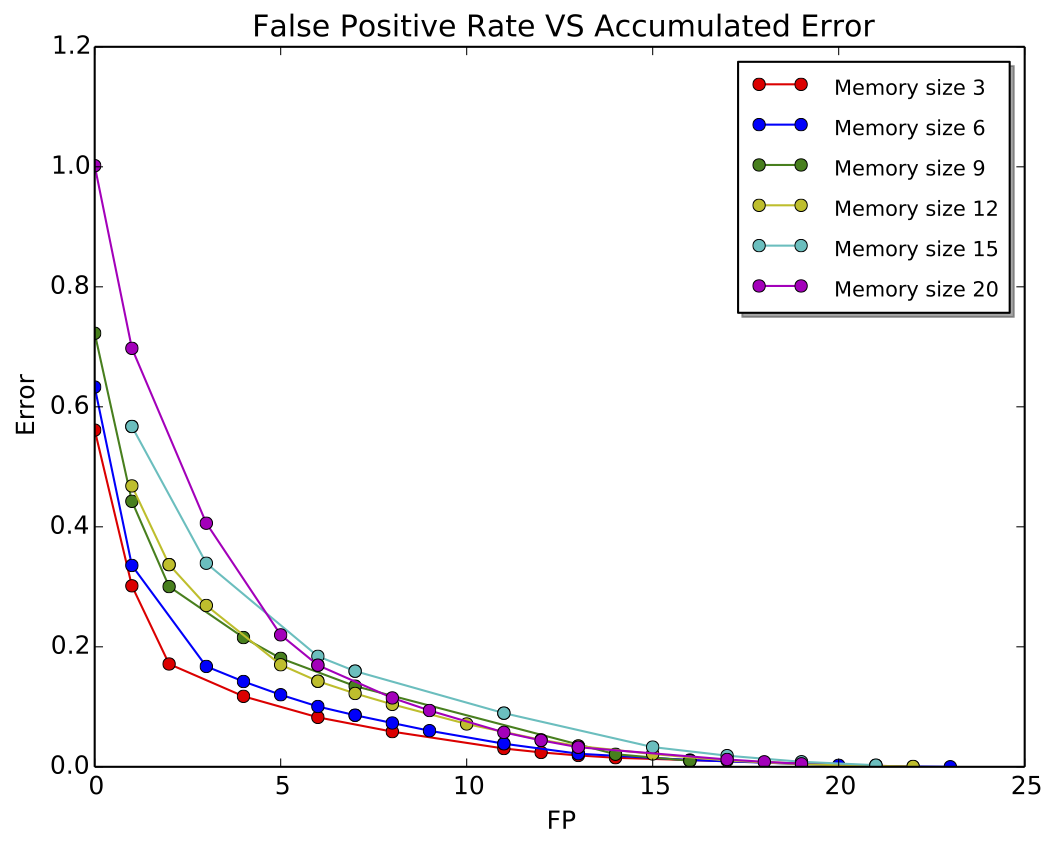}\label{fig:FPvsError_memory_DTW_RLS_median_MHA}}
	\subfigure[][]{\includegraphics[width=.44\textwidth]{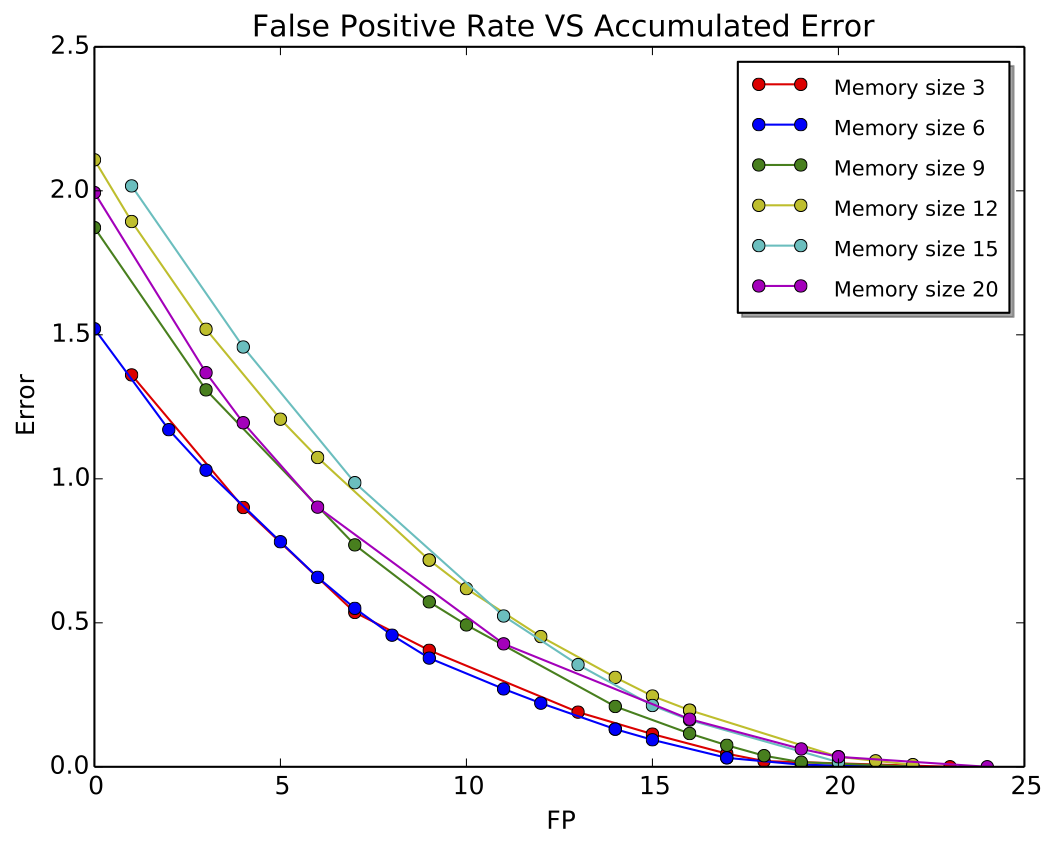}\label{fig:FPvsError_memory_hausdorff_RLS_median_MHA}}
	\subfigure[][]{\includegraphics[width=.44\textwidth]{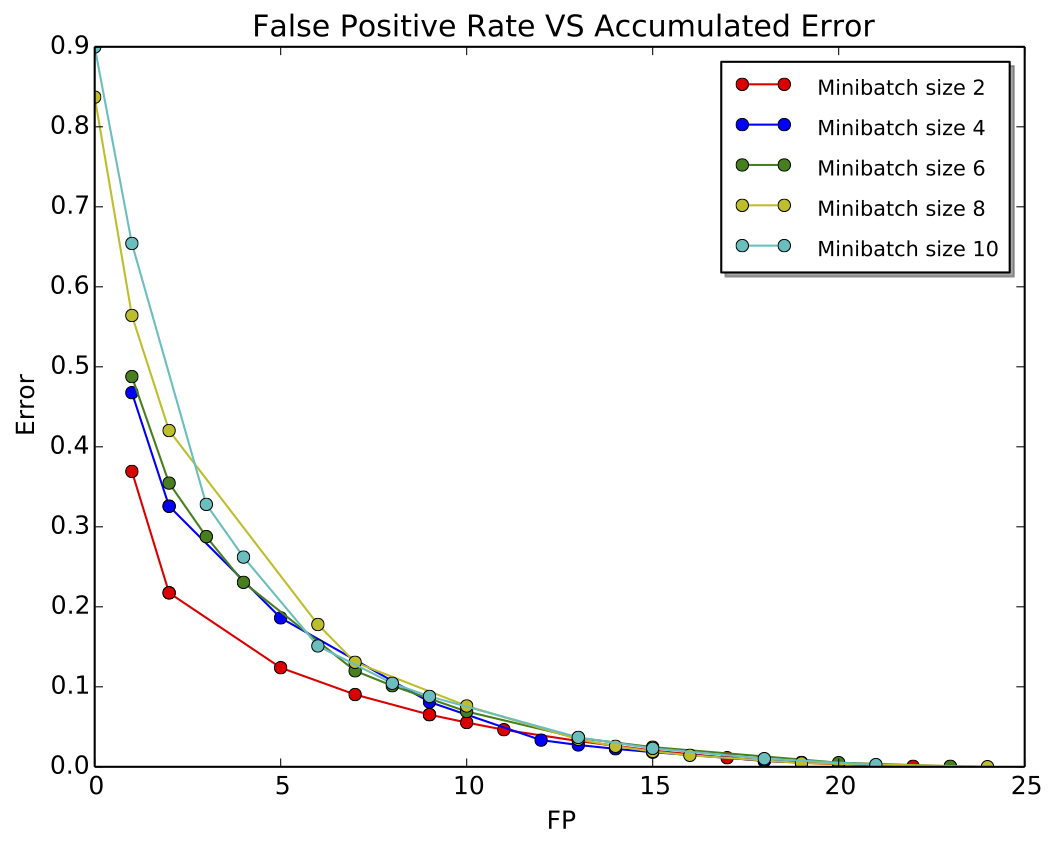}\label{fig:FPvsError_minibatch_DTW_RLS_median_MHA}}	
	\subfigure[][]{\includegraphics[width=.44\textwidth]{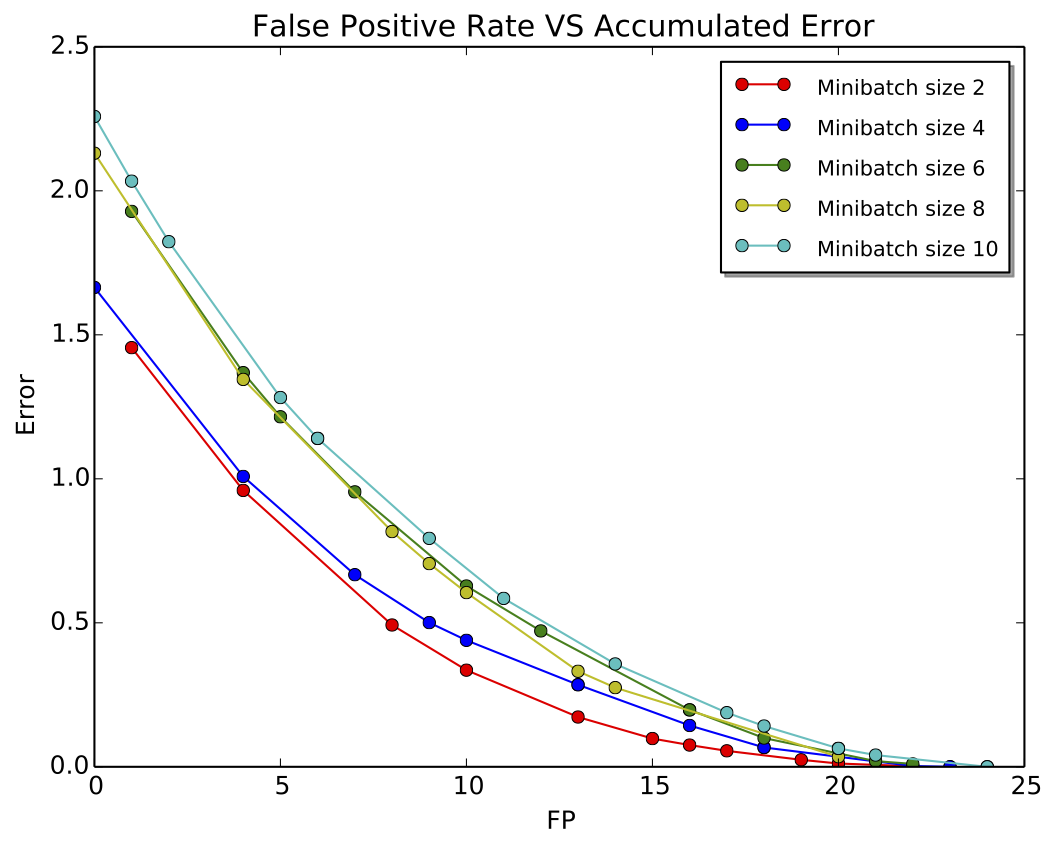}\label{fig:FPvsError_minibatch_hausdorff_RLS_median_MHA}}
	\caption[False positive rate and accumulated error relation on C.U. scenario with RLS median regularization for:
			\subref{fig:FPvsError_memory_DTW_RLS_median_MHA} \emph{memory} variation and DTW metric;			
			\subref{fig:FPvsError_memory_hausdorff_RLS_median_MHA} \emph{memory} variation and hausdorff metric;
			\subref{fig:FPvsError_minibatch_DTW_RLS_median_MHA} \emph{minibatch} variation and DTW metric;
			\subref{fig:FPvsError_minibatch_hausdorff_RLS_median_MHA} \emph{minibatch} variation and hausdorff metric.]{False positive rate and accumulated error relation on C.U. scenario with RLS median regularization for:
			\subref{fig:FPvsError_memory_DTW_RLS_median_MHA} \emph{memory} variation and DTW metric;			
			\subref{fig:FPvsError_memory_hausdorff_RLS_median_MHA} \emph{memory} variation and hausdorff metric;
			\subref{fig:FPvsError_minibatch_DTW_RLS_median_MHA} \emph{minibatch} variation and DTW metric;
			\subref{fig:FPvsError_minibatch_hausdorff_RLS_median_MHA} \emph{minibatch} variation and hausdorff metric.}
	\label{fig:CU_FPvsError}	
\end{figure}

In this case, \emph{memory cell} size has a less stable behaviour assuming the relation between the error and the FP rate. However, in general lower values present better results. \emph{Minibatch} size has even more variability, and we verify that medium and large values perform better than lower values. The selected regularisation was the median RLS. In terms of distance function, we reached the same conclusion of the previous scenario, just highlighting the improvement of the Hausdorff metric, which shows a steeper curve on low thresholds, but maintains a large FP rate at higher thresholds (see Figure \ref{fig:CU_FPvsError}).

\begin{figure}[h!]
\centering
	\subfigure[][]{\includegraphics[width=.15\textwidth]{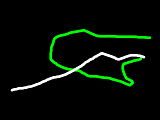}\label{fig:image_relation_01_MHA}}
	\subfigure[][]{\includegraphics[width=.15\textwidth]{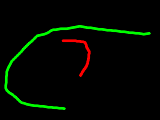}\label{fig:image_relation_02_MHA}}
	\subfigure[][]{\includegraphics[width=.15\textwidth]{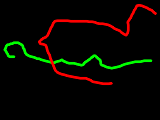}\label{fig:image_relation_03_MHA}}
	\subfigure[][]{\includegraphics[width=.15\textwidth]{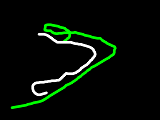}\label{fig:image_relation_04_MHA}}
	\subfigure[][]{\includegraphics[width=.15\textwidth]{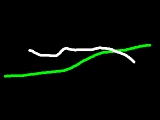}\label{fig:image_relation_05_MHA}}
	\subfigure[][]{\includegraphics[width=.15\textwidth]{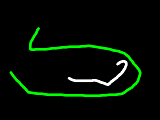}\label{fig:image_relation_06_MHA}}
	\caption[Assignment using DTW metric (\emph{minibatch}=8, \emph{memory}=10, C.U. scenario):
			\subref{fig:image_relation_01_MHA} false positive not detected;			
			\subref{fig:image_relation_02_MHA}, \subref{fig:image_relation_03_MHA} miss-match detected successfully;
			\subref{fig:image_relation_04_MHA}, \subref{fig:image_relation_05_MHA}, \subref{fig:image_relation_06_MHA} matches detected.]{Assignment using DTW metric (\emph{minibatch}=8, \emph{memory}=10):
			\subref{fig:image_relation_01_MHA} false positive not detected;			
			\subref{fig:image_relation_02_MHA}, \subref{fig:image_relation_03_MHA} miss-match detected successfully;
			\subref{fig:image_relation_04_MHA}, \subref{fig:image_relation_05_MHA}, \subref{fig:image_relation_06_MHA} matches detected.}
	\label{fig:assignments_MHA}
\end{figure}

In general, Figure \ref{fig:assignments_MHA} permits to take the same analysis as in the previous scenario. We visually verified that our system extracts trajectories that are very similar to the manual ones. However, our evaluation system was unable to consider them in the final matching results. 
This factor leads us to conclude that, despite the larger matching differences, the system performs well on such demanding scenario.

\subsubsection{Dense Multi-Tracking (D.MT.) Scenario} \label{sec:results_multitracking_dense}
This scenario is also very challenging. Normally, multi-tracking approaches are applied to it and flow-based approaches conduct to poor results. There are 19 manual trajectories to match.

\begin{figure}[h!]
\centering
	\subfigure[][]{\includegraphics[width=.44\textwidth]{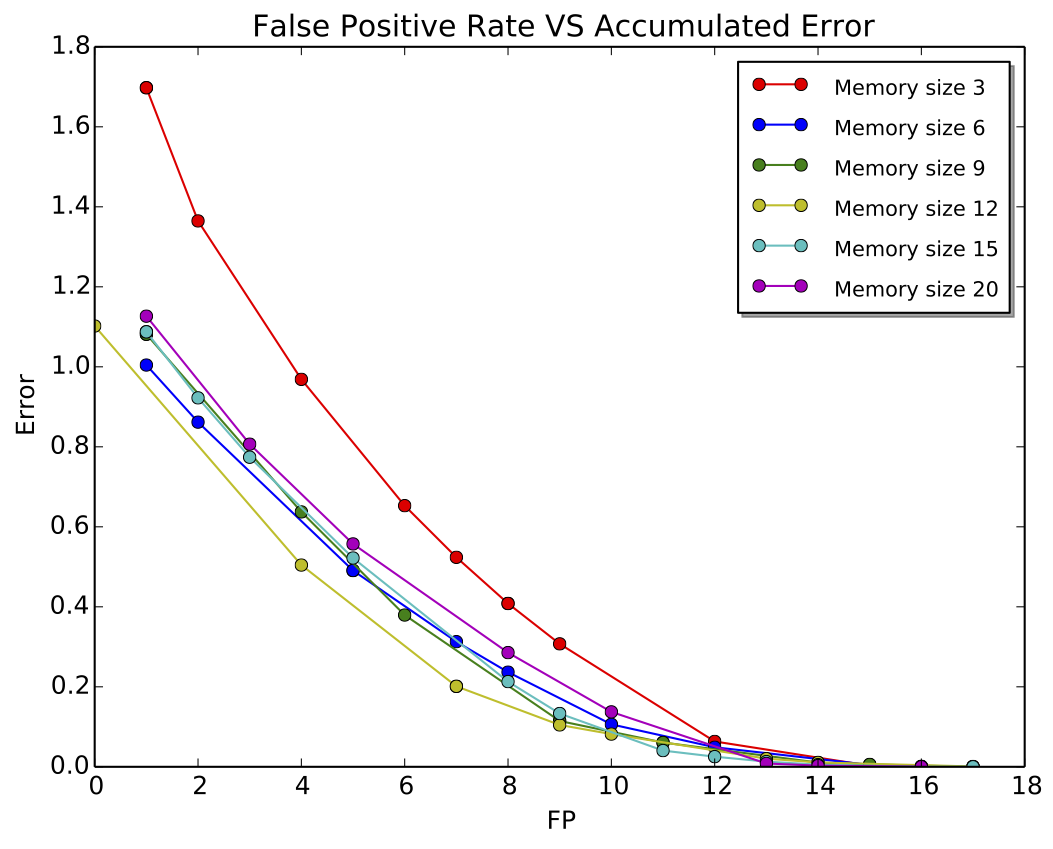}\label{fig:FPvsError_memory_DTW_RLS_median_PETS2013_S2_L1_time12-34_view001}}
	\subfigure[][]{\includegraphics[width=.44\textwidth]{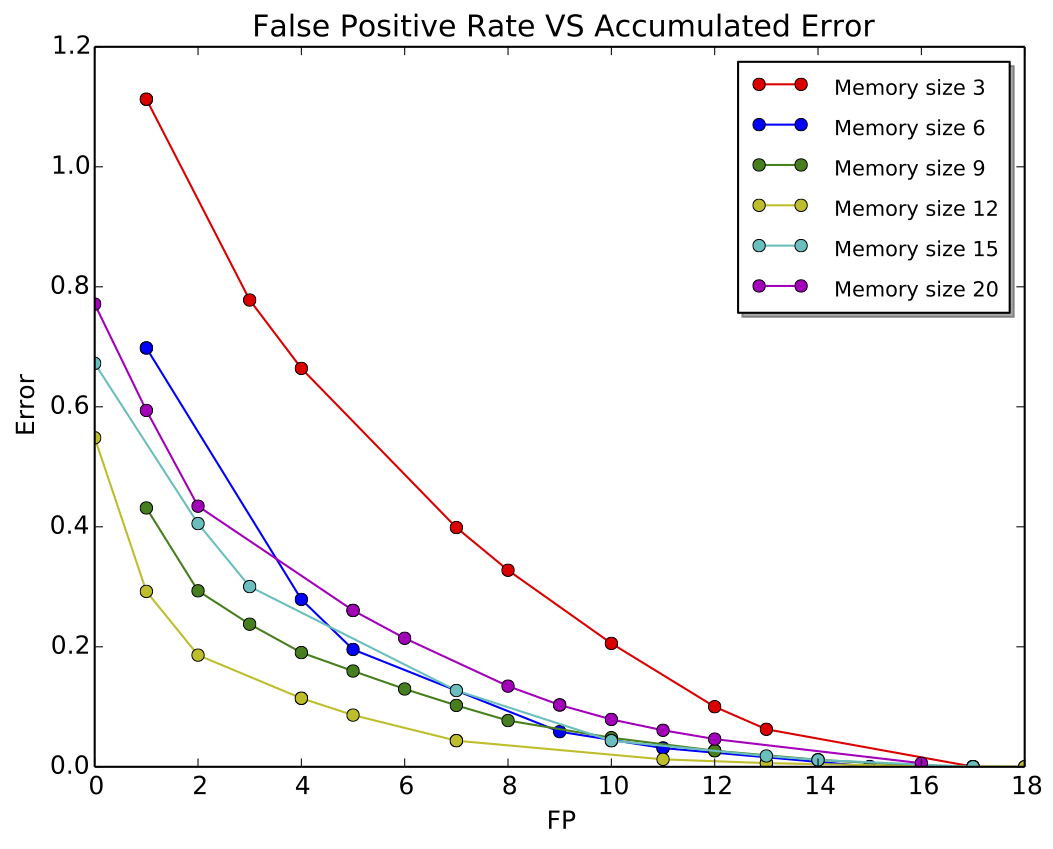}\label{fig:FPvsError_memory_hausdorff_RLS_median_PETS2013_S2_L1_time12-34_view001}}
	\subfigure[][]{\includegraphics[width=.44\textwidth]{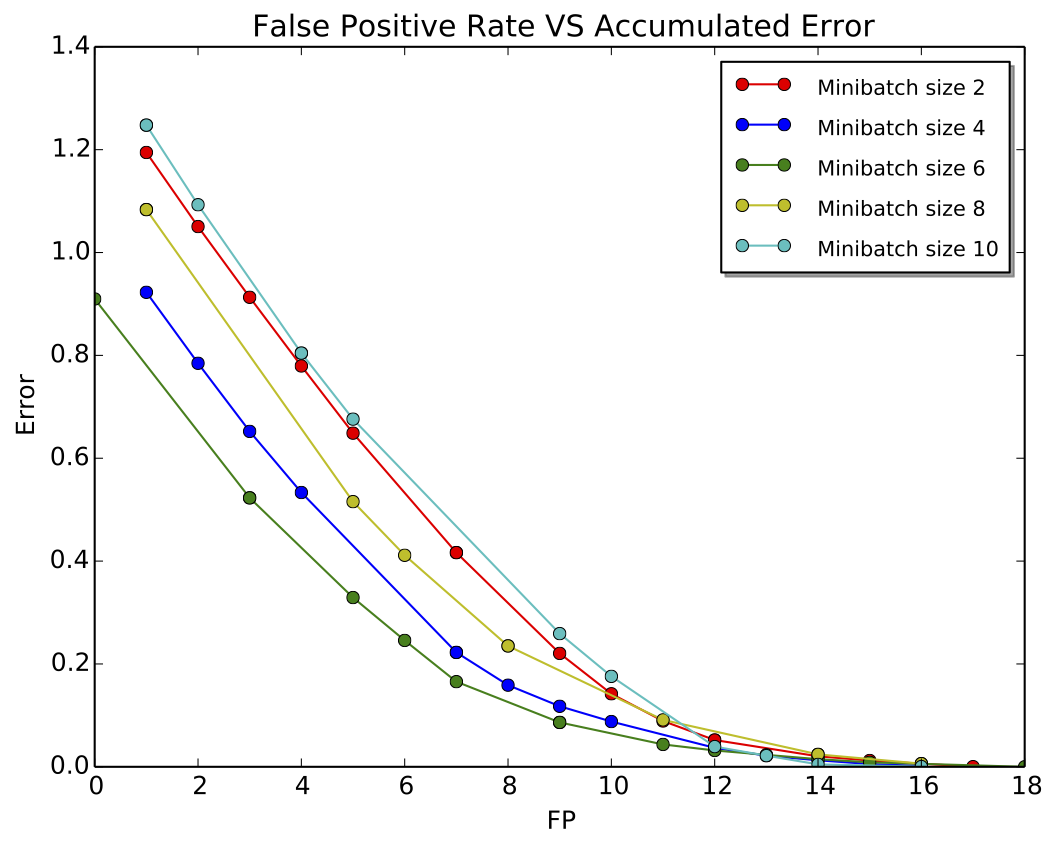}\label{fig:FPvsError_minibatch_DTW_RLS_median_PETS2013_S2_L1_time12-34_view001}}	
	\subfigure[][]{\includegraphics[width=.44\textwidth]{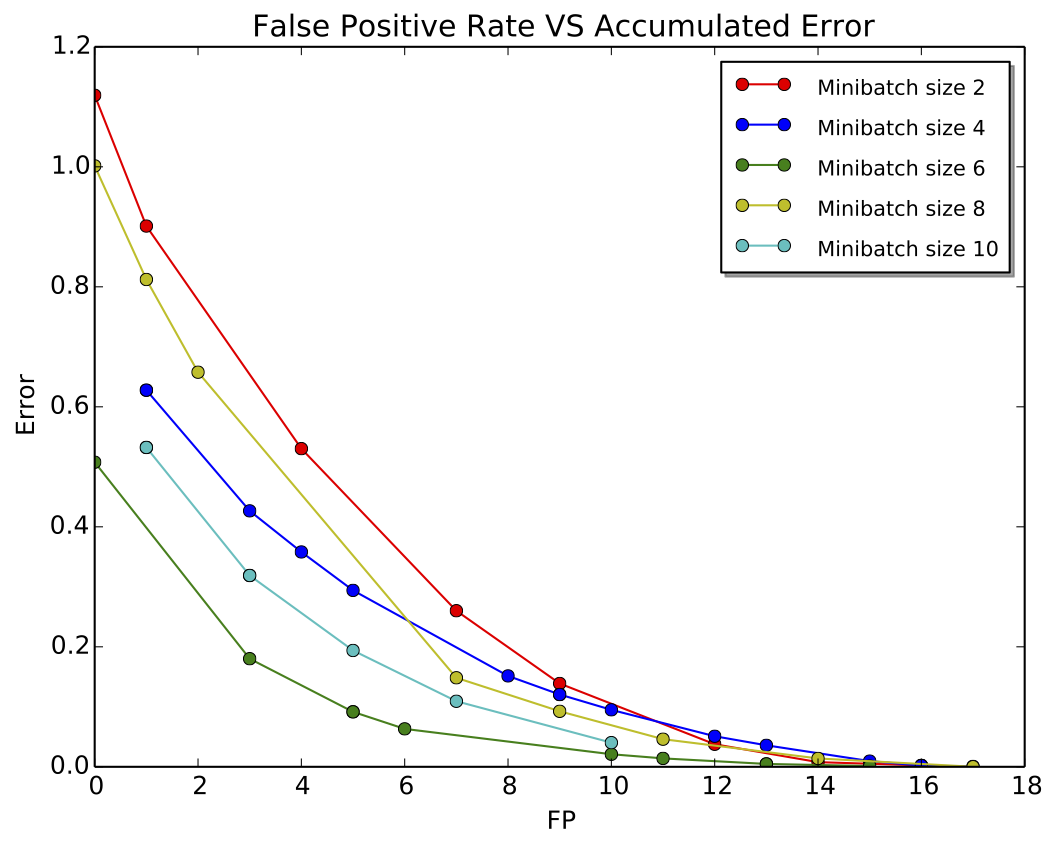}\label{fig:FPvsError_minibatch_hausdorff_RLS_median_PETS2013_S2_L1_time12-34_view001}}
	\caption[False positive rate and accumulated error relation on D.MT. scenario with RLS median regularization for:
			\subref{fig:FPvsError_memory_DTW_RLS_median_PETS2013_S2_L1_time12-34_view001} \emph{memory} variation and DTW metric;			
			\subref{fig:FPvsError_memory_hausdorff_RLS_median_PETS2013_S2_L1_time12-34_view001} \emph{memory} variation and hausdorff metric;
			\subref{fig:FPvsError_minibatch_DTW_RLS_median_PETS2013_S2_L1_time12-34_view001} \emph{minibatch} variation and DTW metric;
			\subref{fig:FPvsError_minibatch_hausdorff_RLS_median_PETS2013_S2_L1_time12-34_view001} \emph{minibatch} variation and hausdorff metric.]{False positive rate and accumulated error relation on D.MT. scenario with RLS median regularization for:
			\subref{fig:FPvsError_memory_DTW_RLS_median_PETS2013_S2_L1_time12-34_view001} \emph{memory} variation and DTW metric;			
			\subref{fig:FPvsError_memory_hausdorff_RLS_median_PETS2013_S2_L1_time12-34_view001} \emph{memory} variation and hausdorff metric;
			\subref{fig:FPvsError_minibatch_DTW_RLS_median_PETS2013_S2_L1_time12-34_view001} \emph{minibatch} variation and DTW metric;
			\subref{fig:FPvsError_minibatch_hausdorff_RLS_median_PETS2013_S2_L1_time12-34_view001} \emph{minibatch} variation and hausdorff metric.}
	\label{fig:DMT_FPvsError}
\end{figure}

An analysis shows that low values for \emph{memory cell} and \emph{minibatch} size decrease performance, while medium values perform better. The selected regularisation was also the median RLS. Regarding the distance functions, previous conclusions still hold true. The Hausdorff metric reaches the best performance, which presents a low and steeper error for low false positive rates (see Figure \ref{fig:DMT_FPvsError}).

\begin{figure}[h!]
\centering
	\subfigure[][]{\includegraphics[width=.15\textwidth]{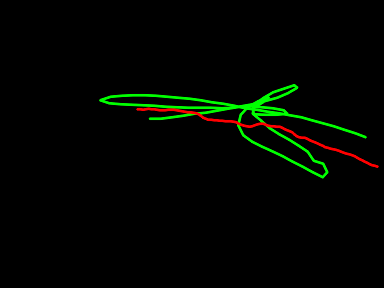}\label{fig:image_relation_01_PETS2013_S2_L1}}
	\subfigure[][]{\includegraphics[width=.15\textwidth]{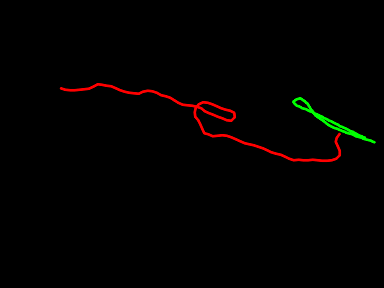}\label{fig:image_relation_02_PETS2013_S2_L1}}
	\subfigure[][]{\includegraphics[width=.15\textwidth]{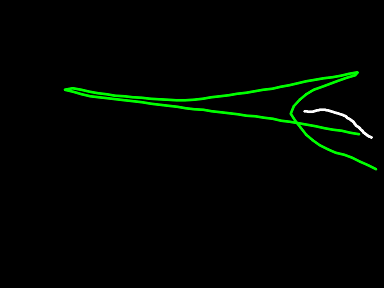}\label{fig:image_relation_03_PETS2013_S2_L1}}
	\subfigure[][]{\includegraphics[width=.15\textwidth]{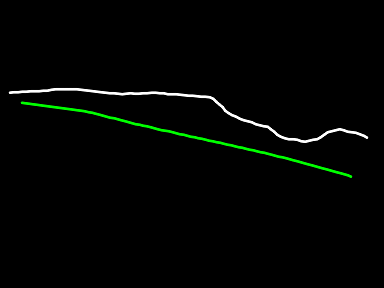}\label{fig:image_relation_04_PETS2013_S2_L1}}
	\subfigure[][]{\includegraphics[width=.15\textwidth]{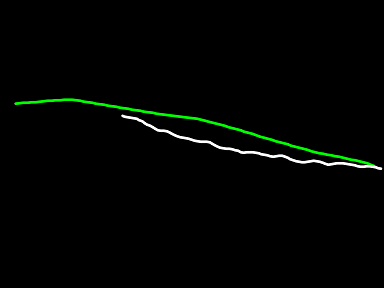}\label{fig:image_relation_05_PETS2013_S2_L1}}
	\subfigure[][]{\includegraphics[width=.15\textwidth]{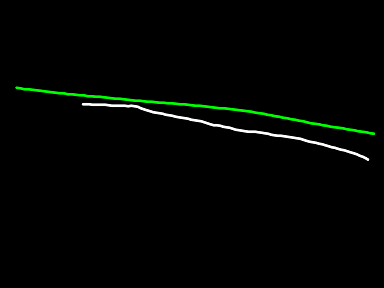}\label{fig:image_relation_06_PETS2013_S2_L1}}
	\caption[Assignment using hausdorff metric (\emph{minibatch}=5, \emph{memory}=12, D.MT. scenario):
			\subref{fig:image_relation_01_PETS2013_S2_L1}, \subref{fig:image_relation_02_PETS2013_S2_L1} miss-match detected successfully;
			\subref{fig:image_relation_03_PETS2013_S2_L1} false positive not detected;
			\subref{fig:image_relation_04_PETS2013_S2_L1}, \subref{fig:image_relation_05_PETS2013_S2_L1}, \subref{fig:image_relation_06_PETS2013_S2_L1} matches detected.]{Assignment using hausdorff metric (\emph{minibatch}=5, \emph{memory}=12):
			\subref{fig:image_relation_01_PETS2013_S2_L1}, \subref{fig:image_relation_02_PETS2013_S2_L1} miss-match detected successfully;
			\subref{fig:image_relation_03_PETS2013_S2_L1} false positive not detected;
			\subref{fig:image_relation_04_PETS2013_S2_L1}, \subref{fig:image_relation_05_PETS2013_S2_L1}, \subref{fig:image_relation_06_PETS2013_S2_L1} matches detected.}
	\label{fig:assignments_PETS2013_S2_L1}
\end{figure}

Figure \ref{fig:assignments_PETS2013_S2_L1} shows the most difficult trajectories to match in this dataset. In fact, Figure \ref{fig:image_relation_01_PETS2013_S2_L1} presents a complex manual trajectory, which is very long and has several direction changes. However, it could be divided on two cross sections, where one of them is automatically captured by the system, nevertheless it could not be identified as a correct match. 

\begin{figure}[!ht]
\centering
	\subfigure[][]{\includegraphics[width=.44\textwidth]{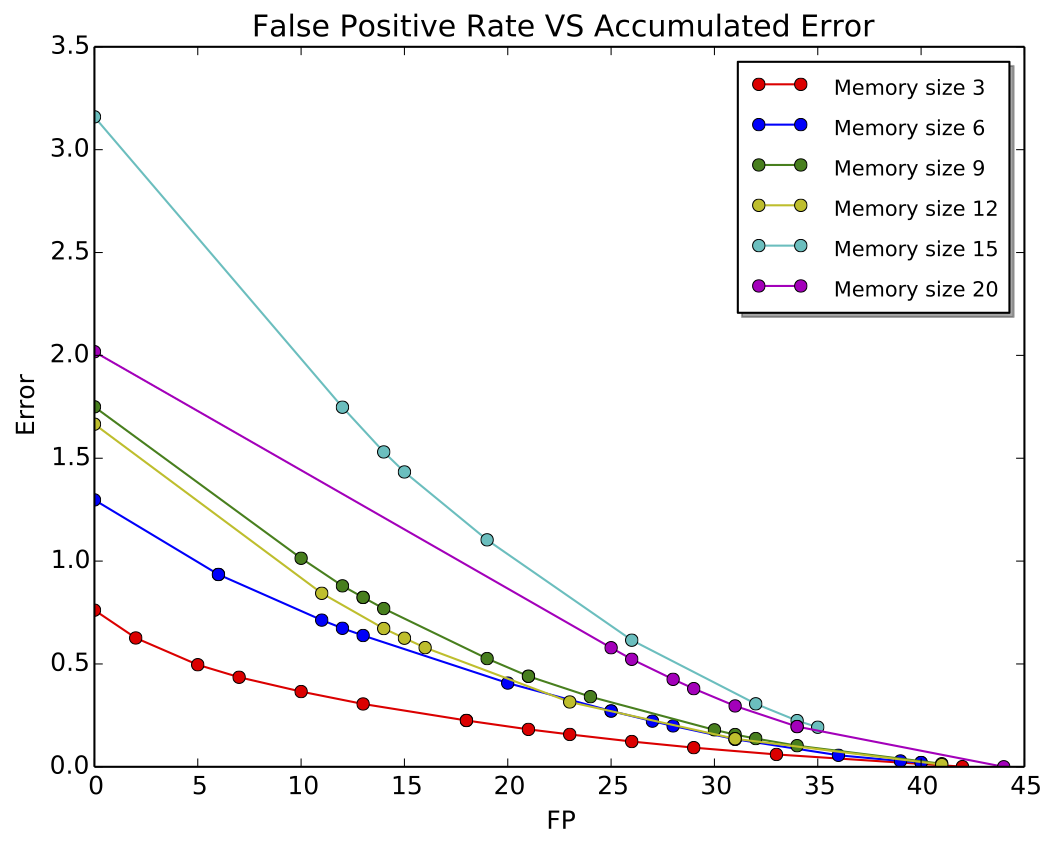}\label{fig:FPvsError_memory_DTW_truncated_cluster_PETS2013_S2_L3_time14-41_view001}}
	\subfigure[][]{\includegraphics[width=.44\textwidth]{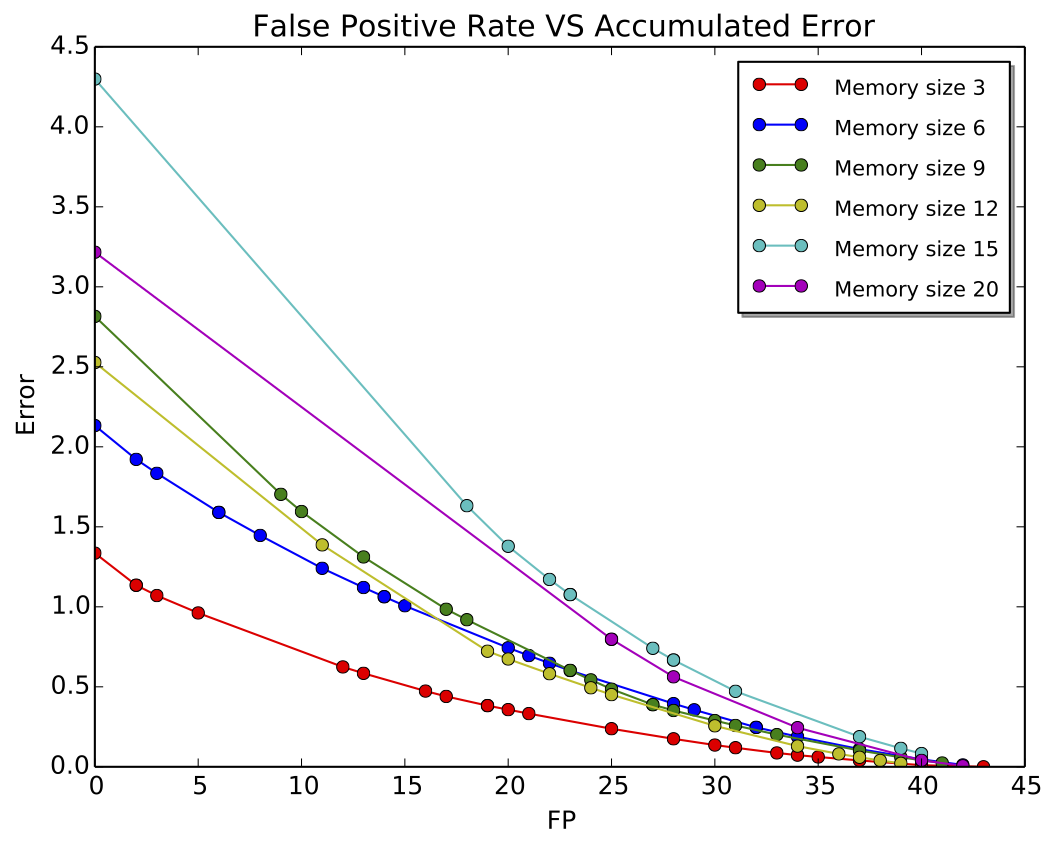}\label{fig:FPvsError_memory_euclidean_truncated_cluster_PETS2013_S2_L3_time14-41_view001}}
	\subfigure[][]{\includegraphics[width=.44\textwidth]{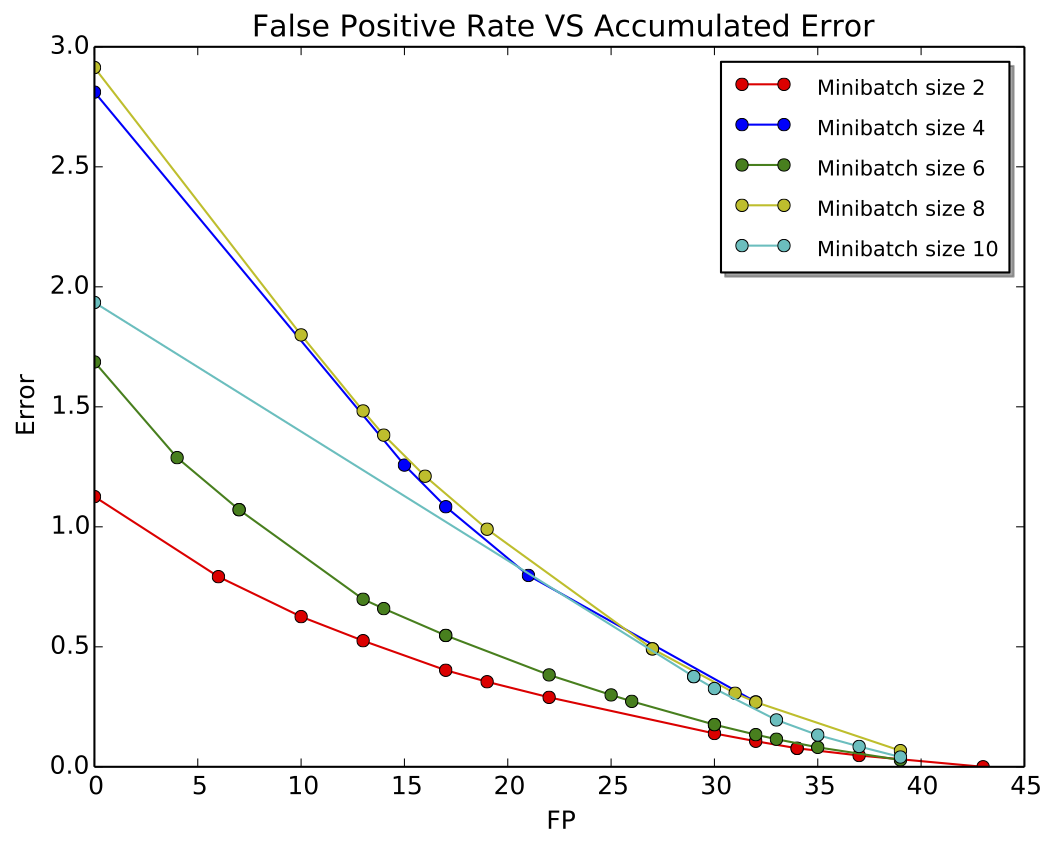}\label{fig:FPvsError_minibatch_DTW_truncated_cluster_PETS2013_S2_L3_time14-41_view001}}	
	\subfigure[][]{\includegraphics[width=.44\textwidth]{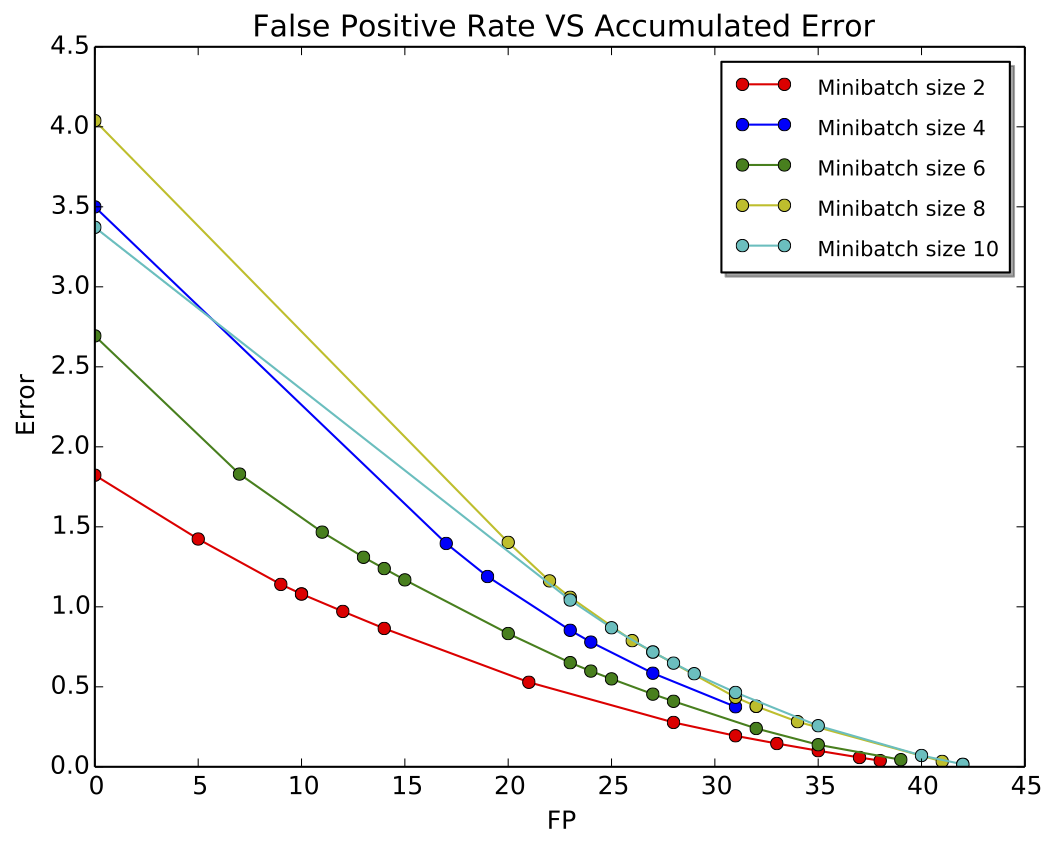}\label{fig:FPvsError_minibatch_euclidean_truncated_cluster_PETS2013_S2_L3_time14-41_view001}}
	\caption[False positive rate and accumulated error relation on S.MT. scenario with clustering threshold regularization for:
			\subref{fig:FPvsError_memory_DTW_truncated_cluster_PETS2013_S2_L3_time14-41_view001} \emph{memory} variation and DTW metric;			
			\subref{fig:FPvsError_memory_euclidean_truncated_cluster_PETS2013_S2_L3_time14-41_view001} \emph{memory} variation and euclidean metric;
			\subref{fig:FPvsError_minibatch_DTW_truncated_cluster_PETS2013_S2_L3_time14-41_view001} \emph{minibatch} variation and DTW metric;
			\subref{fig:FPvsError_minibatch_euclidean_truncated_cluster_PETS2013_S2_L3_time14-41_view001} \emph{minibatch} variation and euclidean metric.]{False positive rate and accumulated error relation on S.MT. scenario with clustering threshold regularization for:
			\subref{fig:FPvsError_memory_DTW_truncated_cluster_PETS2013_S2_L3_time14-41_view001} \emph{memory} variation and DTW metric;			
			\subref{fig:FPvsError_memory_euclidean_truncated_cluster_PETS2013_S2_L3_time14-41_view001} \emph{memory} variation and euclidean metric;
			\subref{fig:FPvsError_minibatch_DTW_truncated_cluster_PETS2013_S2_L3_time14-41_view001} \emph{minibatch} variation and DTW metric;
			\subref{fig:FPvsError_minibatch_euclidean_truncated_cluster_PETS2013_S2_L3_time14-41_view001} \emph{minibatch} variation and euclidean metric.}
	\label{fig:SMT_FPvsError}	
\end{figure}

\subsubsection{Sparse Multi-Tracking (S.MT.) Scenario} \label{sec:results_multitracking_sparse}
This is another scenario where multi-tracking approaches are normally applied. This dataset has an additional difficulty related to illumination variations that could lead to erroneous flow calculation. There are 44 manual trajectories.

The performance dependence on \emph{memory cell} and \emph{minibatch} is as in the S.C. scene. The regularisation of the distance matrix was based on the clustering threshold alternative, and the DTW was, again, the metric which presented the best results (see Figure \ref{fig:SMT_FPvsError}).

\begin{figure}[!ht]
\centering
	\subfigure[][]{\includegraphics[width=.15\textwidth]{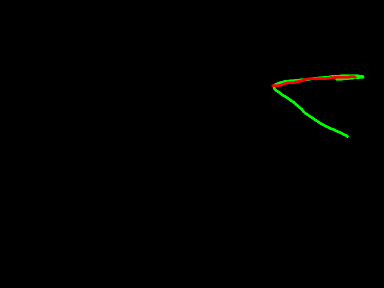}\label{fig:image_relation_01_PETS2013_S2_L3}}
	\subfigure[][]{\includegraphics[width=.15\textwidth]{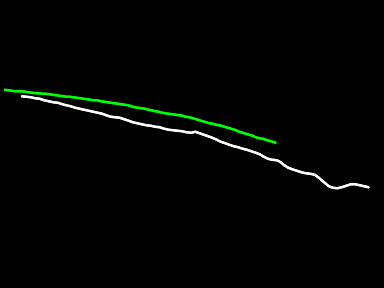}\label{fig:image_relation_02_PETS2013_S2_L3}}
	\subfigure[][]{\includegraphics[width=.15\textwidth]{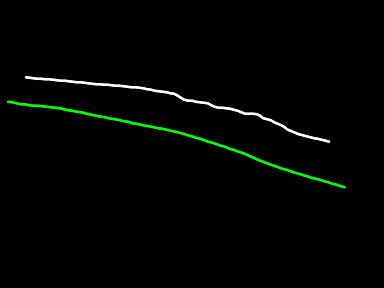}\label{fig:image_relation_03_PETS2013_S2_L3}}
	\subfigure[][]{\includegraphics[width=.15\textwidth]{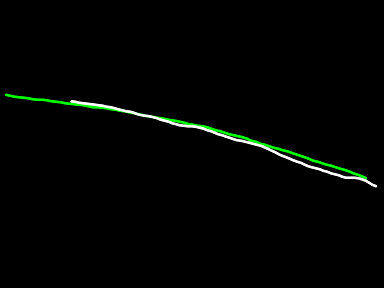}\label{fig:image_relation_04_PETS2013_S2_L3}}
	\subfigure[][]{\includegraphics[width=.15\textwidth]{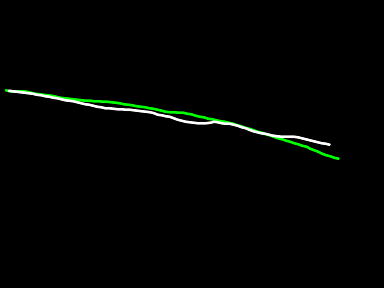}\label{fig:image_relation_05_PETS2013_S2_L3}}
	\subfigure[][]{\includegraphics[width=.15\textwidth]{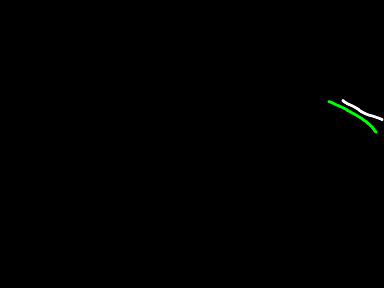}\label{fig:image_relation_06_PETS2013_S2_L3}}
	\caption[Assignment using DTW metric (\emph{minibatch}=5, \emph{memory}=3, S.MT. scenario):
			\subref{fig:image_relation_01_PETS2013_S2_L3} miss-match detected successfully;
			\subref{fig:image_relation_02_PETS2013_S2_L3}, \subref{fig:image_relation_03_PETS2013_S2_L3}, \subref{fig:image_relation_04_PETS2013_S2_L3}, \subref{fig:image_relation_05_PETS2013_S2_L3}, \subref{fig:image_relation_06_PETS2013_S2_L1} matches detected.]{Assignment using DTW metric (\emph{minibatch}=5, \emph{memory}=3):
			\subref{fig:image_relation_01_PETS2013_S2_L3} miss-match detected successfully;
			\subref{fig:image_relation_02_PETS2013_S2_L3}, \subref{fig:image_relation_03_PETS2013_S2_L3}, \subref{fig:image_relation_04_PETS2013_S2_L3}, \subref{fig:image_relation_05_PETS2013_S2_L3}, \subref{fig:image_relation_06_PETS2013_S2_L1} matches detected.}
	\label{fig:assignments_PETS2013_S2_L3}
\end{figure}

This dataset presents the best results (see Figure \ref{fig:assignments_PETS2013_S2_L3}), which is an excellent conclusion since it proves the versatility of the system. It correctly captures short and long trajectories, as well as false positives. However, we notice that the matches presented on Figures \ref{fig:image_relation_02_PETS2013_S2_L3} and \ref{fig:image_relation_03_PETS2013_S2_L3} could be inversely assigned to obtain a better matching.

As general evaluation, we conclude that low values should be considered for \emph{memory cell} size, while \emph{minibatch} size can be varied from low to medium values depending on scene category. Therefore, \emph{minibatch} size presents a larger functional interval, while avoiding error propagation derived from numeric calculations of the motion advection scheme. The lower values for \emph{memory cell} size could be explained by the intuition that it permits to capture less linear trajectories, eliminating the possibility to miss the detection of trajectory segments with high curvature, since the linking process could favour direction continuity. In terms of regularisation, it is obvious to conclude that a non-linear technique improves the results. In terms of the distance function, we verify that metrics that consider point-to-point geometrical information and shape benefit the matching process. 

\setlength{\tabcolsep}{0.1em}
\begin{table}[h!] 
\centering
\begin{tabular} {llll}

	\includegraphics[width=.23\textwidth]{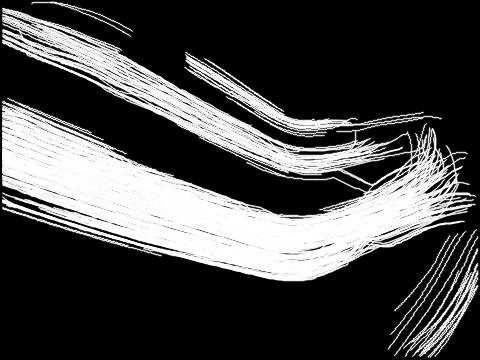}\label{fig:auto_UCF} &
	\includegraphics[width=.23\textwidth]{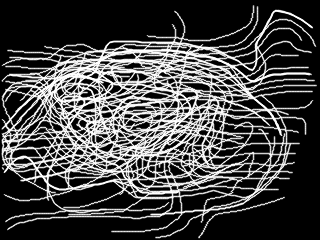}\label{fig:auto_MHA} &
	\includegraphics[width=.23\textwidth]{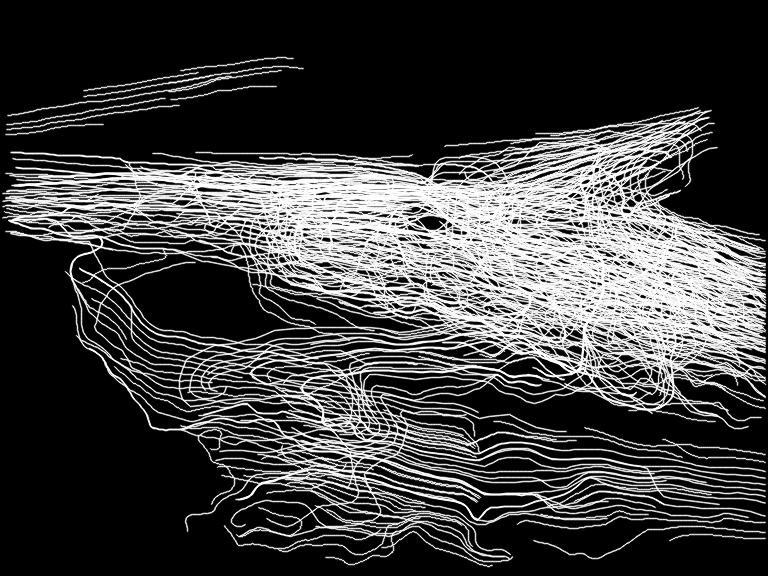}\label{fig:auto_PETS-12-34} &
	\includegraphics[width=.23\textwidth]{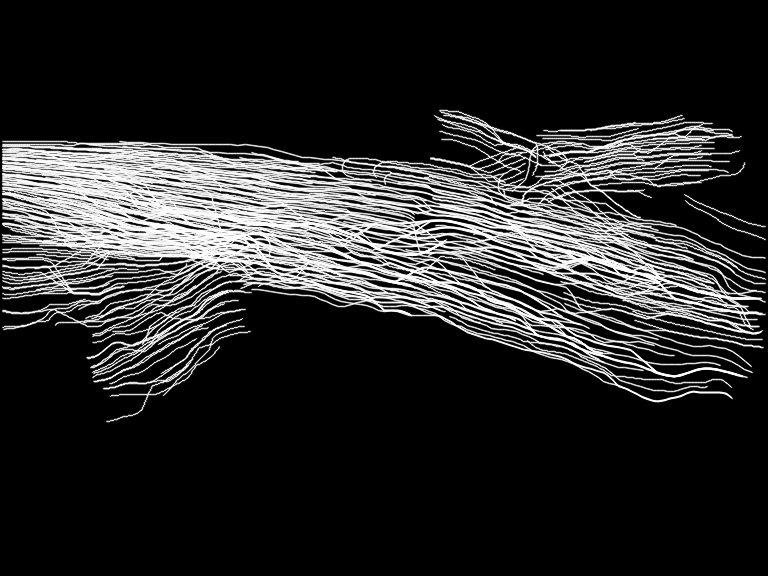}\label{fig:PETS-14-41}\\
	
	\includegraphics[width=.23\textwidth]{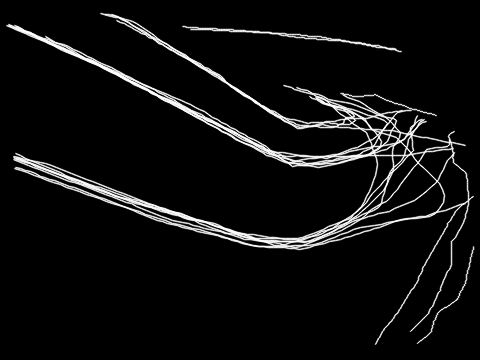}\label{fig:gt_UCF} &
	\includegraphics[width=.23\textwidth]{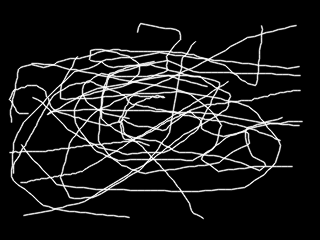}\label{fig:gt_MHA} &	
	\includegraphics[width=.23\textwidth]{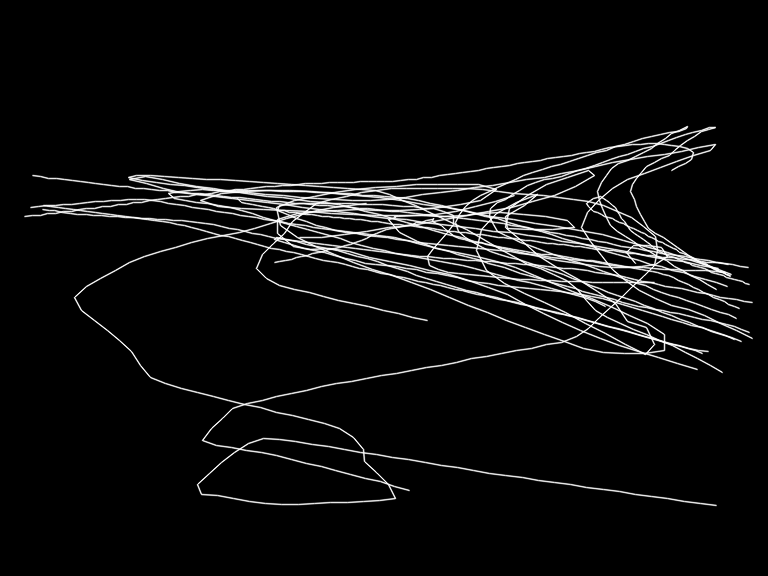}\label{fig:gt_PETS-12-34} &	
	\includegraphics[width=.23\textwidth]{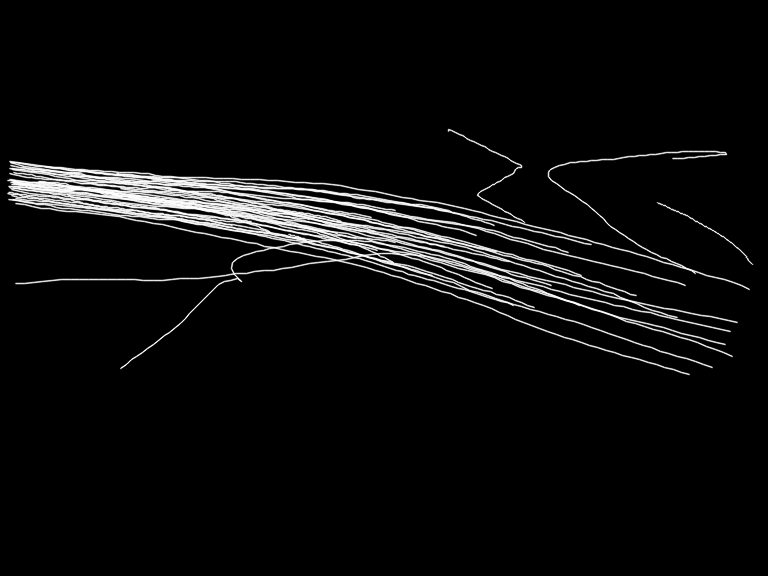}\label{fig:PETS-14-41}\\
	
\end{tabular}
	\captionof{figure}{Comparison between all automatic extracted trajectories and manually annotated trajectories. Top row: automatic. Bottom row: manual.} 
	\label{fig:autoVsgt}			
\end{table}

For qualitative validation, Figure \ref{fig:autoVsgt} presents the spatiotemporal comparison between the extracted trajectories and the manual ones. The density obtained clearly shows the robustness and efficiency of our system. Is important to mention that the assignment process always finds a correspondence between a manual trajectory and an auto-generated one, which corroborates the density evaluation.

\subsection{Motion Segmentation}\label{subsec:motion_segmentation}
This section reports the usefulness of our system related to human activity tasks. In this case, we evaluate our motion segmentation performance against two state-of-the-art works \cite{MehranMooreShah_ECCV10, conf/cvpr/AliS07}. We follow both the qualitative and quantitative analysis approach of \cite{MehranMooreShah_ECCV10} to present and compare the dynamic segmentations of different behaviour states through a video sequence. For the qualitative analysis, we used the same datasets as in \cite{MehranMooreShah_ECCV10}, namely Argentina and Boston \footnote[3]{\url{http://www.cs.ucf.edu/~ramin/?page_id=99}}. However, for the quantitative evaluation we do not report the results in the Boston dataset because it is a very time-consuming task to annotate all motion objects (14130 frames). We should highlight that we were unable to reproduce the results in \cite{MehranMooreShah_ECCV10}, even using the author's source code and the same parameterisation \footnote[4]{We acknowledge and thank the efforts of the first author of \cite{MehranMooreShah_ECCV10} to try to overcome such discrepancy, although without success.}. 

\begin{table}[h!] 
\centering
\begin{tabular} {lll}

	\includegraphics[width=.31\textwidth]{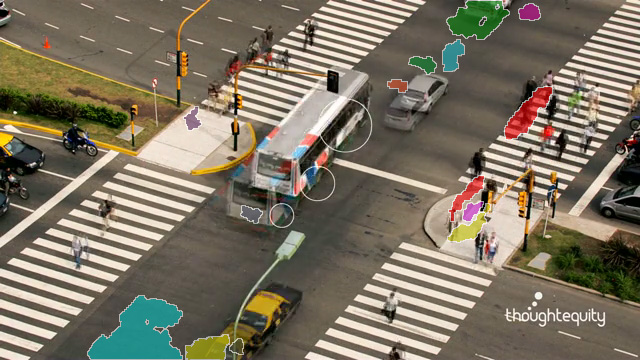}\label{fig:segmentation_115_pathlines} &
	\includegraphics[width=.31\textwidth]{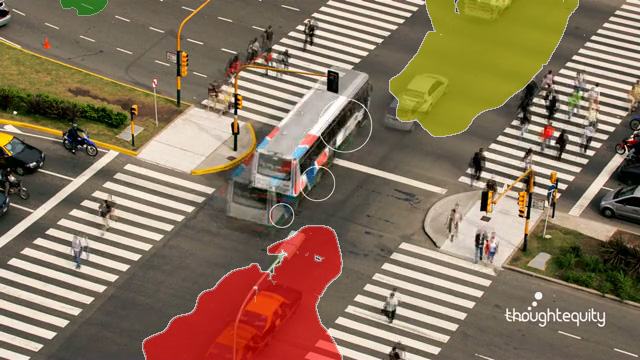}\label{fig:segmentation_115_streaklines} &
	\includegraphics[width=.31\textwidth]{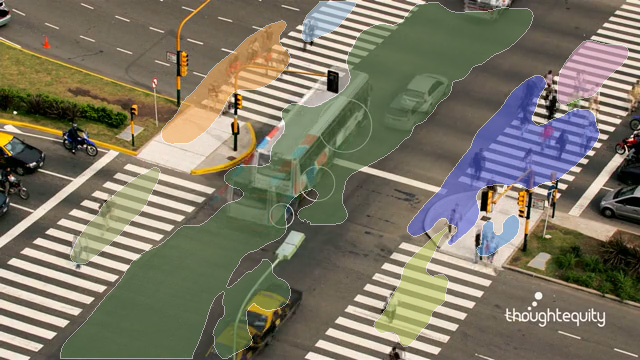}\label{fig:segmentation_115_our} \\
	
	\includegraphics[width=.31\textwidth]{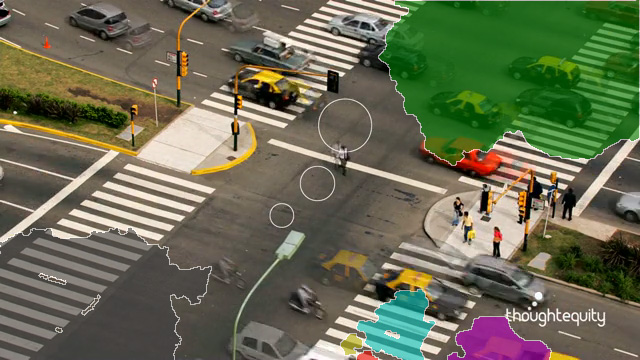}\label{fig:segmentation_213_pathlines} &
	\includegraphics[width=.31\textwidth]{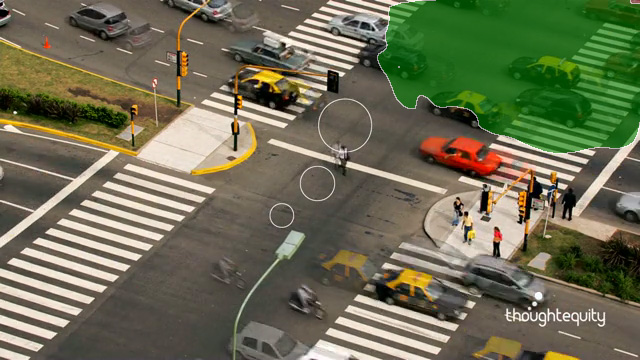}\label{fig:segmentation_213_streaklines} &	
	\includegraphics[width=.31\textwidth]{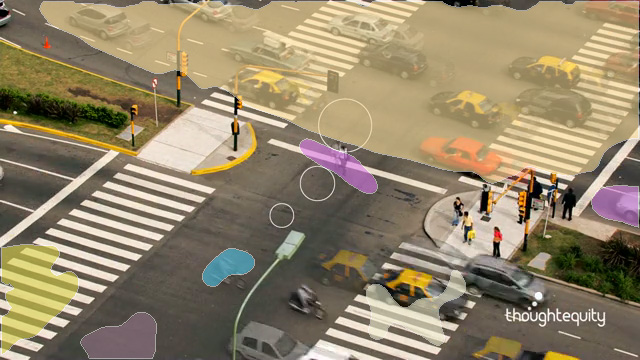}\label{fig:segmentation_213_our} \\
	
	\includegraphics[width=.31\textwidth]{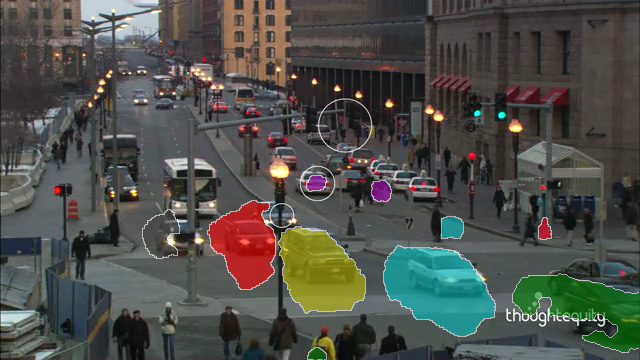}\label{fig:segmentation_213_pathlines} &
	\includegraphics[width=.31\textwidth]{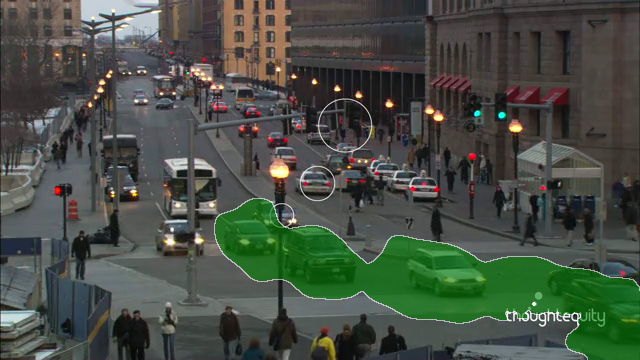}\label{fig:segmentation_213_streaklines} &	
	\includegraphics[width=.31\textwidth]{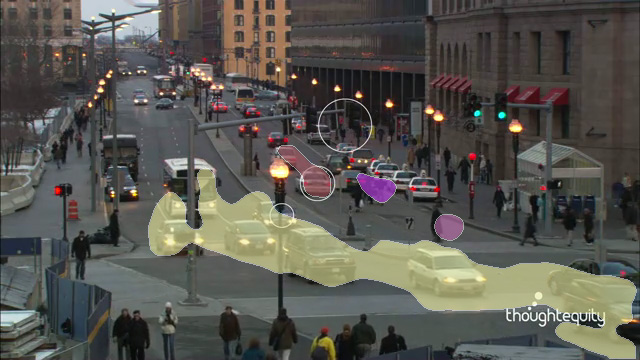}\label{fig:segmentation_213_our} \\
	
	\includegraphics[width=.31\textwidth]{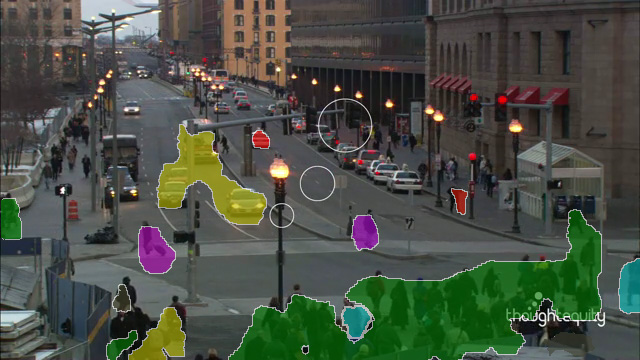}\label{fig:segmentation_213_pathlines} &
	\includegraphics[width=.31\textwidth]{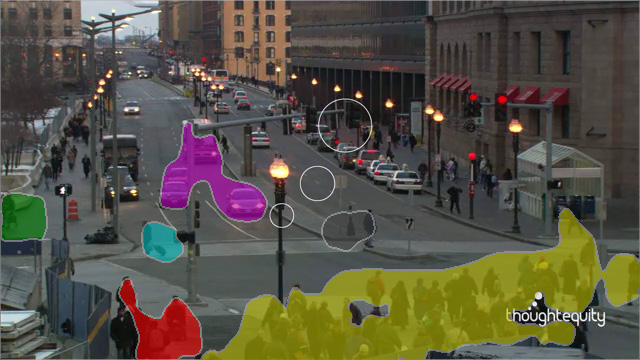}\label{fig:segmentation_213_streaklines} &	
	\includegraphics[width=.31\textwidth]{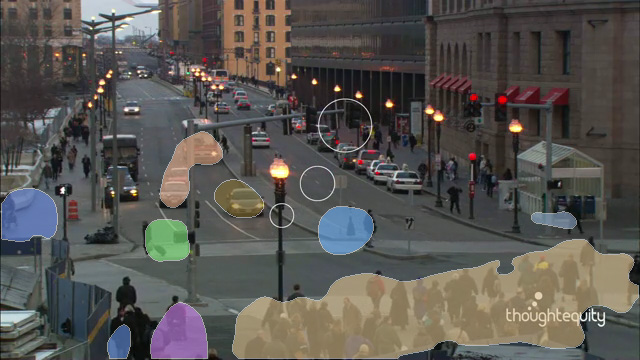}\label{fig:segmentation_213_our} \\
	
	\includegraphics[width=.31\textwidth]{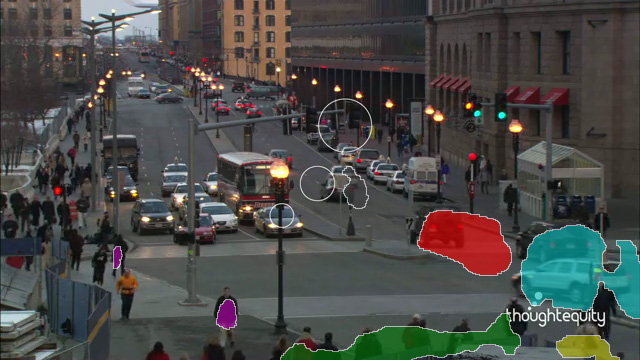}\label{fig:segmentation_213_pathlines} &
	\includegraphics[width=.31\textwidth]{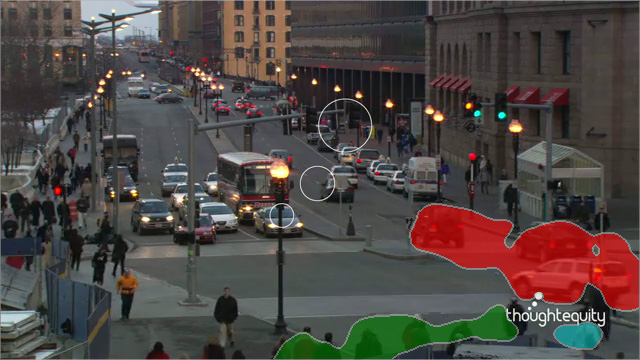}\label{fig:segmentation_213_streaklines} &	
	\includegraphics[width=.31\textwidth]{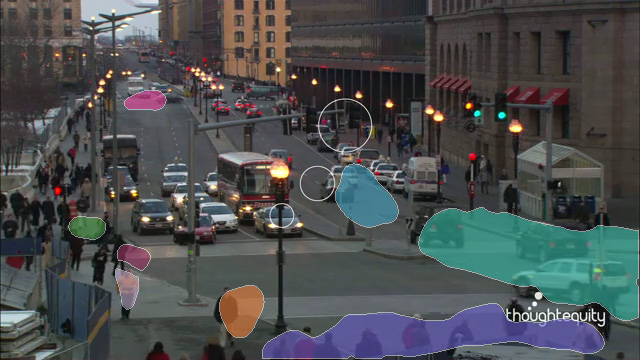}\label{fig:segmentation_213_our} \\
	
\end{tabular}
	\captionof{figure}[Qualitative comparison of segmentation results in Argentina and Boston datasets. By row: frame 115 (1st row), and frame 213 (2nd row) of Argentina; frame 40 (3rd row), frame 433 (4th row), and frame 2042 (5th row) of Boston. By column: from left to right, pathlines, streaklines, and our approach.]{Qualitative comparison of segmentation results in Argentina and Boston datasets. By row: frame 115 (1st row), and frame 213 (2nd row) of Argentina; frame 40 (3rd row), frame 433 (4th row), and frame 2042 (5th row) of Boston. By column: from left to right, pathlines, streaklines, and our approach.} 
	\label{fig:qualitative_segmentation}			
\end{table}

For the experiments, we use the following parameters for each approach: 
\begin{inparaenum}[i)]
\item pathlines, \emph{integration time}=15;
\item streaklines, \emph{streak length}=40;
\item our, \emph{minibatch}=3, \emph{memory}=5.
\end{inparaenum}
 In order for the comparison to be as fair as possible we use the same optical flow result per pair of frames for each approach, namely the Classic+NL from \cite{Sun10secretsof}.
 
Frames 115 and 213 illustrate two behavioural phases in which traffic lights change and north/south flow of pedestrians emerges, and west/east and north/south bound vehicles flow develop. Remaining frames refer to Boston dataset and also consider different behaviors when traffic light changes an east/west flow of pedestrians emerges simultaneously with a north/south bound vehicle flow. Figure \ref{fig:qualitative_segmentation} demonstrates that streaklines are spatially and temporally pronounced and more accurate on capturing dynamic objects than the pathlines, which show fragmented segments of movement. However, our approach clearly shows longer motions and robustly segments different flows, even on cluttered conditions. For instance, our approach is able to detect flow of pedestrians on each sidewalks and distinguish them from north and south bound (frame 115), and detect and separate standalone motions (frames 40, 433 and 2042). None of the other approaches are able to do this, even inspecting the results reported in \cite{MehranMooreShah_ECCV10}. We clarify that our segmentation approach is computed per pixel, considering the similarity of the cosine difference in a 8-connected neighborhood, i.e. it follows the same process of the streak flow similarity of \cite{MehranMooreShah_ECCV10}. In our case, the flow is derived from the long-range trajectories, where each one is resampled and their segments are considered as the flow vectors.

\begin{figure}[h!]
\centering
	\subfigure[][]{\includegraphics[width=.32\textwidth]{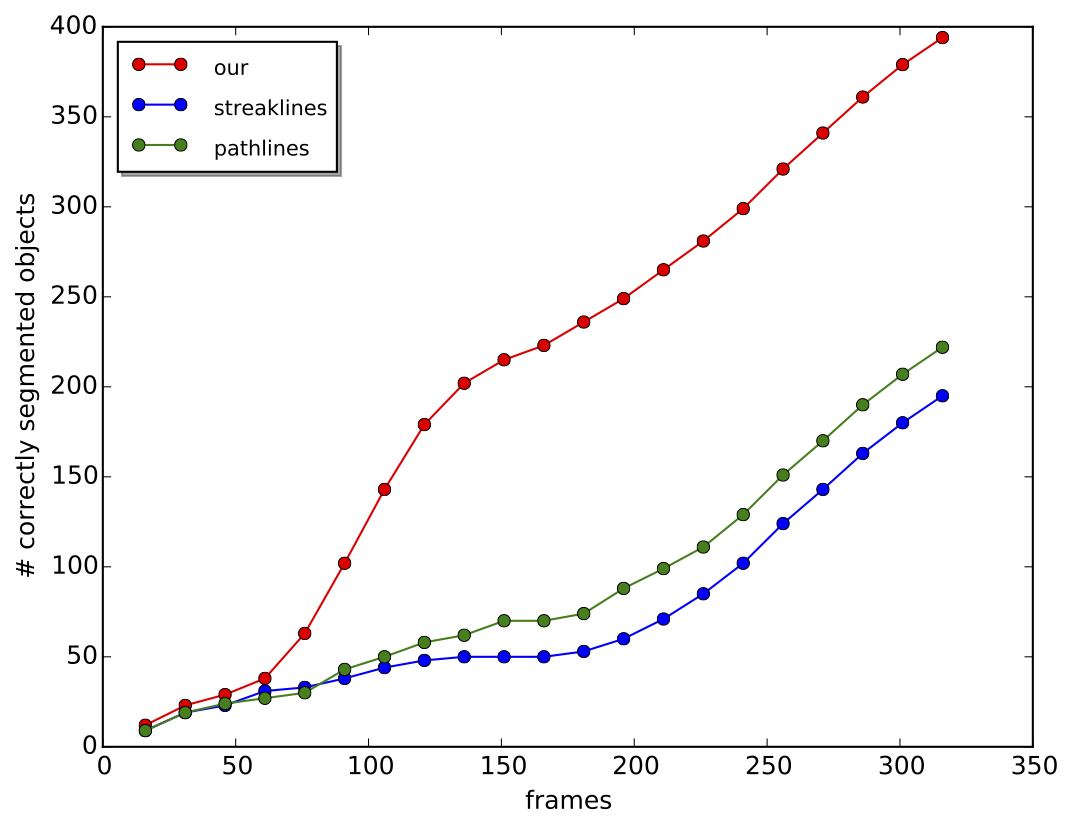}\label{fig:correctly_detected}}
	\subfigure[][]{\includegraphics[width=.32\textwidth]{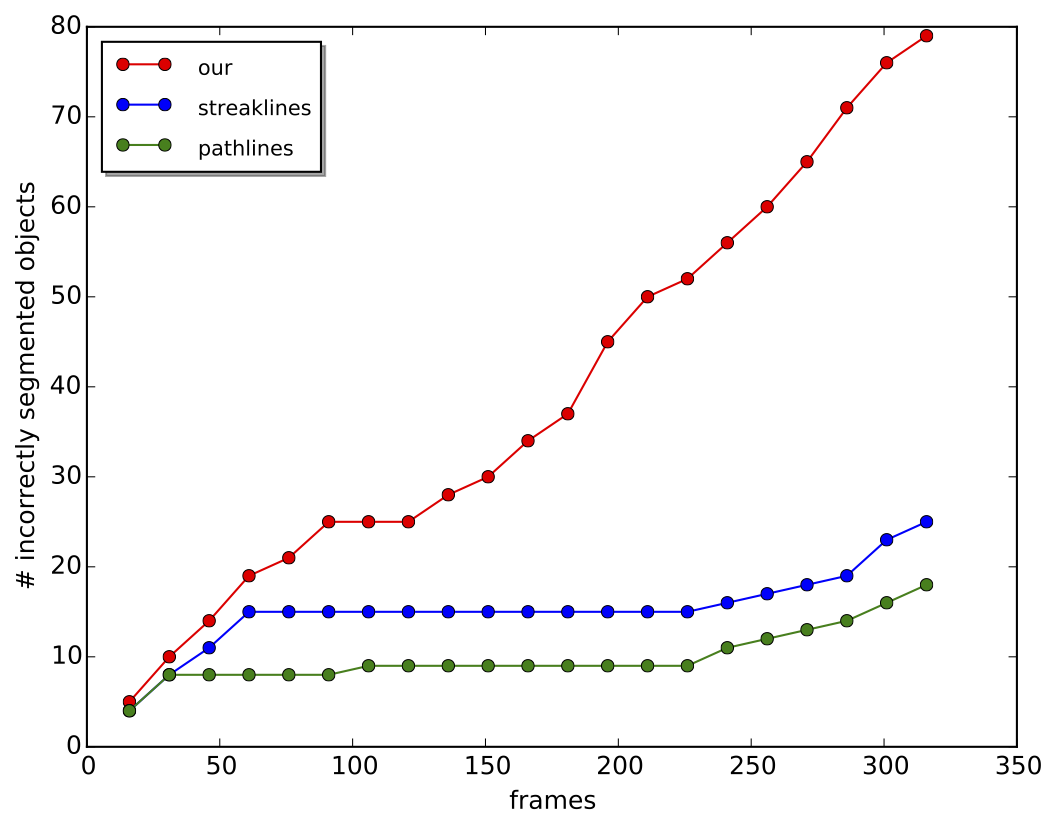}\label{fig:incorrectly_detected}}
	\subfigure[][]{\includegraphics[width=.32\textwidth]{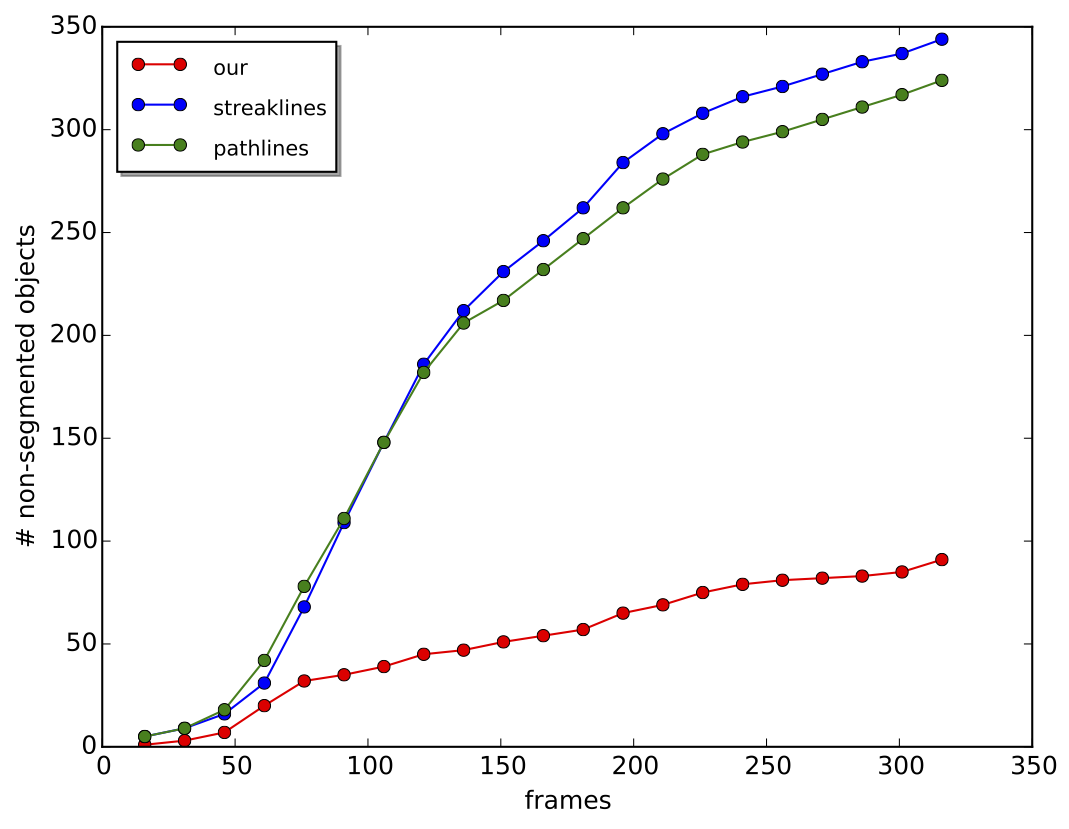}\label{fig:non_detected}}
	\caption[Quantitative comparison of segmentation results in Argentina dataset:
			\subref{fig:correctly_detected} correctly segmented objects;
			\subref{fig:incorrectly_detected}, incorrectly segmented objects; \subref{fig:non_detected}, non-segmented objects.]{Quantitative comparison of segmentation results in Argentina dataset:
			\subref{fig:correctly_detected} correctly segmented objects;
			\subref{fig:incorrectly_detected}, incorrectly segmented objects; \subref{fig:non_detected}, non-segmented objects.}
	\label{fig:quantitative_segmentation}
\end{figure}

For the quantitative comparison we also follow the same criterion stated in \cite{MehranMooreShah_ECCV10}. Our approach outperforms by a large margin, more than twice, both state-of-the-art approaches in the number of correctly segmented objects, even considering the results reported in \cite{MehranMooreShah_ECCV10}. The number of non-segmented objects is less than three times of the reported by the remaining approaches, which corroborates the qualitative analysis. However, it also presents a higher number of incorrectly segmented objects, probably because an over-segmentation on clutter regions (see Figure \ref{fig:quantitative_segmentation}).

\section{Conclusion} \label{sec:conclusion}
We present a complete and novel system based on global dense flow and local motion information that extracts meaningful long-range trajectories. It main advantages are its efficiency on different scenarios, and the correlation between the extracted trajectories and the individual pedestrians movements. The system combines a macro and micro analysis of motion, that to the best of our knowledge was not explored before. This work presents important contributions on different stages, namely on the system formulation, motion extraction and propagation scheme, a novel technique for removal of flow vector outliers, a fine-to-coarse flow representation, and integration of local motion information into a global re-correlation algorithm at the tracklet-level to form long-range trajectories. We also show that the system provides robust spatiotemporal motion information that can be used for human activity tasks such as segmentation, where it outperforms state-of-the-art algorithms. 

An evaluation framework is proposed to deal with the matching problem between global and individual trajectories. We demonstrate very good results on different datasets. However, two aspects should be improved: the false positive rate, since sometimes they are wrongly detected; and the assignment process, since in some cases the system produces most similar trajectories than the ones that were assigned by the matching process.

More sequences should be used to optimise system's parameters for each scene context. In that way, we believe that the system could get even better results. For future work, we will conduct a deeper system component analysis, explore the fine-to-coarse representation to model different level of human-activity-related knowledge, and will use the trajectories extracted from other scene contexts, such as duo interaction and individual tracking, to verify their impact on action classification. We believe that detection and classification of human activity could benefit from a system like ours.

\section*{Acknowledgment} \label{sec:acknowledgment}
The first author would like to thank FCT - Funda\c{c}\~ao para a Ci\^encia e Tecnologia (Portuguese Foundation for Science and Technology) for the financial support for the PhD grant with reference SFRH/BD/51430/2011.

\section*{References}
\bibliography{\myreferences}

\end{document}